\documentclass{article} 
\usepackage{iclr2026_conference,times}
\usepackage[utf8]{inputenc} 
\usepackage[T1]{fontenc}    
\usepackage{url}            

\usepackage{tabularx}
\usepackage{caption}
\usepackage{adjustbox} 
\usepackage{booktabs}       
\usepackage{amsfonts}       
\usepackage{nicefrac}       
\usepackage{microtype}      
\usepackage{xcolor}         
\usepackage[T1]{fontenc}
\usepackage{latexsym}
\usepackage{inconsolata}
\usepackage{graphicx}
\usepackage{booktabs, tabularx, multirow, multicol, makecell, colortbl}
\usepackage{enumitem}
\usepackage{bbding}
\usepackage{amsmath, amssymb}
\usepackage{wrapfig}
\usepackage{graphicx}
\usepackage{subcaption} 
\usepackage{float} 
\definecolor{backblue}{RGB}{210, 230, 250}
\definecolor{backgreen}{RGB}{226, 240, 217}
\definecolor{backred}{RGB}{255, 223, 223}
\newcommand{\high}{\cellcolor{backblue}}
\newcommand{\highgreen}{\cellcolor{backgreen}}

\usepackage[colorlinks=true]{hyperref}

\usepackage{tabularx}
\usepackage{subcaption}
\usepackage{booktabs}
\usepackage{ascii}
\usepackage{newunicodechar}
\usepackage{multirow}
\usepackage{enumitem}
\usepackage{wrapfig}
\usepackage{longtable}
\makeatletter

\newcommand{\Rmnum}[1]{\expandafter\@slowromancap\romannumeral #1@}
\makeatother

\usepackage{amsmath,amsfonts,bm}

\def\eqref#1{equation~\ref{#1}}

\def\1{\bm{1}}

\DeclareMathAlphabet{\mathsfit}{\encodingdefault}{\sfdefault}{m}{sl}
\SetMathAlphabet{\mathsfit}{bold}{\encodingdefault}{\sfdefault}{bx}{n}

\usepackage{url}

\title{HSSBench: Benchmarking  \\Humanities and Social Sciences Ability \\for  Multimodal Large Language Models}

\author{Zhaolu Kang$^{1,2,*}$, Junhao Gong$^{1,*}$, \textbf{Jiaxu Yan}$^{2,4,*}$, \textbf{Wanke Xia}$^{3}$, \textbf{Yian Wang}$^{4}$, \textbf{Ziwen Wang}$^{2}$, \\\textbf{Huaxuan Ding}$^{2}$, \textbf{Zhuo Cheng}$^{5}$, \textbf{Wenhao Cao}$^{6}$,\textbf{Zhiyuan Feng}$^{3}$,
\textbf{Siqi He}$^{2}$, \textbf{Shannan Yan}$^{3}$,
\\\textbf{Junzhe Chen}$^{3}$,  \textbf{Xiaomin He}$^{1}$, \textbf{Chaoya Jiang}$^{1}$, \textbf{Wei Ye}$^{1,\dagger}$, \textbf{Kaidong Yu}$^{2,\dagger}$, \textbf{Xuelong Li}$^{2,\dagger}$
\\$^{1}$National Engineering Research Center for Software Engineering, Peking University 
\\$^{2}$Institute of Artificial Intelligence, China Telecom (TeleAI)
\\
$^{3}$Tsinghua University 
$^{4}$Chinese Academy of Sciences\\
$^{5}$University of British Columbia \
$^{6}$Renmin University of China
\\
\texttt{zlkang25@stu.pku.edu.cn}
}

\footnotetext[1]{Equal Contribution. $^{\dagger}$Corresponding Authors.}

\iclrfinalcopy 
\begin{document}

\maketitle

\vspace{-2em}
\begin{figure*}[hbt!]
    \centering
    \includegraphics[width=0.9\linewidth]{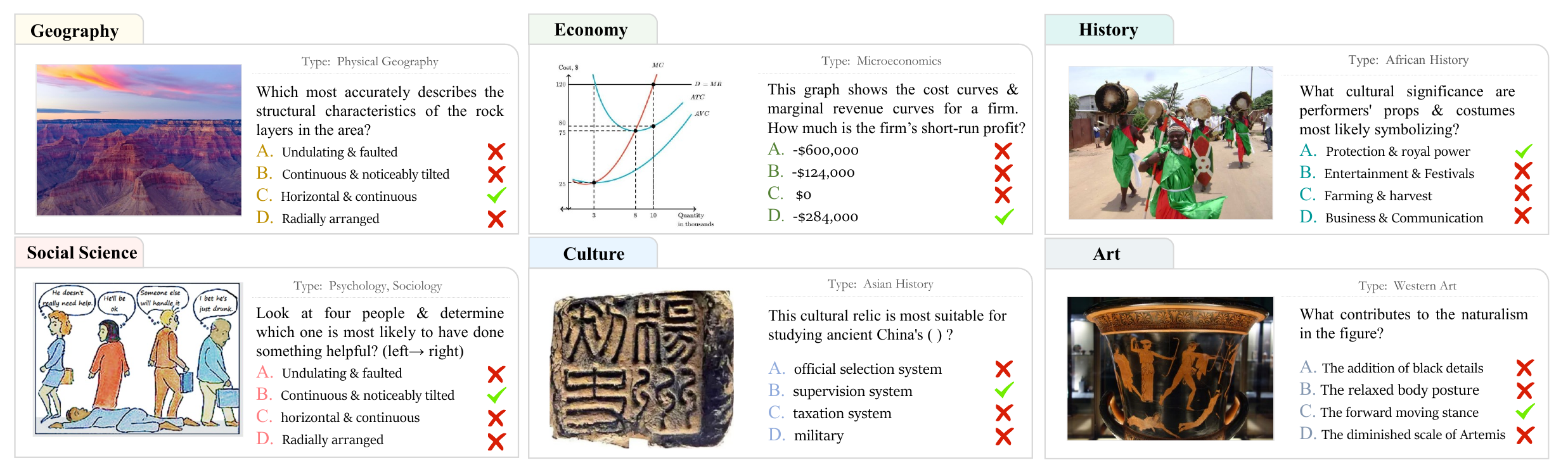}
    \vspace{-0.3cm}
    \caption{We propose HSSBench, a large-scale benchmark spanning six diverse categories and 45 types, comprising 13,152 samples collected in the six official languages of the United Nations.}
    \label{fig:pic1}
    
\end{figure*}
\begin{abstract}
Multimodal Large Language Models (MLLMs) have demonstrated significant potential to advance a broad range of domains.
However, current benchmarks for evaluating MLLMs primarily emphasize general knowledge and vertical step-by-step reasoning typical of STEM disciplines, while overlooking the distinct needs and potential of the Humanities and Social Sciences (HSS).
Tasks in the HSS domain require more horizontal, interdisciplinary thinking and a deep integration of knowledge across related fields, which presents unique challenges for MLLMs, particularly in linking abstract concepts with corresponding visual representations.
Addressing this gap, we present HSSBench, a dedicated benchmark designed to assess the capabilities of MLLMs on HSS tasks in multiple languages, including the six official languages of the United Nations.
We also introduce a novel data generation pipeline tailored for HSS scenarios, in which multiple domain experts and automated agents collaborate to generate and iteratively refine each sample.
HSSBench contains over 13,000 meticulously designed samples, covering six key categories.
We benchmark more than 20 mainstream MLLMs on HSSBench and demonstrate that it poses significant challenges even for state-of-the-art models.
We hope that this benchmark will inspire further research into enhancing the cross-disciplinary reasoning abilities of MLLMs, especially their capacity to internalize and connect knowledge across fields.
\end{abstract}

\footnotetext[1]{HSSBench is publicly available at: \url{https://github.com/Zhaolu-K/HSSBench}.}

\section{Introduction}
\begin{figure}[t]
    \centering
    \includegraphics[width=0.97\textwidth]{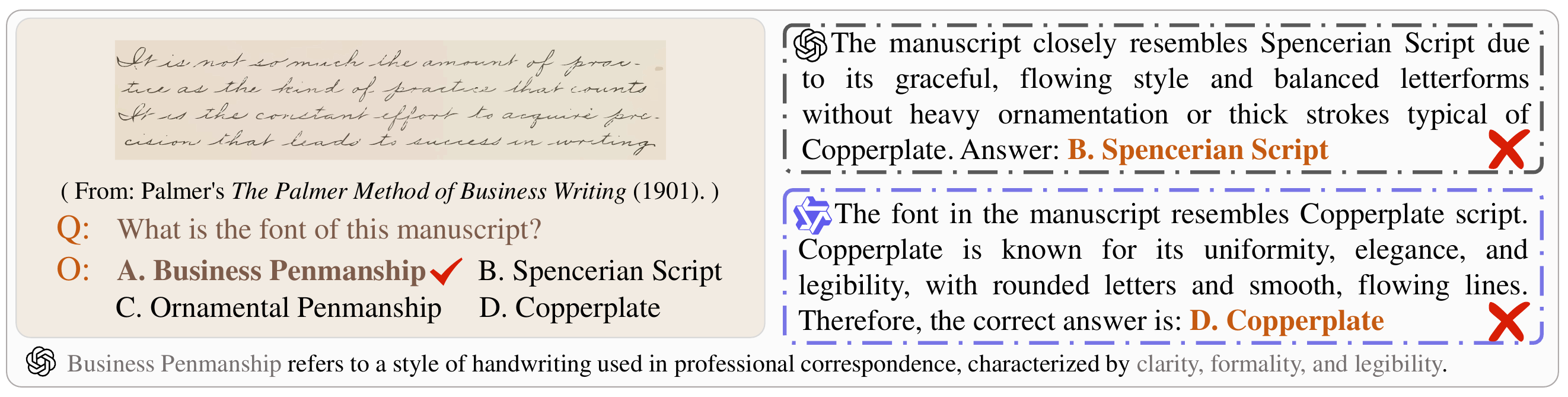}
    \caption{An example of cross-modal knowledge transfer issues in MLLMs within the HSS domain. They struggle to associate "Business Penmanship" font knowledge with relevant images or recognize fonts in image text.}
    \label{fig:introduction}
\end{figure}
Multimodal large language models (MLLMs)~\cite{achiam2023gpt,team2024gemini} have demonstrated remarkable performance across a wide range of tasks, recently achieving or even exceeding human-level capabilities in many of them.
As the performance of MLLMs continues to improve, conducting a comprehensive evaluation of their capabilities has become increasingly essential.
In recent times, numerous benchmarks for evaluating MLLMs have emerged~\cite{hendrycks2020measuring,yue2024mmmu,liu2024mmbench,saikh2022scienceqa,zhang2024cmmmu}.
These tasks are designed to assess the models' ability to jointly understand and reason across multiple modalities, such as images and text, from various perspectives.

Specifically, most multimodal benchmarks are designed either from a general perspective~\cite{liu2024mmbench,liu2023visualinstructiontuning,meng2024vgavisionguiassistant,han2024instinctivebiasspuriousimages,qian2024easyfoolmultimodalllms} or with a focus on scientific disciplines such as mathematics~\cite{wang2024measuringmultimodalmathematicalreasoning,lu2024mathvistaevaluatingmathematicalreasoning} science~\cite{li2023scigraphqalargescalesyntheticmultiturn,liang2024scemqascientificcollegeentrance}, and programming~\cite{song2025csbench}.
Unlike STEM fields that employ "vertical reasoning": a focused sequential process using logical deduction and experimental analysis to arrive at singular correct answers, the {\bf H}umanities and {\bf S}ocial {\bf S}ciences (HSS) emphasize "horizontal reasoning," requiring connections across different contexts and generating multiple valid interpretations rather than single solutions. This fundamental difference stems from the inherent attributes of these disciplines: While STEM fields utilize relatively fixed symbolic systems with standardized reading sequences, HSS disciplines feature symbol systems deeply rooted in regional cultures with meanings that require historical-cultural context interpretation. Furthermore, STEM knowledge can be iteratively developed through logical deduction and experimental analysis, whereas HSS knowledge verification relies on more complex pathways involving cross-referencing literature and expert consensus, with strong dependencies on real-world information.

Although some efforts have been made to explore aspects of the HSS, these attempts lack depth and do not provide a comprehensive and thoughtful examination of MLLMs within the context of these fields.
In the study of HSS-related problems, achieving cross-modal knowledge transfer is crucial. 
Take the scenario illustrated in Figure~\ref{fig:introduction} as an example: humans with basic knowledge can accurately infer "Business Penmanship" from the image content. When directly asked about knowledge points related to "Business Penmanship", the MLLM provides correct answers. However, when queried indirectly through an image, the model fails to recognize the font features in the text, preventing it from associating the relevant knowledge points with the image.
This reveals a problem: most models struggle to establish meaningful mapping relationships between HSS-related images and the abstract concepts they represent.
Although these models may recognize abstract concepts in isolation, they fail to effectively connect these concepts with the corresponding visual content.
From a long-term perspective, a model skilled at solving mathematical problems but unable to interpret historical contexts or understand ethical principles offers an incomplete and potentially harmful form of intelligence.

To address this challenge, we introduce HSSBench, an innovative and comprehensive multilingual benchmark specifically crafted to thoroughly assess the performance of MLLMs in the HSS domain.
Comprising around 13k carefully curated test items, HSSBench is structured into 45 types and covers 6 key categories within the field.
HSSBench utilizes the visual question answering (VQA) format for this purpose, as shown in Figure~\ref{fig:pic1}.
Given the involvement of numerous key domains, our work requires interdisciplinary collaboration.
To this end, we have engaged experts from various fields to design the data framework and ensure quality control, thereby maximizing the representativeness and credibility of HSS-related issues. 
In addition to domain experts contributing data, we leveraged the expertise of both specialists and MLLMs to develop a VQA generation pipeline, which produced a portion of the high-quality data. 
Finally, through evaluations supporting English, Chinese, French, Russian, Spanish, and Arabic, HSSBench enables the assessment of the capabilities of MLLMs in addressing HSS challenges across a wide range of linguistic contexts.

In this study, we evaluate the performance of HSSBench across a range of MLLMs and find that it presents a significant challenge for these models, as their accuracy often falls below 60\%.
We conducted several comparative experiments to analyze the performance of the models.
Our contributions can be summarized as follows:

\begin{itemize}
    \item First, we introduced HSSBench, a novel dataset specifically designed for the HSS domain, which encompasses 6 distinct categories and 45 major types of HSS tasks.
    \item Second, we offered a practical data construction method. It utilizes a multi-agent framework tailored for the HSS domain, allowing batch generation of high-quality, novel datasets.
    \item Finally, we conducted detailed evaluations of over 20 MLLMs on HSSBench across six languages, verifying that HSS tasks still pose significant challenges for MLLMs. This work establishes a foundation for future MLLM research focusing on HSS and serves as a benchmark for future studies in this field.
\end{itemize}

\begin{figure}[!t]
    \centering
    \includegraphics[width=1\linewidth]{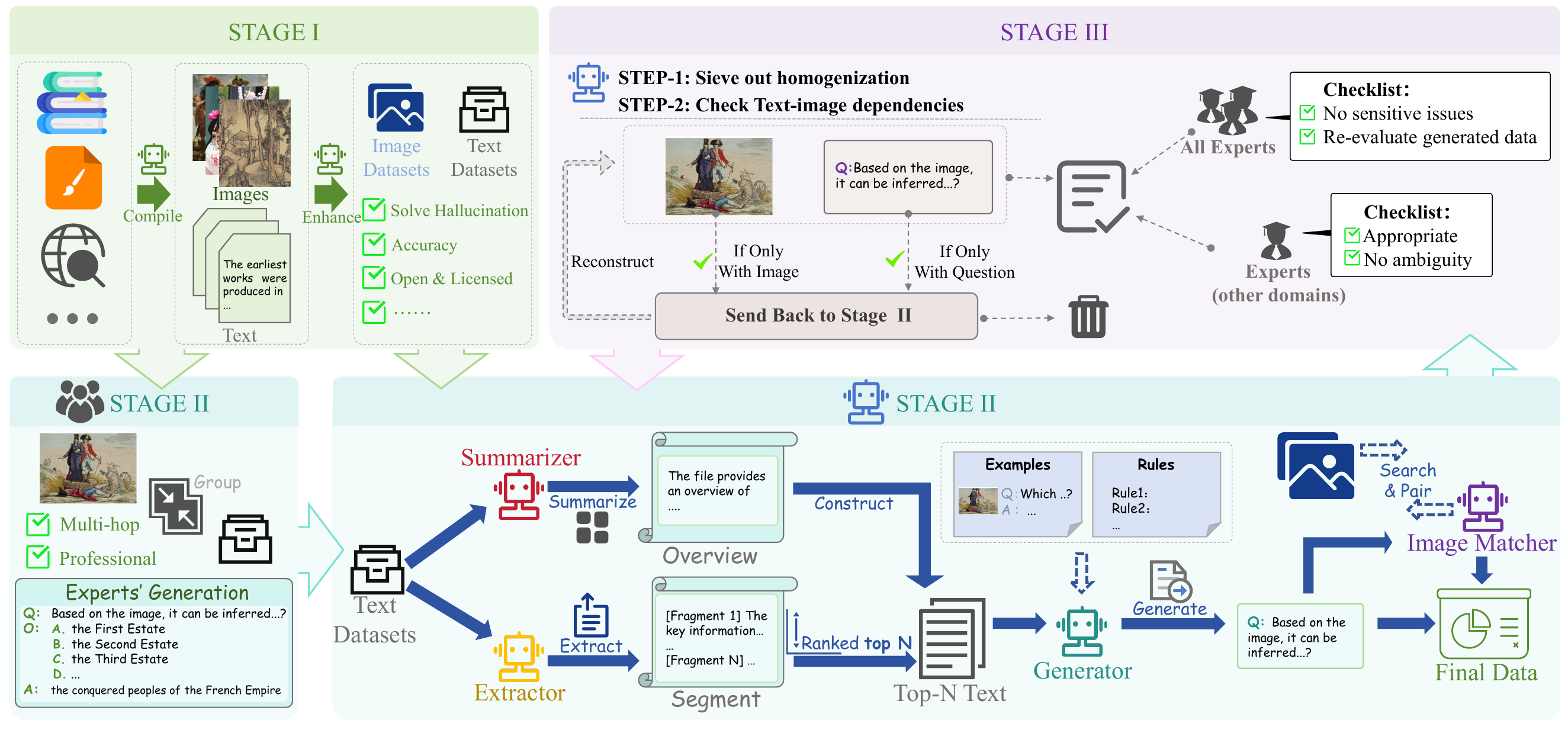}
    \caption{Our pipeline to build HSSBench.}
    \label{fig:image1-test}
    \vspace{-0.3cm}
\end{figure}
\section{HSSBench}

The worldview within the HSS is expansive and lacks unified definitions.
Our dataset focuses on six high-interest categories within the HSS domain: geography, art, culture, social sciences, history, and economics.
Textual data remain the primary medium for disseminating knowledge in the HSS domain, while pictorial data, though less abundant and more challenging to collect, provide valuable complementary information.
This presents significant challenges for experts in the construction of datasets.
For this purpose, we designed and developed a VQA Generation Pipeline (VGP) to generate the dataset.
After engaging experts for annotation, we designed a data construction agent based on their annotation logic, enabling us to generate a sufficient volume of data.

In this section, we will outline the details of the VGP. We have divided this process into three main stages: Dataset Preparation, Dataset Construction, and Validation.
The three stages are summarized in Figure~\ref{fig:image1-test}.
1) During the Dataset Preparation stage, both experts and a networked agent participate in obtaining the raw materials necessary for data construction. Either source is sufficient to provide the foundational content for the subsequent stage.
2) In the Dataset Construction stage, data can be constructed using two methods: expert-constructed and agents-constructed approach. 
3) Within the Validation stage, both experts and agent filtration work jointly to identify high-quality data.
The screened data passes through the second stage again until it meets the required standards.
Data that do not conform to the construction logic are removed.

\subsection{Dataset Preparation, Stage \Rmnum{1}}
\label{sec:Dataset Preparation}
At this stage, we focus on collecting data on target disciplines, covering texts and illustrations. Through systematic organization and strict screening, we lay a solid data foundation for the subsequent development of high-quality questions and answers.

\paragraph{Data Selection and Collection with Experts.}
During the data preparation for HSSBench, ensuring data diversity is crucial. For image and text selection, we adopt a combined approach involving experts and a multi-agent framework.

For images, to avoid data leakage, we initially encourage experts from various fields to use their private images. When acquisition is difficult, the dataset also incorporates images drawn by experts and licensed ones from open-source communities obtained by both experts and the multi-agent framework. Experts and agents collaborate to screen image content, choosing those related to subject areas and targeted knowledge points for constructing high-quality questions.

Regarding text, as experts may face challenges like unfamiliarity with domain-specific knowledge, we recommend they acquire high-quality textual materials from multiple sources, such as academic resource repositories of universities and open-source communities. This enriches information sources and mitigates bias. In the early stages, domain experts obtain credible resources like textbooks, past papers, and digital course materials from various disciplines. 
Since these resources are proofread during compilation, they possess high reliability and dense knowledge content, making them trustworthy sources for constructing questions.
Each expert reviews materials in their field, extracting relevant text passages and images, eliminating redundant information and standardizing the data format.
\paragraph{Networked Information-Aggregation Agent.}
Inspired by the expert-led data collection process, we designed the Networked Information-Aggregation Agent to mimic their workflow. This agent broadens data sources by filtering Internet data while maintaining quality. For each discipline, it first compiles relevant keywords as knowledge point indices. Then, it retrieves online data, classifies it into text and images, and matches data against the keywords to assess relevance. For text, it evaluates aspects like professionalism, uniqueness, logical structure, and the need for cross-validation with images. If the text meets criteria, related images are extracted. Through this process, we obtain high-quality multimodal data at a controlled cost. Finally, domain experts review the curated data to ensure professionalism and reliability.

\subsection{Dataset Construction, Stage \Rmnum{2}}
At this stage, we developed a multidisciplinary visual-question generation pipeline comprising two key phases: Experts Construction and Multi-Agent Construction.
These phases are designed to ensure that the questions produced offer a comprehensive assessment of the model's effectiveness.

\paragraph{Dataset Construction with experts.}

At this stage, experts have two main responsibilities: revising existing questions and creating entirely new ones.
1) When revising existing questions, experts optimize multiple-choice items, encompassing both the original ones provided in the materials and those requiring revision after failing validation tests.
Experts carefully review the relevant textual and visual content, adjusting the question stems and answer choices by integrating real-world knowledge.
The goal is to strengthen the connection between the questions and the accompanying images while improving the plausibility of the incorrect options.
2) For creating new questions, experts generate items based on the given high-quality texts and images.
These new questions are expected to demand identification of knowledge points from the images and complex reasoning based on real-world understanding to arrive at the correct answers.
Following these principles, experts produce a set of high-quality questions that fully utilize the provided multimodal materials to meet the demands of model performance evaluation.

\paragraph{Dataset Construction with Multi-Agent.}
Inspired by the expert annotation process and the high-quality questions it produces, we designed a multi-agent automated construction framework aimed at improving data generation efficiency and reducing human labor.
This agent architecture comprises several roles, including summarizer, extractor, question generator, and image matcher. 

Initially,  the summarizer and extractor independently analyze the document text to produce a comprehensive summary and a set of high-quality text segments, respectively.
The summary provides an overview of the key knowledge points within the document, while the extracted segments offer detailed information suitable for direct question formulation. LLM then scores these text segments based on information density, uniqueness, and logical coherence, selecting the top N segments according to their scores.

The question generator then formulates N questions by leveraging both the full summary and the selected text segments. It operates under detailed guidelines and is supported by multiple examples of human-authored questions, including stems, options, answers, and explanations, to ensure adherence to question design standards. 
Finally, the image matcher pairs images with questions by leveraging either direct image-question matching or matching based on image descriptions, depending on the available data. 
This Multi-Agent approach enables large-scale generation of high-quality question datasets efficiently and with minimal human intervention.

\subsection{Validation, Stage \Rmnum{3}}
This subsection outlines the stringent validation procedures implemented to ensure the quality and relevance of the data used in model evaluation.
Two critical types of validation are performed: Agent Validation and Expert Validation.

\paragraph{Agent Validation.}
This stage aims to eliminate duplicate questions and verify the strength of the correlation between the visual content and the corresponding text.
First, the Validation Agent calculates the textual similarity between questions and filters out highly redundant ones based on predefined criteria to ensure the diversity of the dataset.
Next, for all constructed data, it is necessary to ensure that 1) without providing an image, the question cannot be correctly answered based solely on the text, and 2) without providing a question, the image alone cannot lead to the correct answer.
This requirement stems from our aim to scrutinize the model's capabilities from a multimodal perspective.
If a single modality, text or images, suffices to answer a question, it compromises the quality of the dataset and its ability to comprehensively evaluate the performance of MLLMs.
To meet these two requirements, the Validation Agent assesses the extent to which the question depends on the image.
If the image is deemed unnecessary, the question is sent back to the Stage \Rmnum{2} for revision; if it still fails to meet the requirements after multiple iterations, the question is discarded.

\paragraph{Expert Validation.}

This stage focuses on expert evaluation of the data.
For data annotated by experts, each entry must be 1) validated by other domain experts to confirm its appropriateness and lack of ambiguity, and 2) confirmed by all experts to ensure that it is free from sensitive issues.
In the case of data generated by models, a rigorous evaluation by experts specializing in data generation is required to verify its accuracy and absence of ambiguity.

Furthermore, it requires the collective consent of all experts to confirm that the data are free of sensitive issues.

\subsection{Implementation Details}

\begin{wrapfigure}{r}{0.3\textwidth}
\vspace{-2cm}
\caption{Overview of HSSBench.}
\vspace{-0.4cm}
\includegraphics[width=1\linewidth]{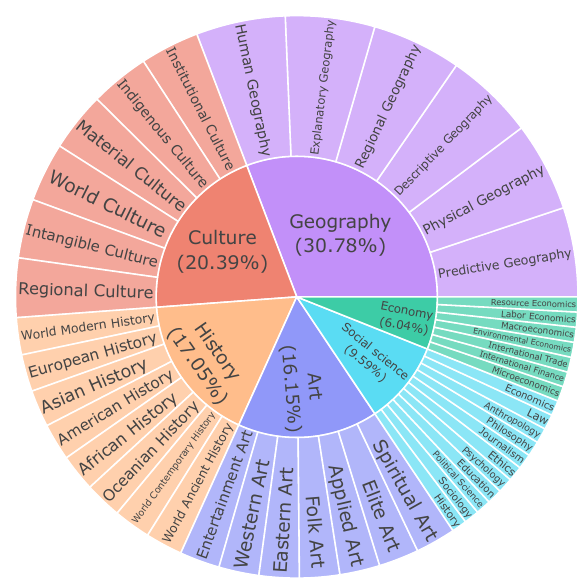} 
\vspace{-1.8cm}
\label{fig:wrapfig}
\end{wrapfigure}

In our VGP, we leverage GPT-4o and GPT-4.1 due to their outstanding capabilities.
Throughout all stages of LLM, we employed the Chain-of-Thought(COT) prompting strategy. 
For more detailed information on VGP, please refer to Appendix~\ref{sec.appendix.detail.VGP}.

\subsection{Data Statistics}

HSSBench consists of 13,152 multiple-choice questions distributed across six major categories—Economy, Art, Culture, Social Sciences, History, and Geography—and further divided into 45 specialized subtypes.

We require each question to be presented in a multiple-choice format with one correct answer and several distractors that are plausible but ultimately incorrect. Although some original questions in the HSS domain naturally have multiple correct answers, these multi-answer questions have been reformulated into single-answer questions for consistency. This was achieved by combining the multiple correct options into a single choice. This approach preserves the original content and complexity of the questions while conforming to a uniform single-answer multiple-choice format.

It supports evaluation in six different languages. Our initial data came from multiple countries and languages, with questions originally created by domain experts in their source languages (examples in Appendix~\ref{sec.appendix.case}). We then used LLM-based translation models to translate the questions into five other languages. All translations were carefully reviewed and validated by bilingual experts to ensure linguistic accuracy and cultural appropriateness, maintaining semantic consistency while respecting cultural nuances.

The statistical data of HSSBench is illustrated in Figure~\ref{fig:wrapfig}, with a detailed description of the dataset provided in Appendix~\ref{sec.appendix.detail.Dataset}.  Appendix~\ref{sec.appendix.humanPerformance} reports human expert accuracy across all categories. Appendix~\ref{sec.appendix.detailsType} provides an in-depth taxonomy of categories and subtypes. Appendix~\ref{sec.appendix.FrequentWords} also includes an analysis of the most frequent content words per category. Appendix~\ref{sec.appendix.Composition.Quality} details the composition of the dataset in terms of contributions of human experts versus automated agents, together with quality validation through model accuracy comparisons.

\section{Experiment}

\subsection{Experimental Setup}\label{expreimentalsettings}
\paragraph{Models.}
For open-source models, we selected Qwen2.5-VL-3B/7B/32B/72B~\cite{Qwen2.5-VL}, Qwen2-VL-72B~\cite{wang2024qwen2vlenhancingvisionlanguagemodels}, QVQ-72B-Preview~\cite{Qwen2-VL}, Deepseek-VL2-Tiny~\cite{wu2024deepseekvl2mixtureofexpertsvisionlanguagemodels}, MiniCPM-o-2.6~\cite{yao2024minicpm}, mPLUG-Owl3-2B/7B~\cite{ye2024mplugowl3longimagesequenceunderstanding}, Llava-onevision-7B~\cite{li2024llavaonevisioneasyvisualtask},  Llama3-llava-next-8b~\cite{li2024llavanext-strong}, InternVL3-8B~\cite{zhu2025internvl3exploringadvancedtraining}, InternVL2.5-8B-MPO~\cite{chen2024expanding}, Phi-3.5-Vision-Instruct~\cite{abdin2024phi3technicalreporthighly}, Janus-Pro~\cite{chen2025janusprounifiedmultimodalunderstanding}, Llava1.5~\cite{liu2024improvedbaselinesvisualinstruction}.
For closed-source commercial models, we utilized GPT-4o/4.1/4.1-mini/4.1-nano~\cite{openai2024gpt4ocard}.

\paragraph{Evaluation Settings.}
1) We conducted evaluations using two types of prompts to examine the MLLMs performance under different prompting strategies. One approach involved prompting the model to directly output the answer without any intermediate reasoning, while the other required the model to generate a COT before providing the final answer.
2) Additionally, we tested each question in six different language versions to investigate how the MLLMs' performance varies across languages for the same question.
3) Furthermore, we optimized the question formats by designing two types of prompts for each item: multiple-choice and open-ended questions. This allowed us to assess whether the model can correctly answer questions without any explicit hints. Detailed prompt settings are provided in Appendix~\ref{sec.appendix.detail.prompt}.
4) For the HSSBench evaluation, we employed two assessment methods and selected GPT-4o as a representative closed-source model and Qwen2.5-7B as a representative open-source model. A detailed comparative analysis of both models is provided in Appendix~\ref{sec.appendix.detail.validation}.

\subsection{Results}

\paragraph{Overall Results.}

Table~\ref{apptab:question_format_score} presents the overall results of various open-source and closed-source MLLMs evaluated on HSSBench in the context of English language tests. 
Ct. and Dt. represent two types of prompts: COT prompts and direct response prompts.
C. and O. denote question types: multiple-choice and open-ended questions. 
"Human" refers to the average final scores achieved by experts in various fields.
Detailed experimental results for Dt. and are presented in Appendix~\ref{sec.appendix.detail.results}.

All of our initial generated data consisted of multiple-choice questions. Since some questions cannot be reasonably converted to open-ended formats, changing all of them into open-ended questions would lead to poor-quality data. This approach would be unfair for evaluating the model and would fail to accurately reflect differences in model performance. Therefore, we chose to modify only those questions that can still be answered meaningfully as open-ended questions to serve as the evaluation data for open-ended question performance. 

The experimental results highlight the performance differences between the various models.
Of the open-source models, Qwen2.5-VL-72B-Instruct delivers the highest performance, although it still falls short of surpassing closed-source models in open-ended question tests.
The GPT-4.1 series achieves state-of-the-art performance in most tasks, with a particularly impressive accuracy of 39.97\% in open questions, almost double that of other closed-source models.
In contrast, some open-source models show considerably lower performance.
We also observe that model performance varies by language. Multilingual evaluation results are provided in the Appendix~\ref{sec.appendix.detail.results.languages}.

\begin{table}[!t]
    \centering
    \renewcommand{\arraystretch}{1.1}
    {\Large
\resizebox{\textwidth}{!}{
    \begin{tabular}{c|cc|cc|cc|cc|cc|cc|cccc}
    \toprule
    \multirow{2}{*}{Model} 
    & \multicolumn{2}{c|}{Geography} 
    & \multicolumn{2}{c|}{Economy} 
    & \multicolumn{2}{c|}{Culture} 
    & \multicolumn{2}{c|}{Social Sciences} 
    & \multicolumn{2}{c|}{History} 
    & \multicolumn{2}{c|}{Art} 
    & \multicolumn{4}{c}{All} \\
    \cmidrule(lr){2-3} \cmidrule(lr){4-5} \cmidrule(lr){6-7} \cmidrule(lr){8-9} \cmidrule(lr){10-11} \cmidrule(lr){12-13} \cmidrule(lr){14-17}
    & Ct.C. & Ct.O. & Ct.C. & Ct.O. & Ct.C. & Ct.O. & Ct.C. & Ct.O. & Ct.C. & Ct.O. & Ct.C. & Ct.O. & Dr.C. & Dr.O. & Ct.C. & Ct.O. \\ \midrule
        Random & 24.93 & 0.00 &  21.92 & 0.00 & 25.00 & 0.00 & 24.90 & 0.00 & 24.91 & 0.00 & 25.00 & 0.00 & 24.62 & 0.00 & 24.62 & 0.00 \\
        Human & 94.14 & - & 93.06 & - & 92.99 &-  &94.44 & - &  93.84& - & 95.53& -& 93.83 & - & 93.83&-\\
        \midrule
        \multicolumn{16}{c}{\textit{Open-source LLM (Scale $<$ 10B)}}\\ \midrule
        Qwen2.5-VL-3B-Instruct &  35.25 & 11.74 & 28.56 & 16.36 & 34.61 & 1.85 & 39.46 & 11.50 & 36.23 & 11.07 & 34.55 & 3.62 & 29.01 & 9.94 & 34.99 & 9.33\\
        Qwen2.5-VL-7B-Instruct & 40.69  & 21.60 & 41.19  & 31.31 & 30.19  & 4.63 & 42.86 & 19.00 & 40.62 & 19.67 & 33.42 & 11.31 & 37.88 & 15.21 & 38.19 & 17.89 \\
        Llava-onevision-7b & 32.05 & 7.04 & 32.23 & 11.68 & 31.02 & 1.85 & 36.63 & 3.00 & 27.07 & 4.51 & 32.92 & 3.17 & 36.20 & 5.73 & 31.56 & 5.20 \\ 
        Llama3-llava-next-8b & 27.59 & 4.23 & 19.82 & 6.54 & 30.53 & 3.24 & 32.89 & 6.50 & 26.87 & 8.61 & 29.26 & 5.43 & 31.20 & 6.50 & 27.93 & 5.81\\ 
        InternVL3-8B & 42.12 & 10.80 & 33.70 & 16.36 & 38.69 & 7.41 & 48.30 & 13.00 & 45.09 & 12.70 & 38.61 & 13.57 & 42.14 &  12.27 & 41.42 & 12.31\\ 
        InternVL2.5-8B-MPO & 37.24 & 17.37 & 34.07 & 20.09 & 35.00 & 8.33 & 43.35 & 15.50 & 40.54 & 16.39 & 35.99 & 13.57 & 39.30 & 11.77 & 37.68 & 15.21\\ 
        Phi-3.5-Vision-Instruct & 25.55 & 9.39 & 26.80 & 13.55 & 29.01 & 3.70 & 28.78 & 3.00 & 20.31 & 7.79 & 28.81 & 4.07 & 35.89 & 10.32 & 26.04 & 6.96\\
        Janus-Pro  & 29.11 & 8.45 & 22.54 & 6.07 & 41.00 & 6.02 & 34.65 & 12.00 & 29.75 & 10.25 & 33.22 & 7.69 & 30.03 & 8.49 & 31.66 & 8.41 \\
        Deepseek-VL2-Tiny &5.78 & 3.29&3.77 &4.95& 14.12&0.00&6.35  & 3.00&  6.37& 2.46& 13.03 &16.74 &29.86  & 3.42& 8.23 & 5.09 \\

        mPLUG-Owl3-2B  & 25.83 & 4.69 &24.63 &  1.35&  31.83&  3.69 & 29.99& 3.00 &  25.96 & 4.51 &29.43 &4.07 & 28.73 &4.02 & 27.71&3.57\\
        mPLUG-Owl3-7B  & 30.60 & 7.04 & 33.15 & 9.91 & 13.73 & 2.30 & 36.40& 7.00 &27.64 &   4.10& 25.69 & 2.30&  33.01& 6.68 & 27.52 & 6.23\\
        MiniCPM-o-2.6  & 26.02 & 5.98 &  19.08&  7.22& 22.22 & 3.57  &26.83 & 8.59 & 25.96 & 4.03 & 22.25& 5.38  & 3.70 & 5.71 &24.11 &5.71\\
        Llava1.5 
         & 12.75 & 4.03 &7.38 & 2.68 & 10.74 & 3.36 & 7.38&  5.37 &10.74  & 4.03& 10.74 &6.04& 8.06&  4.03 & 9.96&4.25\\
        \midrule
        \multicolumn{16}{c}{\textit{Open-source LLM (Scale $>$ 10B)}}\\ \midrule
        Qwen2.5-VL-32B-Instruct  &\highgreen{\textbf{ 52.48}} & 21.33&  52.79 & 6.67  & 38.94 &8.00 &\highgreen{\textbf{53.87}} &24.00&\highgreen{\textbf{  57.03}} &26.67 & 39.20&  3.33& 48.38 & 15.89&\highgreen{\textbf{  50.75}}&  15.00\\
        Qwen2-VL-72B-Instruct  & 50.74 & 17.86 &  52.55&  32.65& 45.91 &8.62   &50.19 & 16.07&  55.53&  16.36& 41.34  &  16.36& \high{54.22} & 20.43 &49.39 &17.21\\
        
        Qwen2.5-VL-72B-Instruct  & \high{\textbf{ 55.59}} & 13.33 &\highgreen{\textbf{  53.83}} & 37.33 &  41.49 &  7.33&\high{\textbf{  57.77}}&  17.57& \high{\textbf{60.30}} & 28.19 &40.84& 14.67 & \highgreen{\textbf{ 54.17 }} &18.17 &\high{\textbf{51.87}}  & 19.73\\
        QVQ-72B-Preview  & 19.93 &3.33  &21.87  & 17.33 & 29.67 & 3.33 &26.54  & 7.43 &  28.86& 18.79 & 23.80  & 9.33& 25.60 & 10.37 &24.69 &9.92\\ 
        
        \midrule
        \multicolumn{16}{c}{\textit{Closed-source LLM}}\\ \midrule
        GPT-4o  & 46.88 & 22.07 & 52.97 & 35.14 &  45.61 & 14.29 & 45.26&16.00  &48.36   & 15.98& 43.42& 12.67 &46.09&20.05 &46.88 &19.36\\
        GPT-4.1  & 39.81 & \high{\textbf{40.38}} & 48.08 & \high{\textbf{52.70}} &  \high{\textbf{48.95}}&\high{\textbf{24.88}}  & 35.36& \high{\textbf{49.25}} &  41.91&\high{\textbf{48.77}} &\highgreen{\textbf{43.51}}& \high{\textbf{23.53}}& 45.02 &\high{\textbf{25.38}} &42.66 &\high{\textbf{39.97}}\\
        GPT-4.1-mini   & 47.67 &   \highgreen{\textbf{ 34.27}}&  \high{\textbf{58.27}} &  \highgreen{\textbf{49.10}}& \highgreen{\textbf{ 48.26}} & \highgreen{\textbf{ 20.74}}& 45.05&\highgreen{\textbf{ 36.68}} & 47.84 & \highgreen{\textbf{ 36.89 }}& \high{\textbf{43.88}}&\high{\textbf{23.53}} &45.75  &\highgreen{\textbf{  24.32}} &48.03 &\highgreen{\textbf{ 33.59}}\\
        GPT-4.1-nano  & 33.21 &30.74  & 39.71 & 41.44 &  38.70& 7.83  &37.52 & 28.14   &  34.01 &30.74  & 35.83& 20.36&  36.33 &21.12  &35.83 &26.22\\ 
        \bottomrule
    \end{tabular}
    }}\vspace{0.07cm}
    \caption{Scores (\%) of MLLMs on HSSBench (EN-\Rmnum{1}). The highest and second highest scores are marked in \textcolor{blue}{blue} and \textcolor[RGB]{0,119,51}{green}, respectively.}
    \label{apptab:question_format_score}
    \vspace{-0.7cm}
\end{table}

\paragraph{Model performance on different categories.}

Among the six categories evaluated on HSSBench (EN-\Rmnum{1}), the economic-related tasks consistently emerge as the most challenging for the models. 
The average score of all models in this category is the lowest, indicating that addressing economic problems requires a deep understanding of various economic theories and the ability to apply them in complex reasoning. 
However, current open-source MLLMs still exhibit significant deficiencies in these aspects. In contrast, closed-source models perform exceptionally well in economically related tasks, and this advantage may stem from their exposure to large amounts of high-quality training data in the economic domain.

On the other hand, the Geography category appears to be the easiest task for the models, with the highest average scores observed across the board. This trend implies that geographic knowledge, which is often more factual and less abstract compared to other humanities and social sciences domains, is better captured by the training data and reasoning capabilities of the models. 

Interestingly, in certain categories of multiple-choice tasks, such as Culture and Social Sciences, some open-source models outperform their closed-source counterparts. For example, larger open-source models like Qwen2.5-VL-32B and Qwen2.5-VL-72B demonstrate competitive or even superior results compared to closed-source models like GPT-4o in these domains.
This phenomenon may be partially attributed to the fact that most of the data experts we invited are Chinese.
Although the evaluation was conducted in English, it is possible that Qwen benefits from a training data advantage related to Chinese content, which could have influenced its superior performance in these tasks.
It also indicates that the gap between open-source and closed-source models is narrowing in specific HSS tasks.

\subsection{Qualitative Analysis}

\paragraph{Comparison between Direct Answer and COT Prompting.}  
Table~\ref{apptab:question_format_score} shows subtle differences in MLLMs’ performance on HSSBench when using direct answer versus COT prompting. Notably, COT does not always help; some models perform better with direct answers, indicating that longer reasoning can mislead them.

Specifically, COT prompts exacerbate hallucination issues in certain models, where reasoning flaws in textual analysis and misinterpretation of visual inputs lead to the generation of incorrect background knowledge during step-by-step analysis. This causes reasoning to deviate from the correct answer. For example, when tackling geographic questions, models often struggle to accurately interpret spatial elements such as location markers and contour lines—visual features primarily composed of points and lines—resulting in analytical outcomes that contradict the actual image content.

Even when notable errors do not occur in intermediate reasoning steps, the final summarization phase can suffer due to excessive information generation, which exceeds the model's ability to effectively weigh the importance of answer options, leading to prediction mistakes. This highlights the importance of integrating reasoning steps cohesively. Further detailed analysis in Appendix~\ref{sec.appendix.case} reveals that many models fail to internalize visual knowledge during the divergent thinking process of HSS tasks.

\paragraph{Comparison between Multiple-choices and Open-ended.}  
We reformulated some questions as open-ended to test models without answer options. Results reveal that HSS tasks remain very challenging: only a few models exceed 15\% accuracy. This matches expectations, as even experts find these questions difficult without options or background cues.

Answer choices provide prior knowledge that narrows the answer space. Without them, models’ reasoning becomes highly divergent, often drifting far from correct answers. This suggests models mainly rely on shallow visual features (e.g., size, type, motion) but miss deeper symbolic information like cultural context or spatiotemporal cues.

These findings highlight two key limitations: insufficient real-world knowledge and weak integration of visual and textual modalities. The performance drop with COT further reflects models’ struggles to internalize and retrieve complex knowledge. Improving model capabilities on HSS tasks remains a critical challenge.

\paragraph{The role of accurate answers.}  
To test if models truly understand questions rather than exploiting options, we added confusing choices like “None of the above” while keeping the correct answer unchanged, sampling 150 questions per category (results in Table~\ref{apptab:question_contrast2}, Appendix~\ref{sec.appendix.confounding}).

Lower-performing models’ accuracy dropped noticeably. Analysis revealed two patterns: some models were misled to select the confusing option; others, though not choosing it, showed incoherent reasoning, failing to filter out wrong choices and sometimes excluding the correct answer altogether. This indicates models struggle to assess the relative credibility of options and make reliable judgments, exposing their vulnerability to distractors and raising concerns about robustness.

\paragraph{Visual Information Extraction.}  
To investigate the impact of visual information loss on the performance of MLLMs, we designed two complementary experiments. In the first experiment, we used GPT-4.1 to generate detailed textual descriptions of the images. In the second experiment, we invited domain experts to produce comprehensive and precise annotations for each image (results in Table~\ref{apptab:question_contrast}, Appendix~\ref{sec.appendix.visual_extraction}).

Most models’ accuracy dropped when images were replaced by GPT-generated texts, showing that direct visual input contains critical details lost in conversion. However, with expert annotations, accuracy improved noticeably despite no image access. Some models even surpassed previous performance ceilings in Culture and Social Sciences, as expert texts helped focus on crucial visual cues and better link to domain knowledge.

These results confirm earlier observations that current MLLMs have inherent limitations in retrieving and understanding visual information fully. They underscore the importance of improving models’ abilities to extract and integrate key visual features for HSS tasks.

\paragraph{Comparative Analysis of HSSBench with Related Benchmarks.}  
We compared HSSBench with other benchmarks that include some HSS data, such as CMMMU, MME, and MMMU, as well as with STEM benchmarks (details in Appendix~\ref{sec.appendix.related.Benchmark} and \ref{sec.appendix.STEM.Benchmark}).

Across HSS benchmarks, relative model performance trends are consistent, but HSSBench yields lower accuracy, indicating it is more challenging and better captures HSS complexity. Compared to STEM benchmarks, models perform much better on STEM tasks, highlighting the unique difficulties of HSS domains that require nuanced cultural and social understanding. These comparisons demonstrate HSSBench’s value in driving progress on underexplored HSS challenges.

\paragraph{Evaluation of Retrieval-Augmented Generation on HSSBench.}  
We tested Retrieval-Augmented Generation (RAG) by integrating external knowledge from Wikipedia and HSS documents with several smaller MLLMs under direct and COT prompting (results in Tables~\ref{tab:direct_prompting} and \ref{tab:COT_prompting}, Appendix~\ref{sec.appendix.detail.RAG}).

Contrary to expectations, RAG did not consistently improve accuracy. Some modest gains appeared in specific domains or models, but overall performance was often similar or worse than direct prompting alone. This suggests that general retrieval corpora and simple integration methods are insufficient for the nuanced knowledge HSS tasks demand.

The limited effectiveness of RAG in supporting MLLMs for complex multi-hop HSS tasks stems from misalignment between general-purpose corpora and HSS knowledge, as well as deficiencies in current retrieval and integration methods. Moreover, MLLMs struggle to internalize and transfer newly retrieved domain knowledge. These challenges highlight the need for specialized HSS retrieval resources, RAG techniques tailored to HSS, and enhanced mechanisms for MLLMs to acquire, transfer, and apply HSS-specific knowledge.

\section{Related Work}

In recent years, significant progress has been made in the development of multimodal benchmarks and methodologies.
Numerous datasets have emerged that assess models from various perspectives, which can be broadly categorized into Generation Benchmarks~\cite{liu2024mmbench,liu2023visualinstructiontuning,meng2024vgavisionguiassistant,han2024instinctivebiasspuriousimages,qian2024easyfoolmultimodalllms,xu2023lvlmehubcomprehensiveevaluationbenchmark,yin2023lammlanguageassistedmultimodalinstructiontuning,zeng2023matterstraininggpt4stylelanguage,wu2024qbenchbenchmarkgeneralpurposefoundation,luo-etal-2024-codis,feng2025seeing,wang2025mucar}, Reasoning Benchmarks~\cite{wang2024measuringmultimodalmathematicalreasoning,lu2024mathvistaevaluatingmathematicalreasoning,li2023scigraphqalargescalesyntheticmultiturn,liang2024scemqascientificcollegeentrance,song2025csbench}, and Application Benchmarks~\cite{fan2022minedojobuildingopenendedembodied,rawles2023androidwildlargescaledataset,chen2023endtoendembodieddecisionmaking,sermanet2023robovqamultimodallonghorizonreasoning,chen2024egoplanbenchbenchmarkingmultimodallarge,koh2024visualwebarenaevaluatingmultimodalagents,you2024ferretuigroundedmobileui,gao2025laobench,kang2026quanteval,zheng2026should,he2026order}.
Generation benchmarks cover a wide range of tasks, often including not only reasoning challenges but also content related to the humanities and social sciences.
Their main objective is to provide a comprehensive assessment of the performance of MLLMs in various dimensions.
In contrast, reasoning benchmarks are designed primarily around mathematical and scientific problems, seeking to rigorously assess the capabilities of MLLM in logical and analytical reasoning.
Among the various evaluation formats, multiple-choice datasets are the most prevalent, owing to their simplicity in evaluation and ease of comparison~\cite{zeng2023matterstraininggpt4stylelanguage}.

The proliferation of these data sets has significantly accelerated research progress, serving both as training resources and as tools for assessing the multifaceted competencies of MLLMs~\cite{wang2025technical,liu2025training,luo2026unveilingcognitivecompasstheoryofmindguided,guan2026teaching,fu2026advsyngnn,ML-TTA,PVP,feng2026noisy,li2026comprehensive,li2025sepprune,li2025frequency,hucontext,zhang2025perl,zhang2025see,zhang2025pixelcraft,ding2025videozoomer,li2025cama,wang2026visual,ma2026thinkingblueprintsassistingvisionlanguage,liu2025explainable,bi2026prismselfpruningintrinsicselection,wang2025emergent,wang2025pixel,wang2025reverse,wang2025vl,liang2025pixelvla,ma2025generativeregressionbasedwatch,ma2025ms,li2025chemvlm,yu2025vismem,feng2024tasl,feng2023towards,li2021image,li2024unionformer,li2025toward}.
Advances in data set construction are particularly exciting, ranging from human-annotated high-quality datasets~\cite{romero2024cvqaculturallydiversemultilingualvisual} to those generated through pipelines built in LLM~\cite{chandrasegaran2024hourvideo1hourvideolanguageunderstanding}.
Despite these achievements, several critical limitations remain.
1) Most multimodal datasets cover a wide range of categories, but suffer from issues such as limited data sources, relatively simple questions, and insufficient image information.
2) Many large-scale datasets are collected through web crawling without thorough manual annotation and verification, which can introduce biases in evaluation results.
3) Current reasoning datasets predominantly focus on STEM tasks, relying heavily on scientific and mathematical data to assess reasoning capabilities. 
In contrast, the benchmark we propose is created through interactions among expert agents, targeting the HSS domain.
In contrast, the benchmark we propose focuses on the HSS domain, emphasizing tasks that analyze and understand the abstract concepts embedded in images.
This approach fills a notable gap in the field and lays the foundation for advancing model performance in this underexplored area.

\section{Conclusion}

We present HSSBench, a novel benchmark dataset constructed through a multi-agent pipeline involving experts from diverse fields.
HSSBench is designed to rigorously evaluate the true mastery of tasks by models within the HSS domain. The dataset comprises six categories, each derived from raw data collected from repositories in the six official languages of the United Nations and subsequently processed to generate task-specific data.

We then carried out comprehensive benchmarking of various MLLMs using HSSBench.
Our results reveal that HSSBench poses significant challenges to all tested models, which exhibit poor performance on reasoning tasks in the HSS domain.
In particular, the accuracy of the model decreases substantially when answer choices are not provided as prompts.
We hope that releasing HSSBench will encourage the AI community to place greater emphasis on reasoning over non-STEM data, thereby advancing research on MLLMs from this important perspective.
\bibliography{iclr2026_conference}
\bibliographystyle{iclr2026_conference}

\clearpage

\appendix

\section{More VGP Details} \label{sec.appendix.detail.VGP}
\subsection{Annotation Platform}

\begin{figure}[htbp]
    \centering
    \includegraphics[width=1\linewidth]{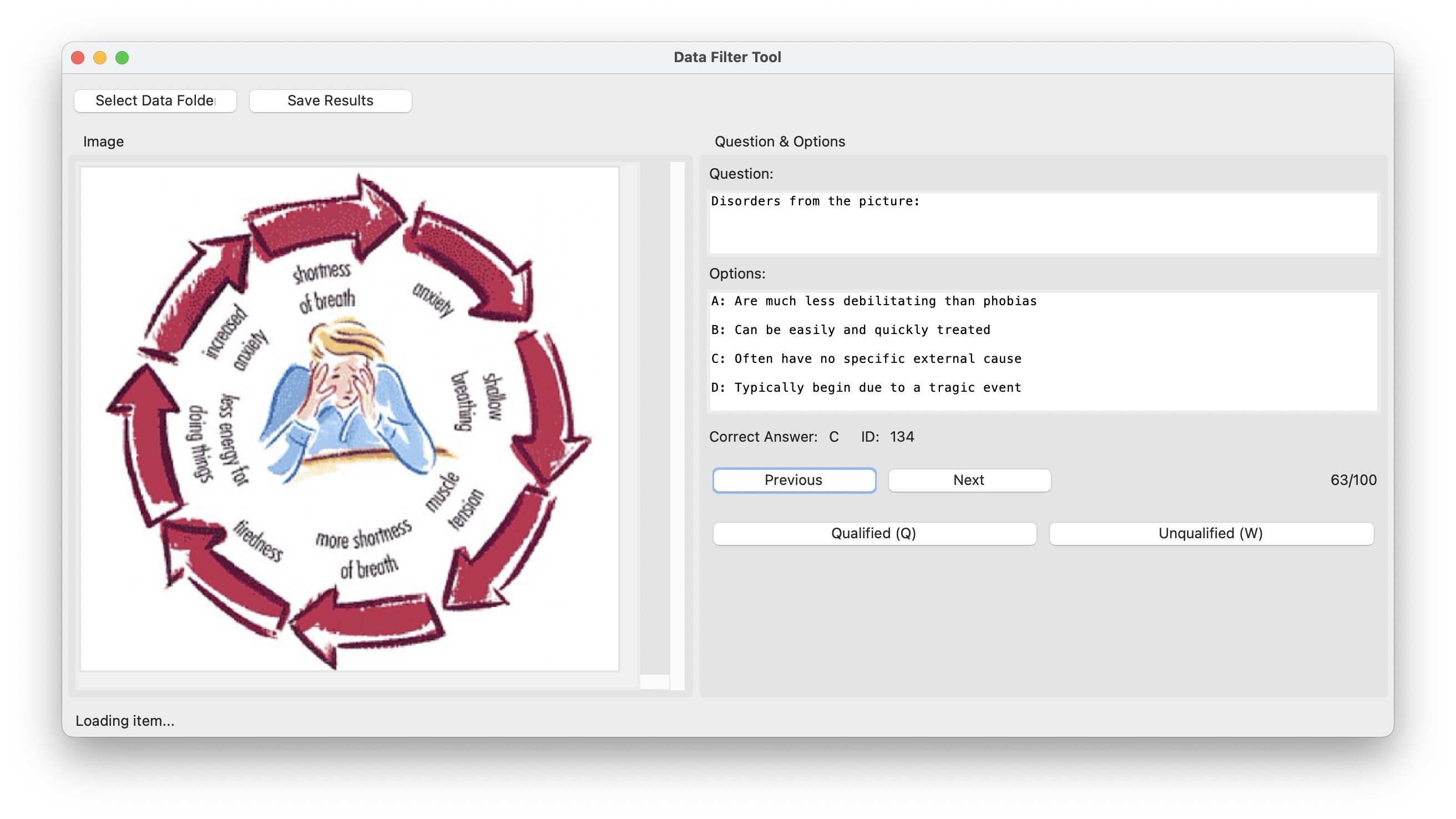}
    \caption{The interface for data construction and validation, allowing experts to use a visual interface to assist their work.}
    \label{fig:platform}
\end{figure}
We use a simple visual interface as our annotation platform.
For question input, experts can upload or write questions in the form.
The interface is shown in Figure~\ref{fig:platform}. During the validation process, experts can see all the data submitted by other experts. They can select entries to view detailed previews of the entries.

\subsection{Information about Experts}
\begin{table}[htbp]
  \centering
  \begin{tabular}{lll}
    \toprule
    \textbf{Name} & \textbf{Languages} & \textbf{Professional Background} \\
    \midrule
    Expert 1  & Chinese, English           & Computer Science, Art, Economy \\
    Expert 2  & Chinese, Japanese, English & Computer Science, Culture \\
    Expert 3  & English, Chinese           & Computer Science, Geography \\
    Expert 4  & Chinese, English           & Economy, History \\
    Expert 5  & Chinese, English, Japanese & Language and Linguistics, History, Culture \\
    Expert 6  & Chinese, English, Russian  & Language and Linguistics, Social Science \\
    Expert 7  & Chinese, English, Arabic   & Language and Linguistics, Social Science \\
    Expert 8  & Chinese, English, Arabic   & Economy, Social Science \\
    Expert 9  & Chinese, English           & Geography \\
    Expert 10 & Chinese, English, French   & Economy, Art \\
    Expert 11 & English, Chinese, Japanese & Art, Culture \\
    Expert 12 & Chinese, English           & Computer Science, Geography \\
    Expert 13 & Chinese, English           & Geography, History \\
    Expert 14 & Chinese, English, Spanish  & Computer Science, Geography \\
    \bottomrule
  \end{tabular}
  \caption{Information about the experts involved in dataset construction.}
  \label{experts}
\end{table}
The team of experts who contributed to the creation of HSSBench comes from a variety of cultural and linguistic backgrounds, as well as interdisciplinary academic fields spanning the humanities and social sciences. This diversity is reflected in their linguistic fluency, professional expertise, and international academic experiences. Table~\ref{experts} summarizes the profiles of the experts involved in the dataset construction.

Many of the experts have international academic experiences and interdisciplinary training across various humanities and social science domains. While individual cultural perspectives may naturally influence the emphasis or framing of certain questions, we consider this diversity a strength rather than a limitation. It enriches the dataset by incorporating multiple viewpoints and insights, which is essential for a benchmark designed to be multilingual and cross-cultural.

To mitigate potential cultural bias, we carefully curated the source materials provided to the experts during the initial stage of dataset construction. These reference materials originate from multiple countries and languages, ensuring that the constructed questions and answers reflect widely accepted, global knowledge rather than culturally subjective viewpoints. Furthermore, during the data validation and filtering stages, we rigorously excluded content that could be culturally biased or inconsistent with universal values.

In summary, our approach balances the preservation of valuable cultural diversity with the need to maintain fairness and universality in the dataset. This careful design ensures that HSSBench serves as a robust and inclusive benchmark for evaluating multilingual large language models (MLLMs) across different languages and cultural contexts.

\section{Dataset Details}\label{sec.appendix.detail.Dataset}
This section provides detailed information about our benchmark designed to evaluate MLLMs' visual comprehension abilities through multiple-choice questions.
The benchmark spans six major categories: Economy, Art, Culture, Social sciences, History, and Geography, each containing various specialized subtypes.

\subsection{Human Performance}\label{sec.appendix.humanPerformance}
To establish a performance baseline, we asked three relevant experts in each domain to spend a significant amount of time answering the entire dataset.
Table~\ref{expert_performance} summarizes their performance across the six categories.

\begin{table}[htbp]
    \centering
    \resizebox{0.6\textwidth}{!}{ 
    \begin{tabular}{c|ccc|c}
    \toprule
    Category & Expert 1 & Expert 2 & Expert 3 & Overall \\
    \midrule
    Economy       & 96.02 & 91.70 & 94.70 & 94.14 \\
    Art           & 93.12 & 92.61 & 93.44 & 93.06 \\
    Culture       & 92.50 & 90.83 & 95.64 & 92.99 \\
    Social sciences& 94.49 & 91.77 & 97.07 & 94.44 \\
    History       & 95.91 & 91.86 & 93.75 & 93.84 \\
    Geography     & 94.99 & 92.78 & 95.81 & 94.53 \\
    \midrule
    Average       & 94.84 & 91.76 & 95.40 & 93.83 \\
    \bottomrule
    \end{tabular}
    }
    \vspace{0.3cm}
    \caption{Scores (\%) of Experts.}
    \label{expert_performance}
\end{table}

Human experts demonstrated high proficiency across all categories, with overall accuracy ranging from 92.99\% to 94.53\%. The highest individual performance was observed in Social sciences by Expert 3 (97.07\%), while the lowest was in Culture by Expert 2 (90.83\%). This high level of human performance establishes a challenging benchmark for evaluating large language models.

\subsection{The details of types}\label{sec.appendix.detailsType}

Our dataset is organized into six major categories with various subtypes in each category. Table~\ref{tab:types} presents the detailed breakdown of the dataset structure and the count of questions in each type.
We allow each data entry to have multiple types because the intersection of knowledge across disciplines is essential.

The dataset exhibits varying distributions across categories, with Geography containing the largest number of questions and Economy the smallest. Within categories, there are also significant variations in subtype representation, reflecting the natural distribution of content within these domains.

\begin{table}[htbp]
    \centering
    
    \renewcommand{\arraystretch}{1.1} 
    \setlength{\tabcolsep}{12pt} 
    \begin{tabular}{@{} l l r @{}}
        \toprule
        \textbf{Category} & \textbf{Type} & \textbf{Count} \\
        \midrule
        \multirow{7}{*}{Economy} 
            & Microeconomics & 1,193 \\
            & Macroeconomics & 163 \\
            & Labor Economics & 105 \\
            & Environmental Economics & 63 \\
            & International Trade & 63 \\
            & Resource Economics & 44 \\
            & International Finance & 32 \\
        \midrule
        \multirow{7}{*}{Art} 
            & Folk Art & 1,051 \\
            & Eastern Art & 1,041 \\
            & Western Art & 672 \\
            & Spiritual Art & 630 \\
            & Applied Art & 452 \\
            & Entertainment Art & 298 \\
            & Elite Art & 303 \\     
        \midrule
        \multirow{6}{*}{Culture} 
            & Regional Culture & 1,790 \\
            & Material Culture & 1,382 \\
            & Intangible Culture & 1,218 \\
            & Indigenous Culture & 733 \\
            & Institutional Culture & 266 \\
            & World Culture & 228 \\   
        \midrule
        \multirow{12}{*}{Social sciences} 
            & Sociology & 950 \\
            & Political Science & 550 \\
            & Psychology & 339 \\
            & Anthropology & 190 \\
            & Economics & 147 \\
            & Education & 135 \\
            & Philosophy & 127 \\
            & Law & 99 \\
            & Ethics & 61 \\
            & Journalism & 24 \\       
            & History & 18 \\
        \midrule
        \multirow{8}{*}{History} 
            & Asian History & 1,586 \\
            & World Modern History & 885 \\
            & World Ancient History & 754 \\
            & European History & 602 \\
            & World Contemporary History & 346 \\
            & American History & 295 \\
            & African History & 147 \\
            & Oceanian History & 82 \\
        \midrule
        \multirow{6}{*}{Geography} 
            & Physical Geography & 2,361 \\
            & Regional Geography & 1,694 \\
            & Descriptive Geography & 1,682 \\
            & Explanatory Geography & 1,348 \\
            & Human Geography & 1,319 \\
            & Predictive Geography & 75 \\
        \bottomrule
    \end{tabular}
    \vspace{0.2cm}
    \caption{Dataset Taxonomy and Question Distribution.}
    \label{tab:types}
\end{table}

\subsection{Most-Frequent Words in the Questions}\label{sec.appendix.FrequentWords}
\begin{figure}[htbp]
 \centering
 \begin{subfigure}[b]{0.32\textwidth}
 \centering
 \includegraphics[width=\linewidth]{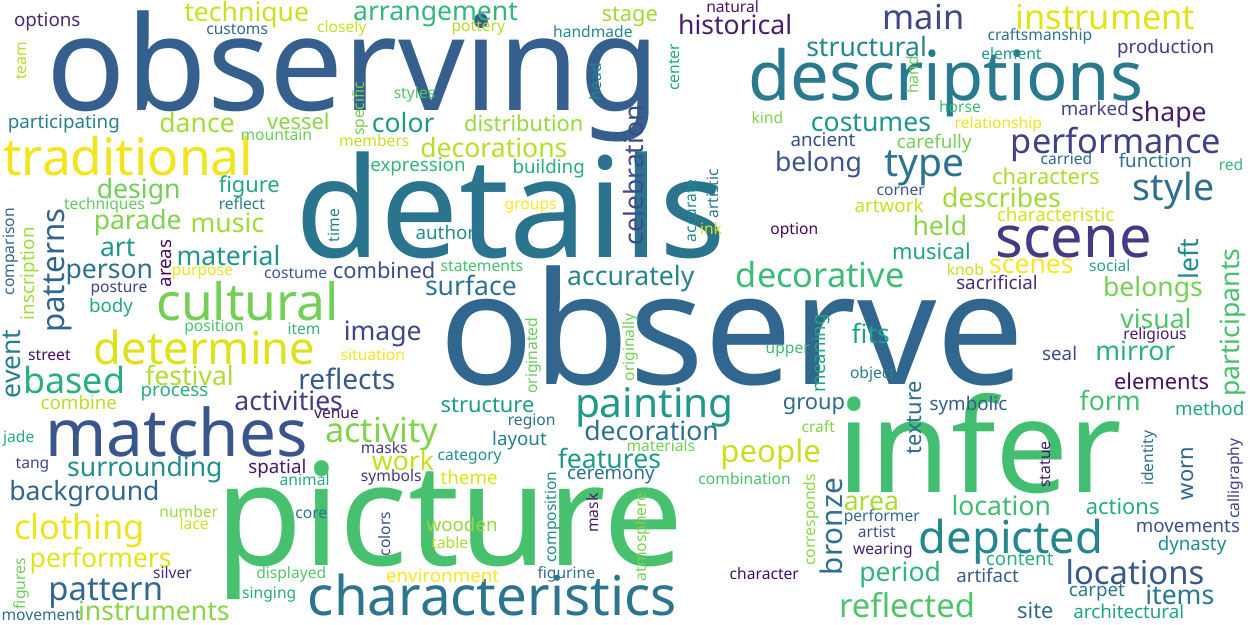}
 \caption{Art}
 \label{fig:1}
 \end{subfigure}
 \hfill
 \begin{subfigure}[b]{0.32\textwidth}
 \centering
 \includegraphics[width=\linewidth]{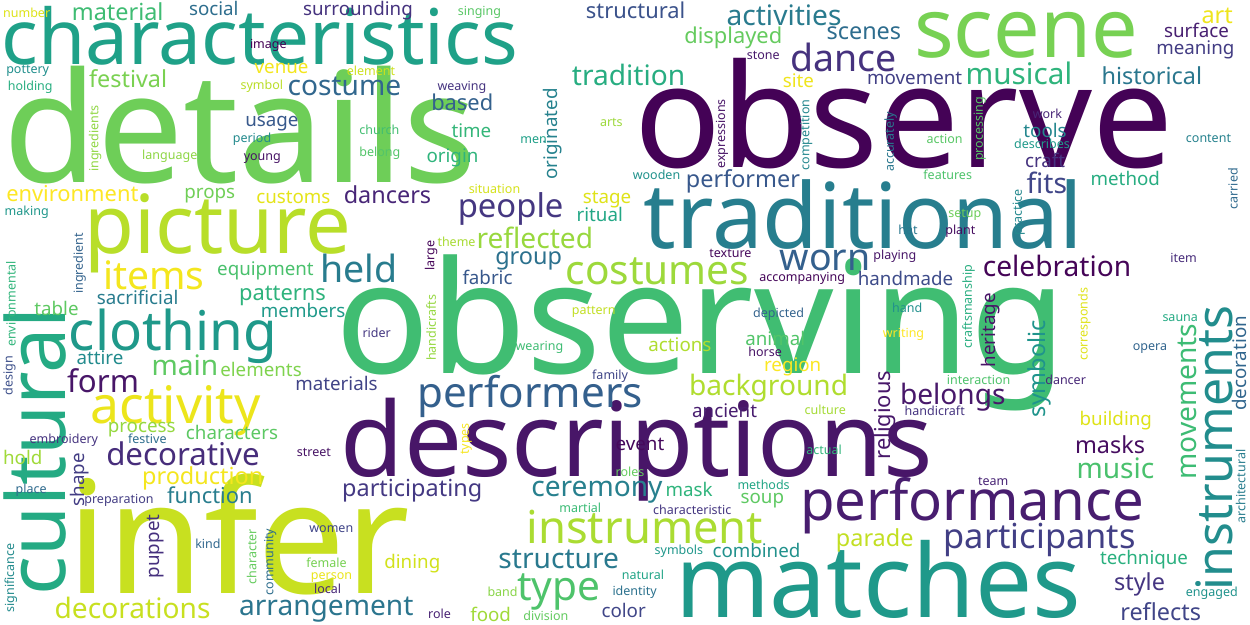}
 \caption{Culture}
 \label{fig:2}
 \end{subfigure}
 \hfill
 \begin{subfigure}[b]{0.32\textwidth}
 \centering
 \includegraphics[width=\linewidth]{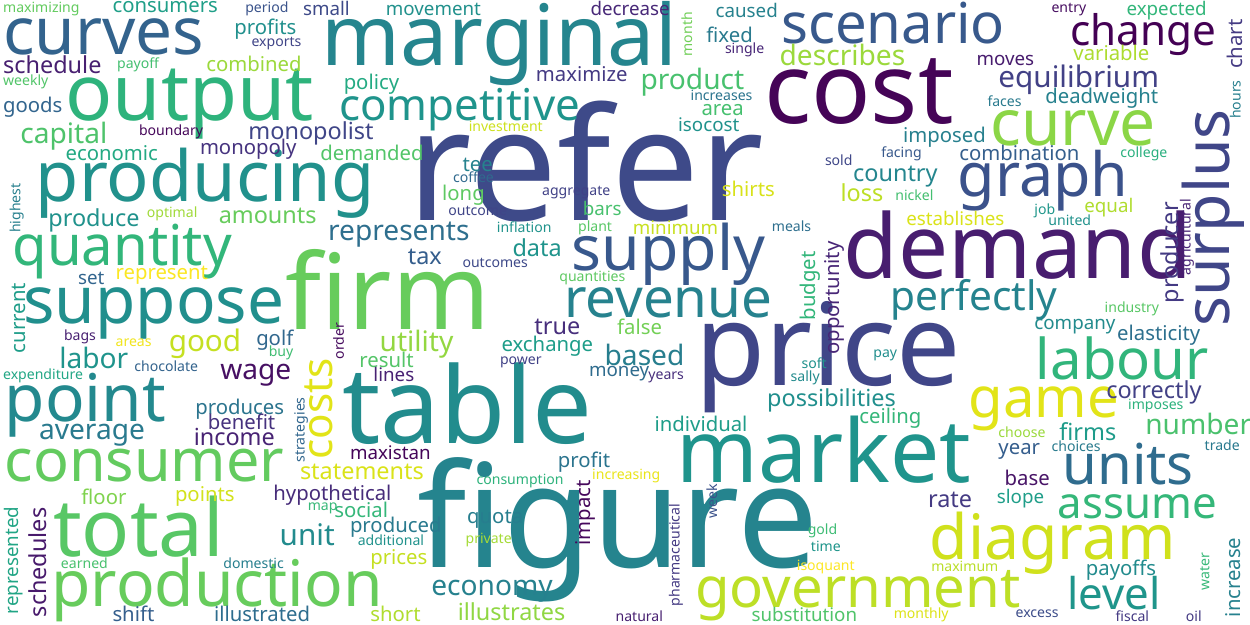}
 \caption{Economy}
 \label{fig:3}
 \end{subfigure}

 \vspace{1em} 

 \begin{subfigure}[b]{0.32\textwidth}
 \centering
\includegraphics[width=\linewidth]{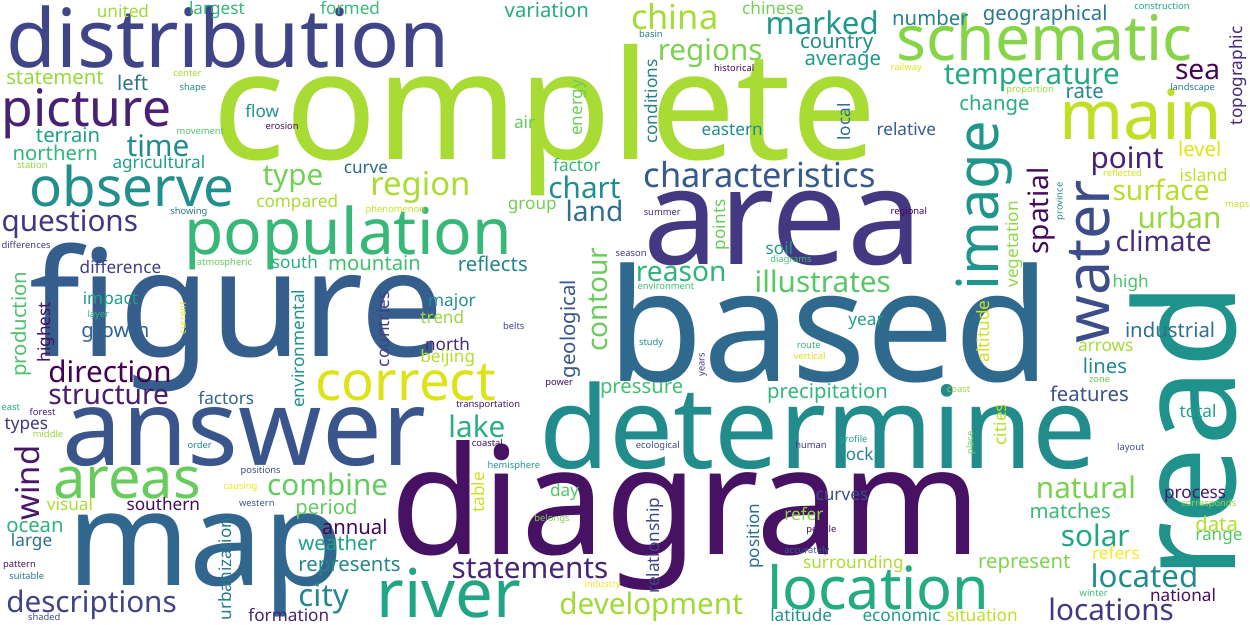}
 \caption{Geography}
 \label{fig:4}
 \end{subfigure}
 \hfill
 \begin{subfigure}[b]{0.32\textwidth}
 \centering
 \includegraphics[width=\linewidth]{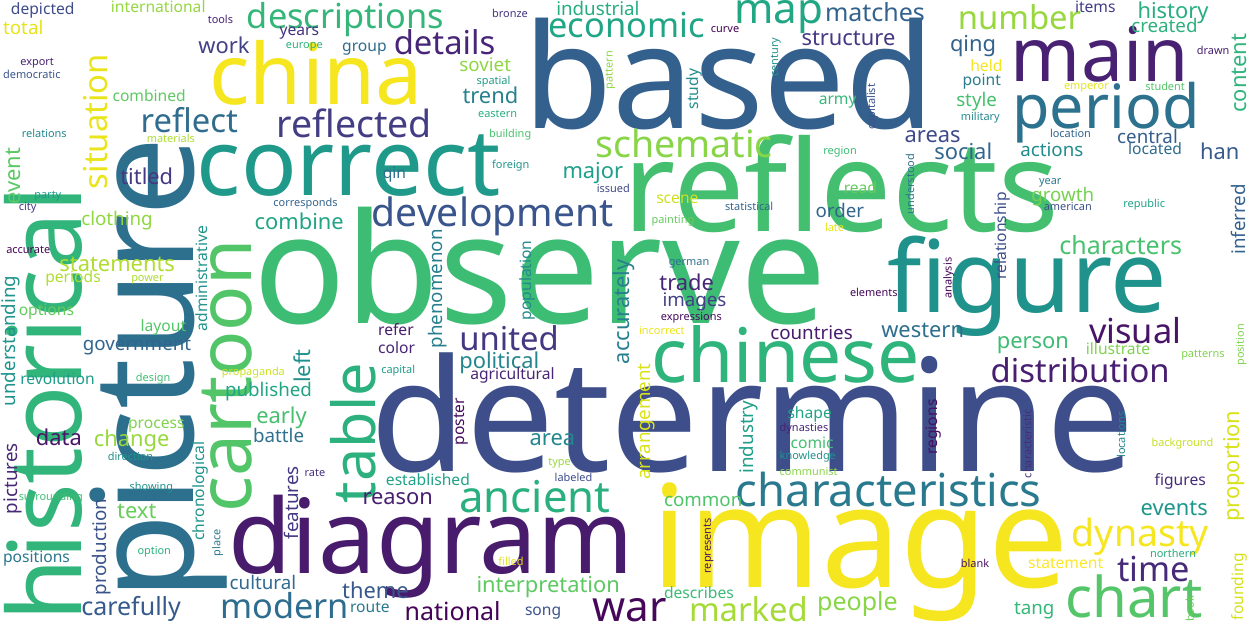}
 \caption{History}
 \label{fig:5}
 \end{subfigure}
 \hfill
 \begin{subfigure}[b]{0.32\textwidth}
 \centering
\includegraphics[width=\linewidth]{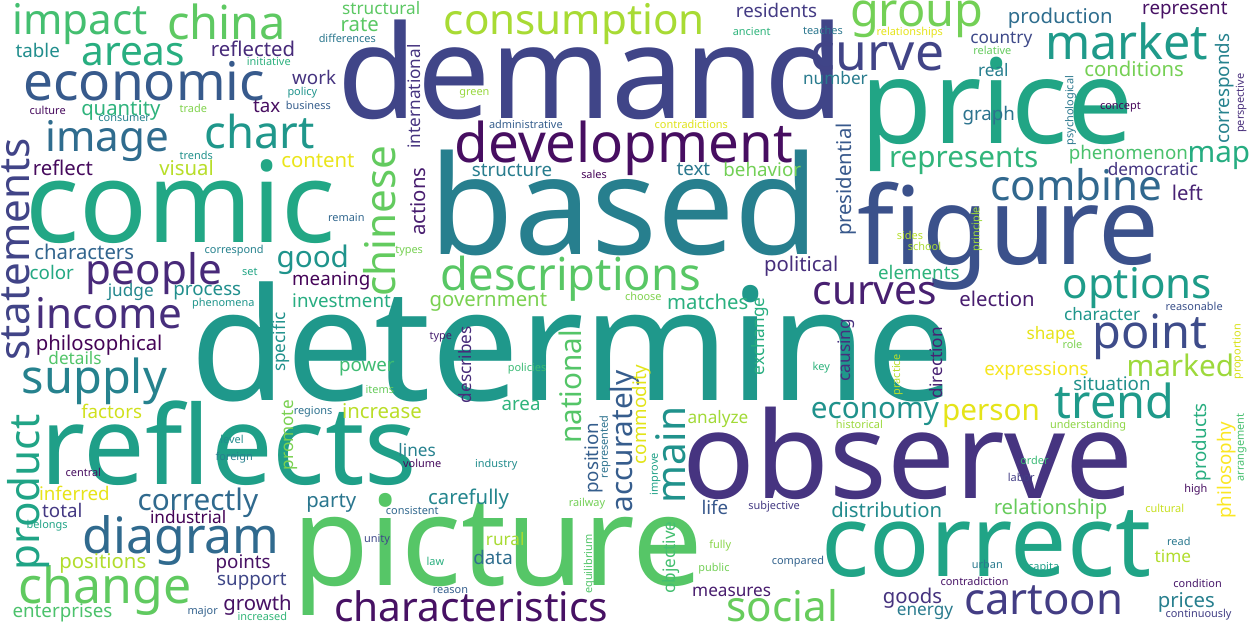}
 \caption{Social sciences}
 \label{fig:6}
 \end{subfigure}

 \caption{Word Cloud in HSSBench per category.}
 \label{fig：six_figs}
\end{figure}

We analyzed the most frequent content words in the questions in all categories to understand the linguistic characteristics and tasksur dataset. Figure~\ref{fig：six_figs} shows the word clouds for the most frequent words in HSSBench per category, excluding common stop words.
The word frequency analysis reveals distinct patterns across categories:

1) Visual observation terms dominate in Art and Culture categories ("observe", "picture", "shown", "scene"), indicating a focus on visual analysis tasks.

2) Economy questions frequently use technical terms ("price", "firm", "cost", "demand", "market", "marginal"), reflecting domain-specific concepts.

3) Geography questions heavily employ spatial and diverse visual comprehension capabilities ("diagram", "map", "area", "distribution"), emphasizing spatial reasoning.

\subsection{Composition and Quality Comparison Between Human and Automated Agents}\label{sec.appendix.Composition.Quality}

We provide a detailed breakdown of the data sources across the six domains in HSSBench, as shown in Table~\ref{tab:data_proportion}.

\begin{table}[htbp]
  \centering
  \begin{tabular}{l|cc|c}
    \toprule
    \textbf{Category} & \textbf{Human Expert} & \textbf{Automated Agent} & \textbf{Total Samples} \\
    \midrule
    Art            & 984  & 1,204 & 2,188 \\
    Geography      & 2,671 & 891 & 3,562 \\
    History        & 1,946 & 550 & 2,496 \\
    Economy        & 974  & 459 & 1,433 \\
    Social Science & 487  & 947 & 1,434 \\
    Culture        & 652  & 1,387 & 2,039 \\
    \bottomrule
  \end{tabular}
  \caption{Distribution of samples constructed by human experts and automated agents across different domains in HSSBench.}
  \label{tab:data_proportion}
\end{table}

To further validate the quality of data generated by both human experts and automated agents, we evaluated the performance of GPT4.1-mini on the respective subsets. The accuracy results are summarized in Table~\ref{tab:accuracy_comparison}.

\begin{table}[htbp]
  \centering
  
  \begin{tabular}{lcc}
    \toprule
    \textbf{Category} & \textbf{Human Expert Acc. } & \textbf{Automated Agent Acc.} \\
    \midrule
    Art            & 42.32 & 45.43 \\
    Geography      & 48.82 & 46.52 \\
    History        & 45.59 & 49.12 \\
    Economy        & 58.87 & 57.67 \\
    Social Science & 43.15 & 46.95 \\
    Culture        & 49.11 & 47.41 \\
    \bottomrule
  \end{tabular}
  \caption{Model accuracy (\%) of GPT4.1-mini on human-expert-generated and automated-agent-generated subsets across different domains.}
  \label{tab:accuracy_comparison}
\end{table}

The comparable accuracy results indicate that data generated by automated agents, after undergoing multi-round expert verification, maintain a quality level consistent with that of human expert contributions. These clarifications and statistics have been incorporated into the revised manuscript to enhance transparency regarding dataset construction and quality assurance.

\section{Experimental details}

\subsection{Details of Results}\label{sec.appendix.detail.results}

Table~\ref{apptab:question_format_score-2} provides additional details on the experimental results presented in Table~\ref{apptab:question_format_score}. The table below reports the overall accuracy of the models under the DIRECT prompt and presents a detailed breakdown of the performance for each category under the same prompt.
\begin{table}[!ht]
    \centering
    \renewcommand{\arraystretch}{1.1}
\resizebox{\textwidth}{!}{
    \begin{tabular}{c|cc|cc|cc|cc|cc|cc|cccc}
    \toprule
    \multirow{2}{*}{Model} 
    & \multicolumn{2}{c|}{Geography} 
    & \multicolumn{2}{c|}{Economics} 
    & \multicolumn{2}{c|}{Culture} 
    & \multicolumn{2}{c|}{Social Sciences} 
    & \multicolumn{2}{c|}{History} 
    & \multicolumn{2}{c|}{Art} 
    & \multicolumn{4}{c}{All} \\
    \cmidrule(lr){2-3} \cmidrule(lr){4-5} \cmidrule(lr){6-7} \cmidrule(lr){8-9} \cmidrule(lr){10-11} \cmidrule(lr){12-13} \cmidrule(lr){14-17}
    & Dr.C. & Dr.O. & Dr.C.& Dr.O.& Dr.C. & Dr.O. & Dr.C. & Dr.O.& Dr.C.& Dr.O.& Dr.C.& Dr.O. & Dr.C.& Dr.O.& Ct.C. & Ct.O. \\ \midrule
        Random & 24.93 & 0.00 &  21.92 & 0.00 & 25.00 & 0.00 & 24.90 & 0.00 & 24.91 & 0.00 & 25.00 & 0.00 & 24.62 & 0.00 & 24.62 & 0.00 \\
        Human & 94.14 & - & 93.06 & - & 92.99 &-  &94.44 & - &  93.84& - & 95.53& -& 93.83 & - & 93.83&-\\
        \midrule
        \multicolumn{16}{c}{\textit{Open-source LLM (Scale $<$ 10B)}}\\ \midrule
        Qwen2.5-VL-3B & 29.84 & 11.27 & 29.44 & 17.76 & 27.83 & 5.09 & 35.22 & 10.00 & 26.91 & 8.61 & 26.78 & 7.24 & 29.01 & 9.94 & 34.99 & 9.33\\
        Qwen2.5-VL-7B & 40.07 & 19.25 & 38.18 & 30.37 & 30.29 & 7.87 & 44.91 & 11.00 & 41.20 & 14.75 & 32.33 & 8.14 & 37.88 & 15.21 & 38.19& 17.89\\
        Llava-onevision-7b & 37.46 & 6.10 & 31.20 & 10.75 & 38.64 & 1.39 & 40.95 & 7.00 & 31.65 & 4.51 & 37.33 & 4.98 & 36.20 & 5.73 & 31.56 & 5.20 \\ 
        Llama3-llava-next-8b & 29.98 & 6.10 & 21.81 & 5.61 & 35.69 & 6.48 & 35.36 & 6.00 & 29.48 & 9.02 & 34.46 & 5.43 & 31.20 & 6.50 & 27.93 & 5.81 \\ 
        InternVL3-8B & 42.88 & 13.62 & 37.08 & 17.97 & 37.71 & 6.91 & 48.94 & 12.00 & 47.22 & 14.34 & 37.48 & 8.60 & 42.14 & 12.27 & 41.42 & 12.31 \\ 
        InternVL2.5-8B-MPO & 38.08 & 14.08 & 34.73 & 16.36 & 39.13 & 6.94 & 45.69 & 10.50 & 40.82 & 11.07 & 38.27 & 11.76 & 39.30 & 11.77 & 37.68 & 15.21\\
        Phi-3.5-Vision-Instruct & 29.84 & 10.33 & 29.44 & 11.21 & 27.83 & 10.65 & 35.22 & 7.00 & 26.91 & 11.48 & 26.78 & 10.86 & 29.01 & 10.32 & 26.04 & 6.96 \\ 
        Janus-Pro  & 27.57 & 8.92 & 16.01 & 7.01 & 42.38 & 6.02 & 31.26 & 9.00 & 28.62 & 12.30 & 32.33 & 7.24 & 30.03 & 8.49 & 31.66 & 8.41 \\ 
        Deepseek-VL2-tiny  & 25.44 &5.16 & 17.38&3.15 & 40.17 &0.46& 30.33& 2.50 &29.97  & 4.51 &35.19 &4.52 & 29.86  & 3.42& 8.23 & 5.09\\
        mPLUG-Owl3-2B  & 25.27 & 4.23 & 30.48 &2.70  &  32.56& 2.30 & 34.17& 4.50 & 26.88  & 6.15 & 30.48 &4.07 & 28.73 &4.02  & 27.71&3.57\\
        mPLUG-Owl3-7B  & 32.71 & 7.04 &33.08&6.31 &  33.45&  4.61&37.31 & 5.50 &  29.97&  8.20& 33.73  & 8.14 & 33.01& 6.68 & 27.52 & 6.23\\
        MiniCPM-o-2.6 & 3.48 & 1.67 &  3.90&  14.16&  2.06 &3.70  &5.75 & 3.03  &4.72& 2.27 & 3.02 &8.70 & 3.70 & 5.71 &24.11 &5.71\\
        Llava1.5 &  3.36 &2.68  & 10.07 & 5.37 &  3.36 & 2.01 & 8.05 & 4.70 &8.05 &  1.34&15.44 &8.05 &8.06&  4.03 & 9.96&4.25\\

        \midrule
        \multicolumn{16}{c}{\textit{Open-source LLM (Scale $>$ 10B)}}\\ \midrule
        Qwen2.5-VL-32B  & 52.54 & 16.00 & 39.38 & 8.00 &39.87  & 10.00  & 53.09&   26.67 & 57.52 & 24.67 & 44.92& 10.00& 48.38 & 15.89 & 50.75&15.00\\
        Qwen2.5-VL-72B  & 56.71 & 16.00 &51.67  & 32.00 & 57.77 & 5.33 &47.43 &  16.89& 59.26&24.16  &63.12& 14.67  &54.17  & 18.17&51.87&19.73\\
        QVQ-72B-Preview  & 21.53 &  3.33& 18.38 & 22.00 &  29.57 &  2.67& 26.26& 10.14 &  29.58& 12.75 & 23.85  &11.33 & 25.60 & 10.37 &24.69 &9.92\\ 
        
         Qwen2-VL-72B &54.22   &  19.30& 41.06 & 24.07 & 55.33 & 5.66 &58.85 & 12.96 & 62.17 & 32.08 & 50.00& 18.87& 54.22  &  17.79 & 49.39&18.82\\ 
        \midrule
        \multicolumn{16}{c}{\textit{Closed-source LLM}}\\ \midrule
        GPT-4o  & 45.68 & 23.47 & 48.92 &  35.14& 48.16 & 15.67 & 46.72 & 18.00 & 47.24 & 14.75 &46.66 &13.57 & 46.99 & 20.05 & 46.88&19.36\\
         GPT-4.1   & 43.80 & 26.29 & 52.34 & 42.34 &  52.77& 18.89 &40.86 & 26.13 & 38.86 &22.13  & 44.74& 16.74& 45.02 & 25.38 &42.66 &39.97\\
        GPT-4.1-mini   & 45.14 &26.29  &  54.92& 45.95  &  52.08& 15.21 & 41.84&22.11& 39.34 & 18.85 &44.70 &17.65 &45.75  & 24.32 &48.03 &33.59\\
        GPT-4.1-nano   &  33.18& 20.66 & 34.96 & 36.04 &  45.61& 14.75 & 38.42&   15.58&32.69  & 20.49 & 36.47&18.55 &36.33  &21.12  & 35.83&26.22\\
        \bottomrule

    \end{tabular}
    }
    \caption{Scores (\%) of MLLMs on HSSBench (EN-\Rmnum{2}).}
    
    \label{apptab:question_format_score-2}
\end{table}

\begin{figure}[htbp]
    \centering
    \begin{subfigure}[b]{0.49\textwidth}
        \centering
        \includegraphics[width=\linewidth]{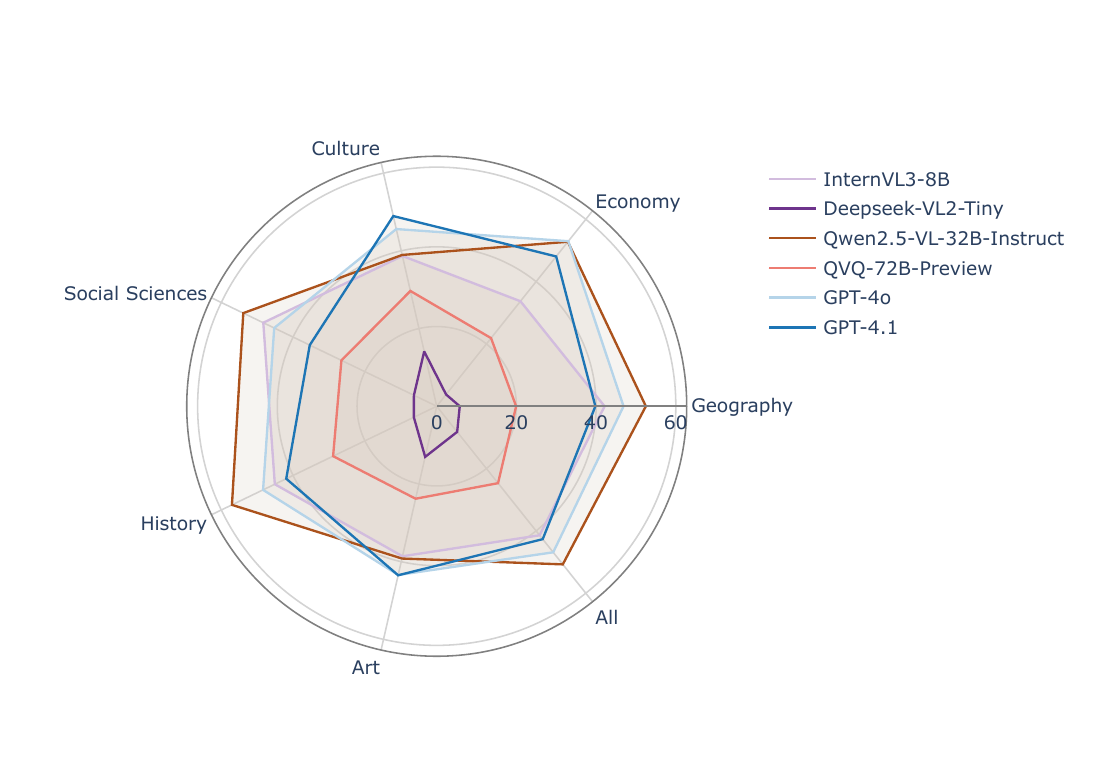} 
        \caption{Effect under multiple-choice question setting}
        \label{fig:left}
    \end{subfigure}
    \hfill
    \begin{subfigure}[b]{0.49\textwidth}
        \centering
        \includegraphics[width=\linewidth]{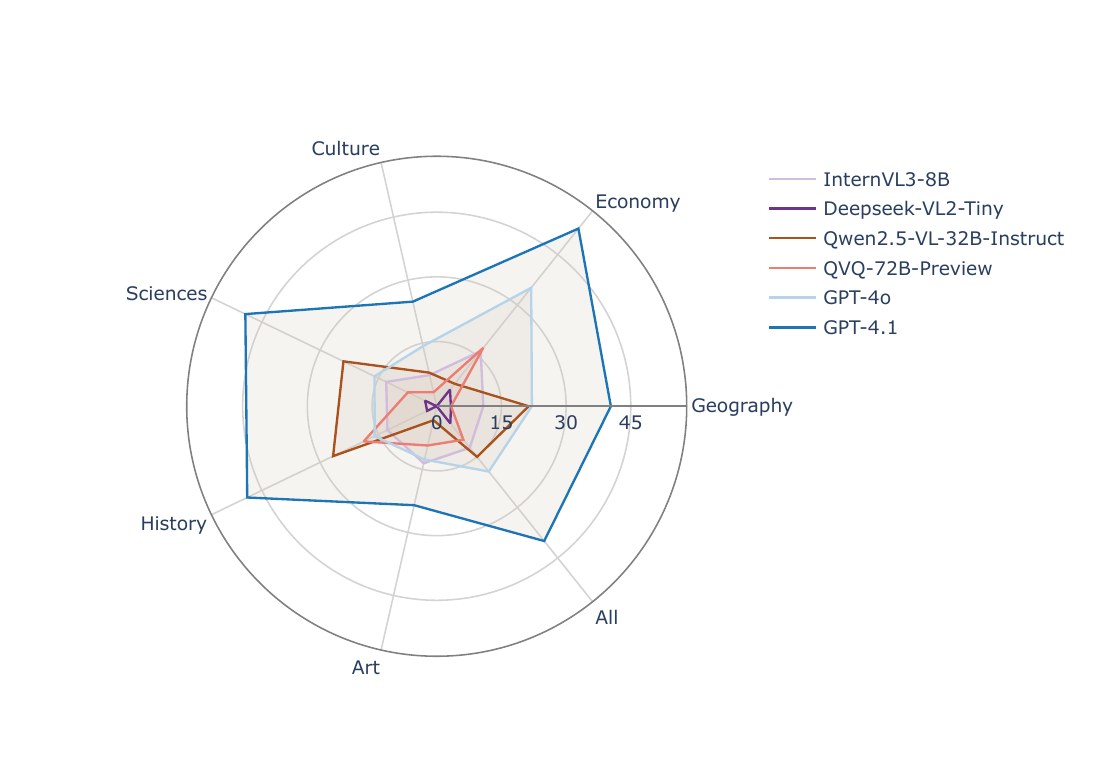} 
        \caption{Effect under open-ended question setting}
        \label{fig:right}
    \end{subfigure}
    \caption{Partial MLLMs' results under CoT prompt settings.}
    \label{fig:cot_effects}
\end{figure}

\subsection{Results in different languages}\label{sec.appendix.detail.results.languages}
Table~\ref{apptab:question_format_score} and Table~\ref{apptab:question_format_score-2} present the test results on English-language data. To analyze the impact of different languages on the model, we also conducted large-scale experiments using Chinese-language data. The final experimental results are shown in Table~\ref{apptab:question_format_score3} and Table~\ref{apptab:score4} below.
\begin{table}[htbp]
    \centering
    \renewcommand{\arraystretch}{1.1}
    
\resizebox{\textwidth}{!}{
    \begin{tabular}{c|cc|cc|cc|cc|cc|cc|cccc}
    \toprule
    \multirow{2}{*}{Model} 
    & \multicolumn{2}{c|}{Geography} 
    & \multicolumn{2}{c|}{Economics} 
    & \multicolumn{2}{c|}{Culture} 
    & \multicolumn{2}{c|}{Social Sciences} 
    & \multicolumn{2}{c|}{History} 
    & \multicolumn{2}{c|}{Art} 
    & \multicolumn{4}{c}{All} \\
    \cmidrule(lr){2-3} \cmidrule(lr){4-5} \cmidrule(lr){6-7} \cmidrule(lr){8-9} \cmidrule(lr){10-11} \cmidrule(lr){12-13} \cmidrule(lr){14-17}
    & Ct.C. & Ct.O. & Ct.C. & Ct.O. & Ct.C. & Ct.O. & Ct.C. & Ct.O. & Ct.C. & Ct.O. & Ct.C. & Ct.O. & Dr.C. & Dr.O. & Ct.C. & Ct.O. \\ \midrule
        Random & 24.93 & 0.00 &  21.92 & 0.00 & 25.00 & 0.00 & 24.90 & 0.00 & 24.91 & 0.00 & 25.00 & 0.00 & 24.62 & 0.00 & 24.62 & 0.00 \\
        Human & 94.14 & - & 93.06 & - & 92.99 &-  &94.44 & - &  93.84& - & 95.53& -& 93.83 & - & 93.83&-\\
        \midrule
        \multicolumn{16}{c}{\textit{Open-source LLM (Scale $<$ 10B)}}\\ \midrule
        Qwen2.5-VL-3B & 40.49 & 10.80 & 28.89 & 19.81 & 36.30 & 0.93 & 42.43 & 9.50 & 43.91 & 11.89 & 33.85 & 4.98 & 35.45 & 11.79 & 35.45 & 9.65 \\ 
        Qwen2.5-VL-7B & 48.43 & 15.96 & 35.00 & 25.00 & 27.60 & 2.78 & 49.36 & 19.50 & 50.54 & 15.98 & 32.90 & 6.79 & 43.12 & 13.17 & 41.86 & 14.24 \\
        Llava-onevision-7b & 39.68 & 4.69 & 30.80 & 10.38 & 32.96 & 1.85 & 39.46 & 7.00 & 32.97 & 4.51 & 35.03 & 2.71 & 37.89 & 4.98 & 35.62 & 5.13 \\ 
        Llama3-llava-next-8b & 27.29 & 3.29 & 19.82 & 3.77 & 33.55 & 0.93 & 32.89 & 2.50 & 26.84 & 3.28 & 29.48 & 2.26 & 31.18 & 5.21 & 28.35 & 2.68 \\ 
        InternVL3-8B & 47.45 & 11.74 & 35.37 & 17.92 & 38.21 & 9.26 & 53.47 & 14.00 & 50.97 & 18.44 & 39.94 & 10.41 & 44.81 & 14.70 & 44.92 & 13.71 \\
        InternVL2.5-8B-MPO & 41.14 & 11.74 & 36.11 & 23.58 & 36.20 & 6.94 & 48.09 & 16.50 & 47.63 & 6.94 & 38.35 & 10.86 & 42.72 & 14.01 & 41.45 & 14.55 \\
        Phi-3.5-vision-instruct  & 22.29 & 6.10 & 25.64 & 4.25 & 30.16 & 2.31 & 23.55 & 1.00 & 23.35 & 4.92 & 28.30 & 1.81 & 34.08 & 6.74 & 25.35 & 3.45 \\ 
        Janus-Pro  & 28.10 & 11.27 & 21.67 & 8.49 & 44.16 & 6.48 & 34.79 & 8.50 & 29.13 & 14.34 & 40.49 & 14.48 & 33.50 & 11.94 & 32.81 & 10.72 \\ 
        mPLUG-Owl3-2B  & 27.20 & 1.88 &29.43  & 0.90&33.74 & 2.30 &29.78  &3.00  & 26.76 & 2.87 & 29.43 &4.52 &  32.25&  1.90 & 28.25&2.58\\
        mPLUG-Owl3-7B  & 35.01 &  5.63& 34.23&3.15 & 26.39  &  1.84& 39.26& 2.50 & 26.08 & 2.46 & 34.23 &  5.43 &  34.22 & 4.71 &31.77 &3.49\\
        MiniCPM-o-2.6  & 26.97 & 7.69 &19.50  & 4.88 &  35.21 &6.25  & 31.23& 3.90 & 30.27 & 6.76 &  30.20&5.98 &1.29  & 6.83 &29.04 &5.94\\
        \midrule
        \multicolumn{16}{c}{\textit{Open-source LLM (Scale $>$ 10B)}}\\ \midrule
        Qwen2.5-VL-32B  & 58.45 & 19.33 & 49.86 &  7.33 & 40.90 &  5.33&56.49 & 22.67& 64.33 &24.67  & 41.62&14.67 & 51.60 & 14.22 &  52.86&15.67 \\
        Qwen2.5-VL-72B  & 61.65 & 12.67 & 52.02 & 33.33 & 43.60 &   10.67& 61.04& 14.19 & 68.08 &18.12  &  43.76&16.00 & 60.04 & 21.96 &55.94 &17.51\\
        QVQ-72B-Preview  & 19.82    &  8.00 &  21.10 & 18.00 & 30.80 &   3.33 & 25.41  & 6.76 & 27.85 &  10.74& 25.90 & 6.67 &22.95  & 9.14 &  24.82&8.92\\
        Qwen2-VL-72B  & 21.28 & 19.30 & 38.25 &  24.07 &21.25   & 5.66  &27.79 &  12.96 &  25.69& 32.08 & 27.65&18.87& 59.11 &   17.79 &25.59 &18.82\\ 
       
        \midrule
        \multicolumn{16}{c}{\textit{Closed-source LLM}}\\ \midrule
        GPT-4.1mini   &  47.53&  35.21 & 55.55 &55.41  & 44.43 & 18.43 & 44.07& 34.17 & 47.92 & 43.03 & 42.50&30.32 &44.19  & 24.54 &46.78 &36.32\\
        \bottomrule
        
    \end{tabular}
    }
    \caption{Scores (\%) of MLLMs on HSSBench (ZH-\Rmnum{1}).}
    \label{apptab:question_format_score3}
\end{table}

\begin{table}[htbp]
    \centering
    \renewcommand{\arraystretch}{1.1}
\resizebox{\textwidth}{!}{
    \begin{tabular}{c|cc|cc|cc|cc|cc|cc|cccc}
    \toprule
    \multirow{2}{*}{Model} 
    & \multicolumn{2}{c|}{Geography} 
    & \multicolumn{2}{c|}{Economics} 
    & \multicolumn{2}{c|}{Culture} 
    & \multicolumn{2}{c|}{Social Sciences} & \multicolumn{2}{c|}{History} 
    & \multicolumn{2}{c|}{Art} 
    & \multicolumn{4}{c}{All} \\
    \cmidrule(lr){2-3} \cmidrule(lr){4-5} \cmidrule(lr){6-7} \cmidrule(lr){8-9} \cmidrule(lr){10-11} \cmidrule(lr){12-13} \cmidrule(lr){14-17}
    & Dr.C.g & Dr.O.g & Dr.C.g & Dr.O.g & Dr.C.g & Dr.O. & Dr.C.g & Dr.O.g & Dr.C.g & Dr.O.g & Dr.C.g & Dr.O.g & Dr.C.g & Dr.O.g & Ct.C.g & Ct.O.g \\ \midrule
       Random & 24.93 & 0.00 &  21.92 & 0.00 & 25.00 & 0.00 & 24.90 & 0.00 & 24.91 & 0.00 & 25.00 & 0.00 & 24.62 & 0.00 & 24.62 & 0.00 \\
        Human & 94.14 & - & 93.06 & - & 92.99 &-  &94.44 & - &  93.84& - & 95.53& -& 93.83 & - & 93.83&-\\
        \midrule
        \multicolumn{16}{c}{\textit{Open-source LLM (Scale $<$ 10B)}}\\ \midrule
        Qwen2.5-VL-3B & 39.51 & 12.21 & 29.26 & 13.21 & 30.94 & 7.87 & 40.03 & 14.50 & 38.36 & 11.48 & 30.03 & 11.76 & 35.45 & 11.79 & 35.45 & 9.65 \\
        Qwen2.5-VL-7B & 52.52 &  17.84 & 33.97 & 20.75 & 27.60 & 4.63 & 51.13 & 15.00 & 52.13 & 11.48 & 31.17 & 9.95 & 43.12 & 13.17 & 41.86 & 14.24 \\ 
        Llava-onevision-7b & 42.20 & 6.57 & 30.21 & 7.55 & 36.59 & 1.39 & 42.93 & 5.00 & 35.61 & 4.10 & 36.12 & 5.43 & 37.89 & 4.98 & 35.62& 5.13 \\ 
        Llama3-llava-next-8b & 29.61 & 4.69 & 19.97 & 6.60 & 38.80 & 5.09 & 34.37 & 5.00 & 28.86 & 3.28 & 34.54 & 6.33 & 31.18 & 5.21 & 28.35 & 2.68 \\ 
        InternVL3-8B & 46.44 & 18.78 & 34.27 & 20.75 & 40.08 & 6.94 & 52.90 & 18.00 & 51.05 & 14.34 & 40.19 & 9.95 & 44.81 & 14.70 & 44.92 & 13.71 \\ 
        InternVL2.5-8B-MPO & 42.26 & 18.78 & 31.32 & 14.15 & 40.72 & 7.87 & 48.87 & 17.50 & 49.11 & 12.30 & 40.78 & 14.03 & 42.72 & 14.01 & 41.45 & 14.55\\
        Phi-3.5-vision-instruct  & 29.92 & 6.57 & 27.49 & 7.08 & 45.58 & 7.87 & 32.89 & 4.50 & 29.71 & 6.56 & 40.68 & 7.69 & 34.08 & 6.74 & 25.35 & 3.45 \\ 
        Janus-Pro & 27.29 & 9.86 & 18.50 & 12.74 & 50.83 & 14.35 & 32.18 & 8.50 & 29.44 & 10.25 & 43.21 & 15.84 & 33.50 & 11.94 & 32.81 & 10.72 \\ 
        mPLUG-Owl3-2B  & 31.25& 0.94 & 23.73 &2.71  & 35.95  & 0.46 & 37.17 &3.00  & 31.89 &   2.46&  33.18& 2.71& 32.25&  1.90 & 28.25&2.58\\
        mPLUG-Owl3-7B  & 35.85 & 7.51 &35.65&  6.79& 35.70 & 1.38 &40.59 &  3.00&28.17  & 4.10 & 35.65 & 6.79&  34.22 &  4.71&31.77 &3.49\\
        MiniCPM-o-2.6   & 0.93 & 7.55 &0.77&  8.06& 1.42 & 1.61 &  1.99& 5.41 & 0.60 & 20.00 & 2.42& 2.27 & 1.29  &6.83  &29.04 &5.94\\ 
        
        \midrule
        \multicolumn{16}{c}{\textit{Open-source LLM (Scale $>$ 10B)}}\\ \midrule
        Qwen2.5-VL-32B  & 56.29 &   20.67 &42.83 & 6.00 & 41.83 & 4.00 &57.42 &18.00  & 63.00 & 25.33  & 42.21&11.33 & 51.60 &  14.22&  52.86& 15.67\\
        Qwen2.5-VL-72B  & 68.15 &  22.67& 48.40 & 30.00 & 50.07 &  12.00& 63.73&  20.27 &  71.02& 31.54 &49.02 & 15.33& 60.04 & 21.96 &55.94 &17.51\\
        QVQ-72B-Preview  & 19.40  & 6.67 & 17.90 &20.00  &  29.72  &  4.00 & 24.34  & 11.49 &  28.13   & 8.05 & 23.44  &4.67 &22.95  & 9.14 &  24.82&8.92\\
        Qwen2-VL-72B  &  64.42& 28.07 & 41.83 & 16.67 & 55.98 &  7.55& 62.00& 14.81 &68.77  & 22.64 &51.49 & 16.98&59.11 &   17.79 &25.59 &18.82\\
        \midrule
        \multicolumn{16}{c}{\textit{Closed-source LLM}}\\ \midrule
        GPT-4.1mini  & 45.93 &26.76  &53.31  &43.69  & 46.25 & 14.29 & 39.61&  20.10 & 39.26 & 24.18 & 42.09& 17.65& 44.19 & 24.54 & 46.78&36.32\\
        \bottomrule
        
    \end{tabular}
    }
    \caption{Scores (\%) of MLLMs on HSSBench (ZH-\Rmnum{2}).}
    \label{apptab:score4}
\end{table}

\begin{figure}[htbp]
    \centering
    \includegraphics[width=1\linewidth]{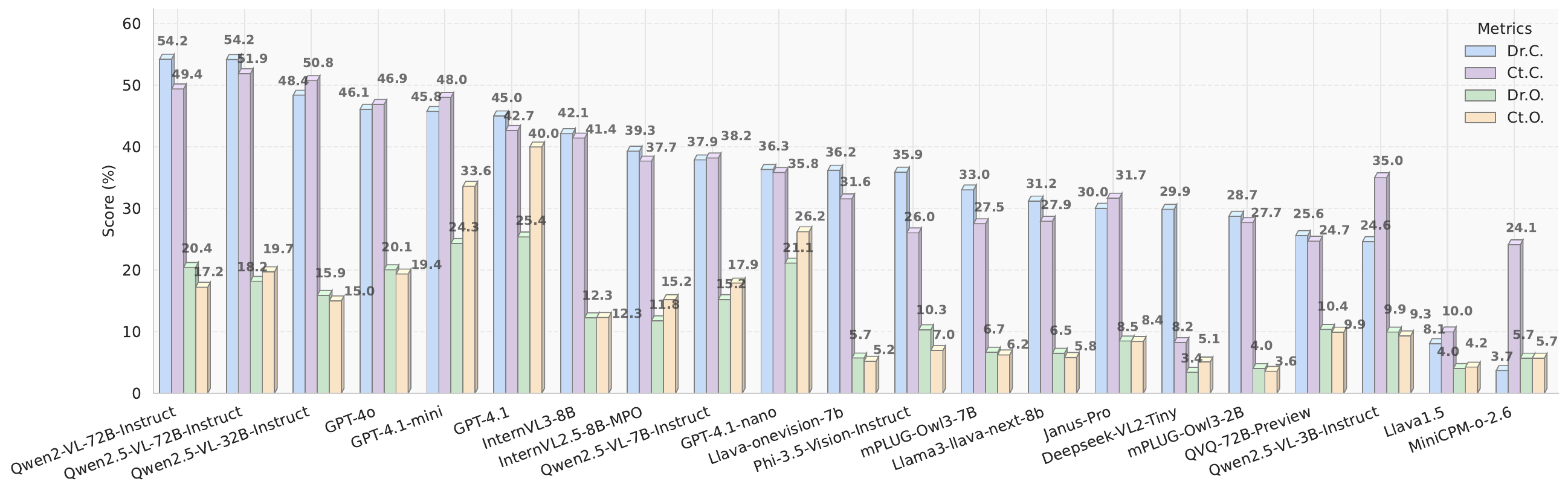}
    \caption{Comparison of Model Performances Across Four Prompt Settings.}
    \label{all_image}
\end{figure}

Figure~\ref{all_image} illustrates the performance of various models under four different prompt configurations.

\begin{figure}[htbp]
    \centering  
    \includegraphics[width=0.6\linewidth]{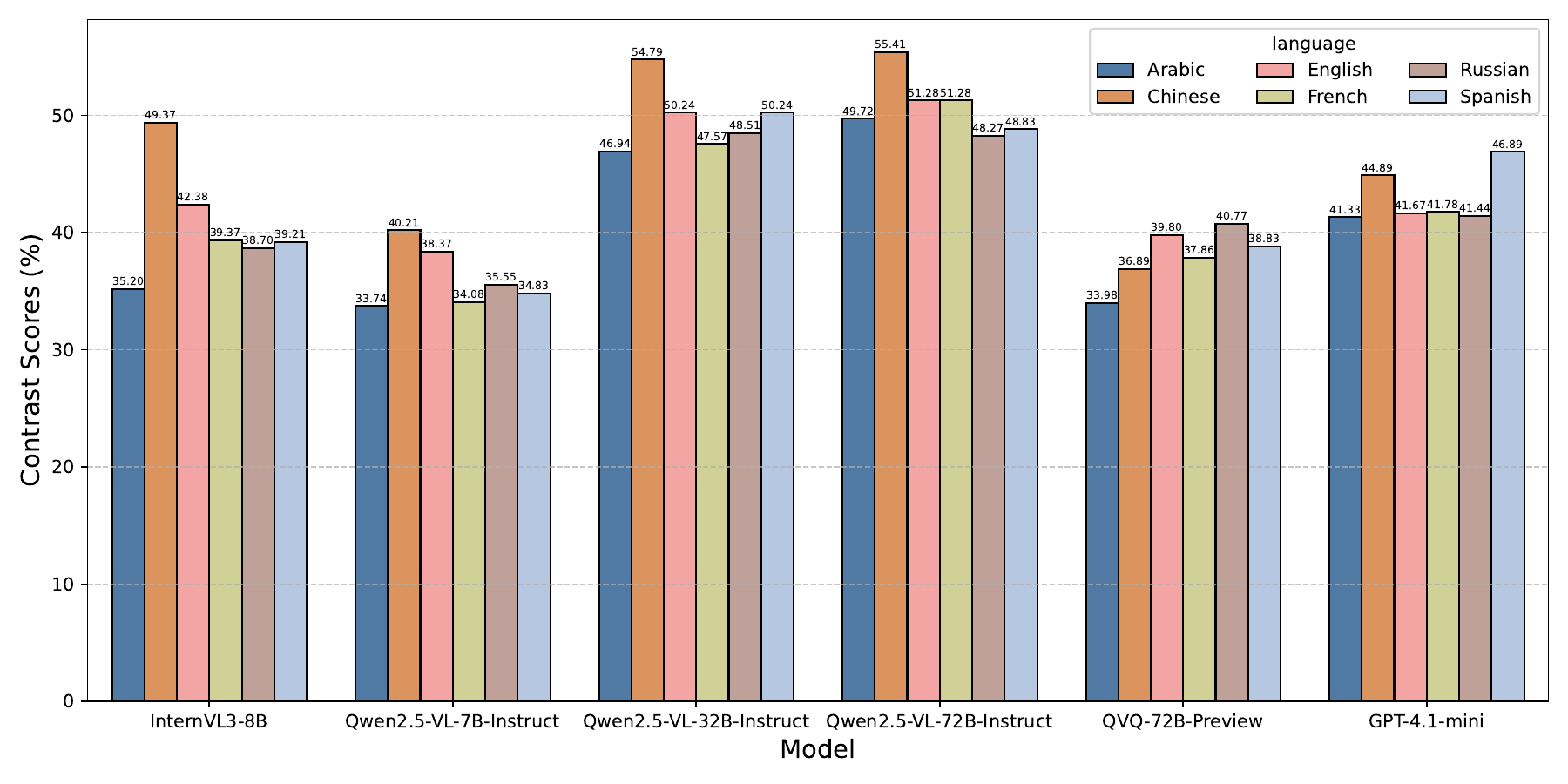}
    \caption{Contrast Scores (\%) of MLLMs on HSSBench for six UN languages and six models.}
    \label{table7}
\end{figure}

Table~\ref{apptab:question_contrast-language} and Figure~\ref{table7}
shows the performance of the model on datasets in different languages. Our dataset includes data organized in six languages. The table below presents the experimental results on a stratified sample of 900 instances.

\subsection{Multiple-Choice Confounding Option Experiment Details}\label{sec.appendix.confounding}

\begin{table}[htbp]
    \centering
    \resizebox{0.6\textwidth}{!}{ 
    \begin{tabular}{c|cccccc}
    \toprule
    Model & Arabic & Chinese & English & French & Russian & Spanish \\
    \midrule
    InternVL3-8B   &  35.20   &    49.37   &    42.38   &    39.37   &    38.70   &   39.21      \\
    Qwen2.5-VL-7B-Instruct  &  33.74   &   40.21    &    38.37   &    34.08   &   35.55    &    34.83     \\
    Qwen2.5-VL-32B-Instruct  &  46.94    &  54.79     &   50.24    &   47.57    &  48.51     &  50.24     \\
    Qwen2.5-VL-72B-Instruct     &  49.72 &  55.41  &   51.28  &51.28&48.27    & 48.83     \\
    QVQ-72B-Preview      &     33.98   &    36.89   &    39.80   &   37.86    &   40.77    &   38.83     \\
    GPT-4.1-mini&   41.33    & 44.89      &  41.67     &  41.78     &     41.44  &      46.89 \\
    \midrule
     Average     & 40.82 & 46.93 & 43.96 & 41.99 & 42.87 & 43.47 \\
    \bottomrule
    \end{tabular}
    }
    \caption{Contrast Scores (\%) of MLLMs on HSSBench for six UN languages and six models.}
    \label{apptab:question_contrast-language}
\end{table}

\begin{table}[http]
    \centering
    \renewcommand{\arraystretch}{1}
    \resizebox{\textwidth}{!}{
    \begin{tabular}{c|cc|cc|cc|cc|cc|cc|cc}
    \toprule
    \multirow{2}{*}{Model} 
    & \multicolumn{2}{c|}{Geography} 
    & \multicolumn{2}{c|}{Art} 
    & \multicolumn{2}{c|}{Culture} 
    & \multicolumn{2}{c|}{Social Sciences} 
    & \multicolumn{2}{c|}{History} 
    & \multicolumn{2}{c|}{Economy} 
    & \multicolumn{2}{c}{Total} \\
    \cmidrule(lr){2-3} \cmidrule(lr){4-5} \cmidrule(lr){6-7} \cmidrule(lr){8-9} \cmidrule(lr){10-11} \cmidrule(lr){12-13} \cmidrule(lr){14-15}
    & Ct.C. & Ct.C.Conf. & Ct.C. &  Ct.C.Conf.& Ct.C. &  Ct.C.Conf.& Ct.C. &  Ct.C.Conf.& Ct.C. &  Ct.C.Conf. & Ct.C. &  Ct.C.Conf. & Ct.C. & Ct.C.Conf. \\
    \midrule
    Qwen2.5-VL-7B & 48.00 & 44.67 & 32.12 & 31.69 & 29.33 & 26.67 & 39.86 & 34.00 & 55.33 & 49.33 & 35.51 & 36.05 & 40.21 & 37.12 \\
    InternVL3-8B & 50.67 & 46.00 & 41.55 & 34.01 & 45.33 & 49.33 & 53.38 & 48.67 & 60.00 & 56.00 & 39.86 & 45.99 & 49.37 & 46.01 \\
    MiniCPM-o & 27.51 & 21.47 & 39.33 & 34.67 & 49.33 & 50.67 & 35.81 & 29.73 & 38.26 & 26.85 & 27.33 & 23.33 & 36.34 & 31.22 \\
    Qwen2.5-VL-32 &  50.83& 49.83 & 46.10 & 40.90 & 41.55 & 30.99 & 59.61 & 51.92 & 61.98 & 56.77 & 50.94 & 53.77 & 51.85 & 47.74 \\
    QvQ-72B-Preview &  16.67 & 16.67 & 29.33 & 28.00 & 35.78 & 29.67 & 26.35 & 18.91 & 28.18 & 27.51 & 25.33 & 24.66 & 26.98 & 24.41 \\
    Qwen2.5-VL-72 &  60.67 & 55.33 & 45.33 & 42.05 & 49.33 & 44.67 & 56.76 & 55.40 & 65.77 & 67.11 & 54.67 & 55.33 & 55.41 & 53.29 \\
    GPT-4.1-mini &46.67  &47.33  &  56.67& 54.67 & 52.67 &47.33  & 40.00 & 42.67 & 54.67 &  44.00& 45.33 &  42.00& 49.33 &46.33 \\
    \bottomrule
    \end{tabular}
    }
    \caption{Contrast Scores (\%) with confounding option of MLLMs on HSSBench.}
    \label{apptab:question_contrast2}
\end{table}

\begin{figure}[htbp]
 \centering
 \begin{subfigure}[b]{0.32\textwidth}
 \centering
 \includegraphics[width=\linewidth]{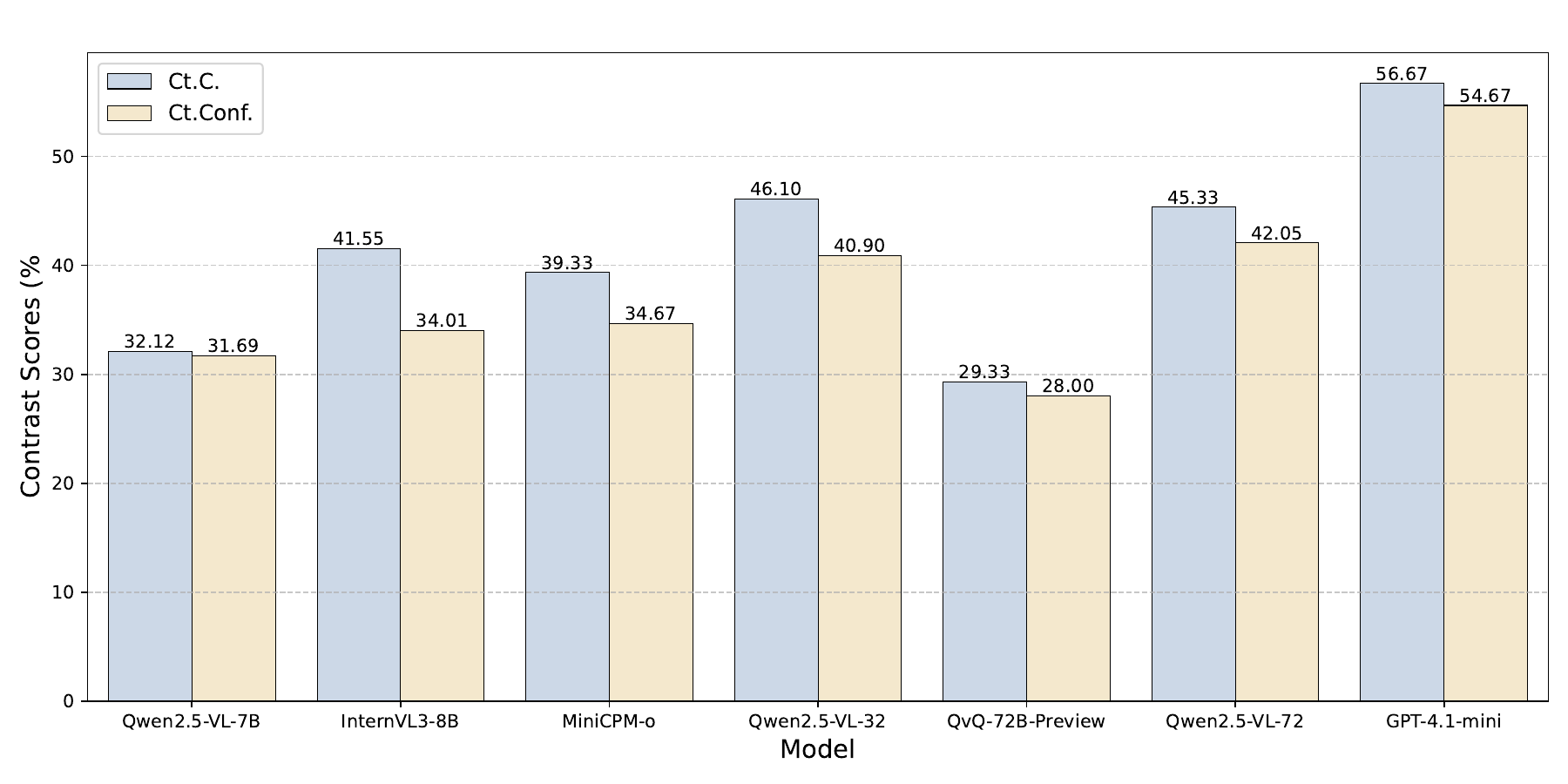}
 \caption{Art}
 \label{fig：1}
 \end{subfigure}
 \hfill
 \begin{subfigure}[b]{0.32\textwidth}
 \centering
 \includegraphics[width=\linewidth]{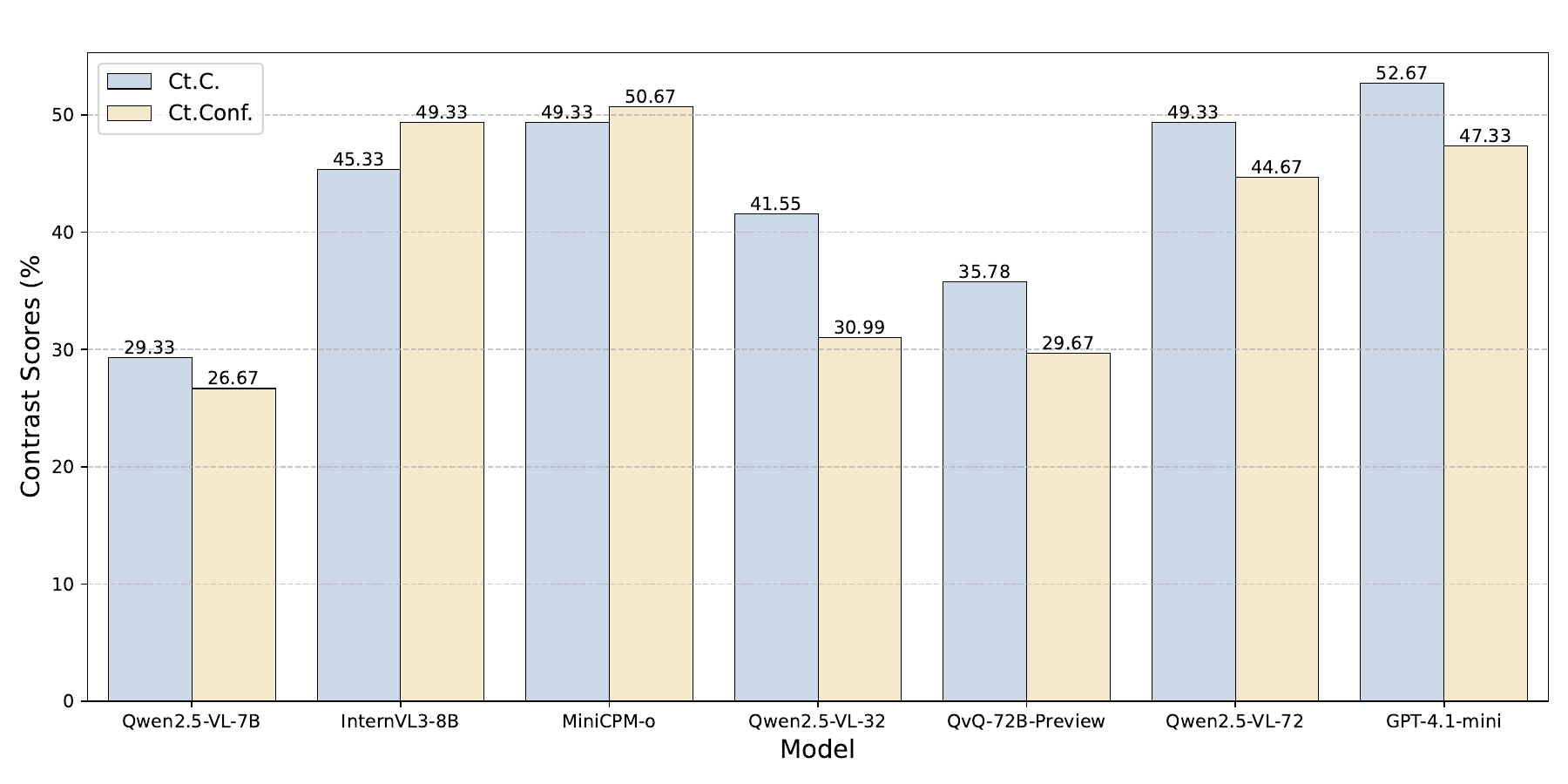}
 \caption{Culture}
 \label{fig：2}
 \end{subfigure}
 \hfill
 \begin{subfigure}[b]{0.32\textwidth}
 \centering
 \includegraphics[width=\linewidth]{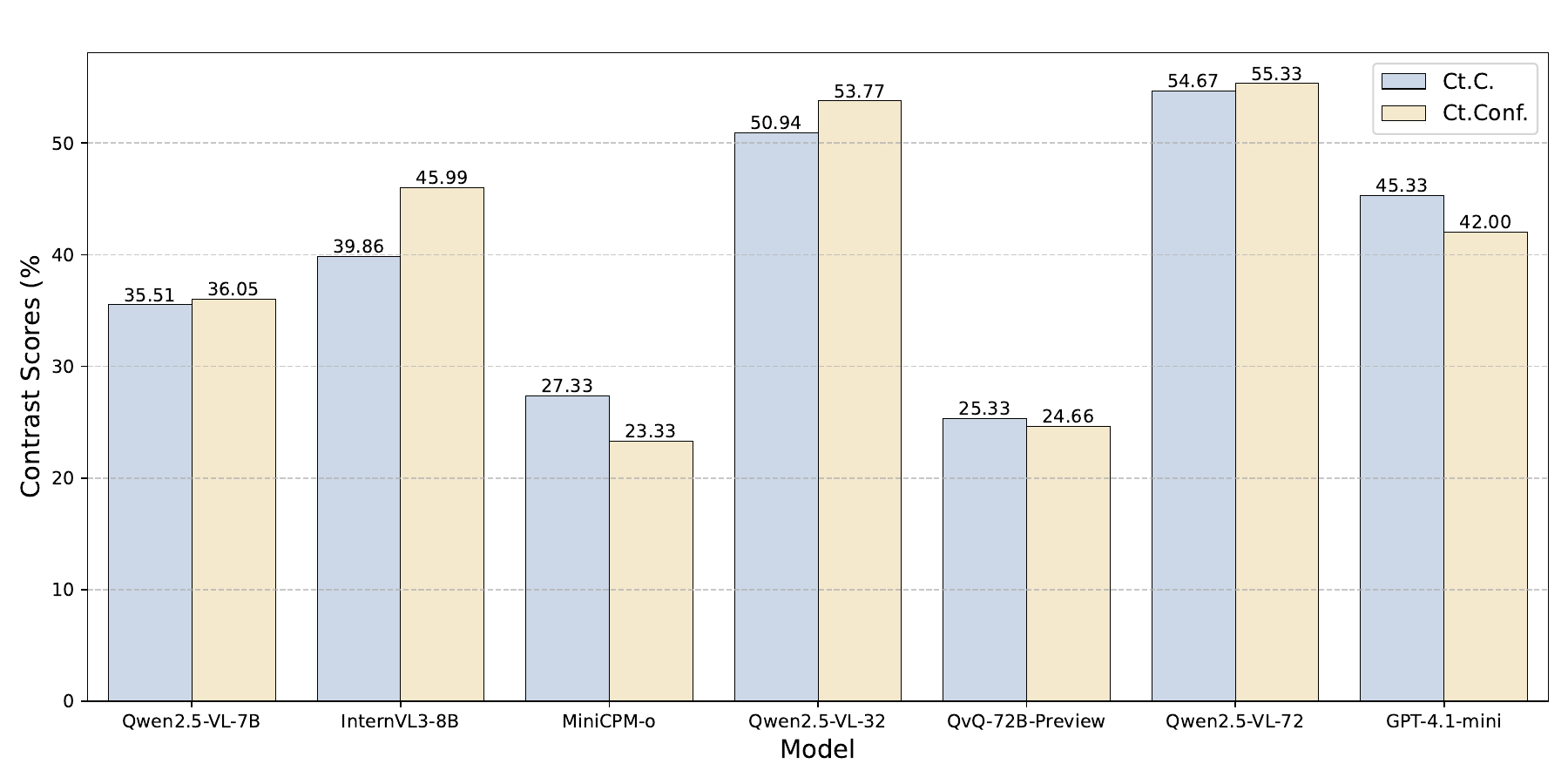}
 \caption{Economy}
 \label{fig：3}
 \end{subfigure}

 \vspace{1em} 

 \begin{subfigure}[b]{0.32\textwidth}
 \centering
\includegraphics[width=\linewidth]{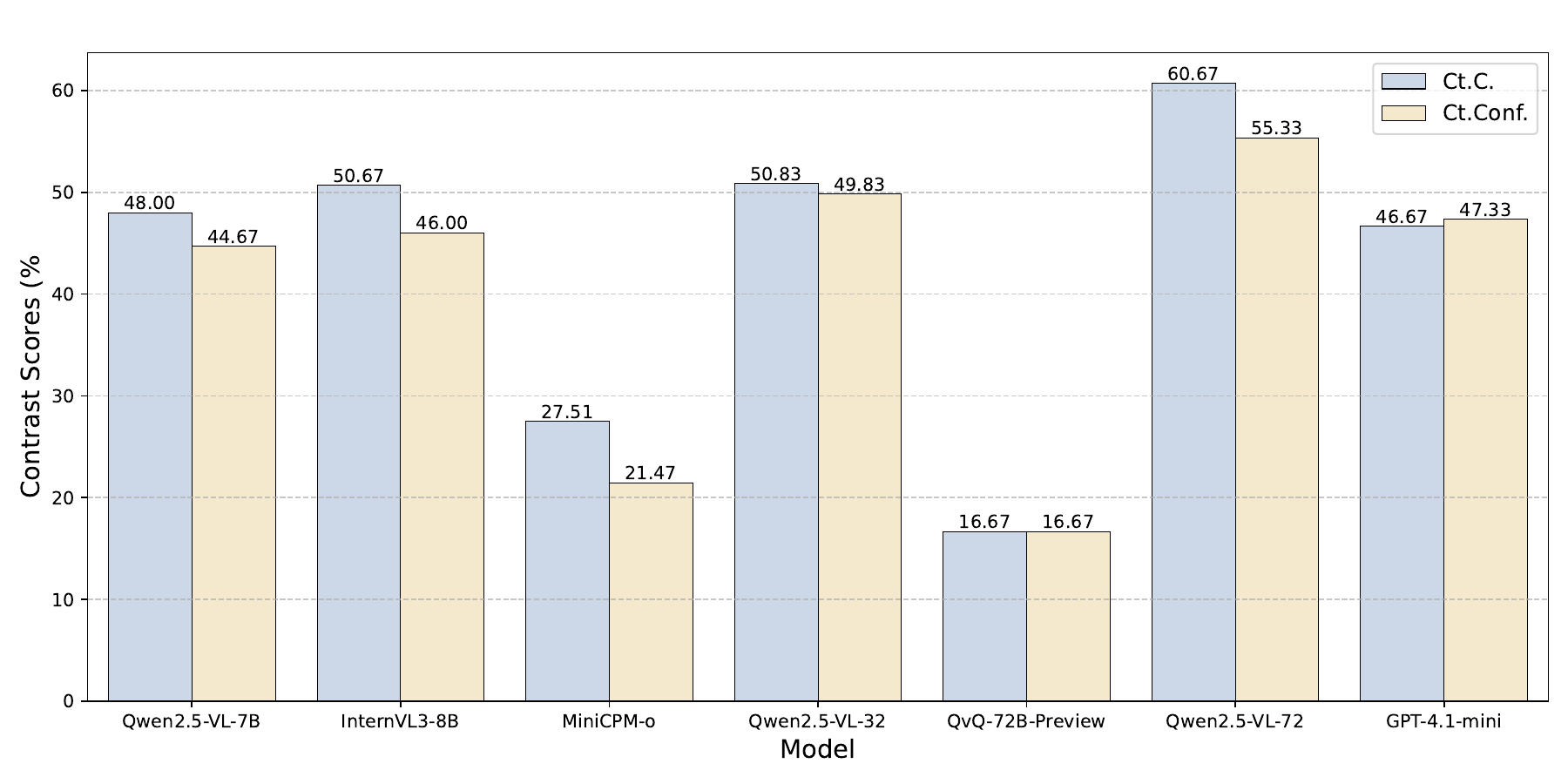}
 \caption{Geography}
 \label{fig：4}
 \end{subfigure}
 \hfill
 \begin{subfigure}[b]{0.32\textwidth}
 \centering
 \includegraphics[width=\linewidth]{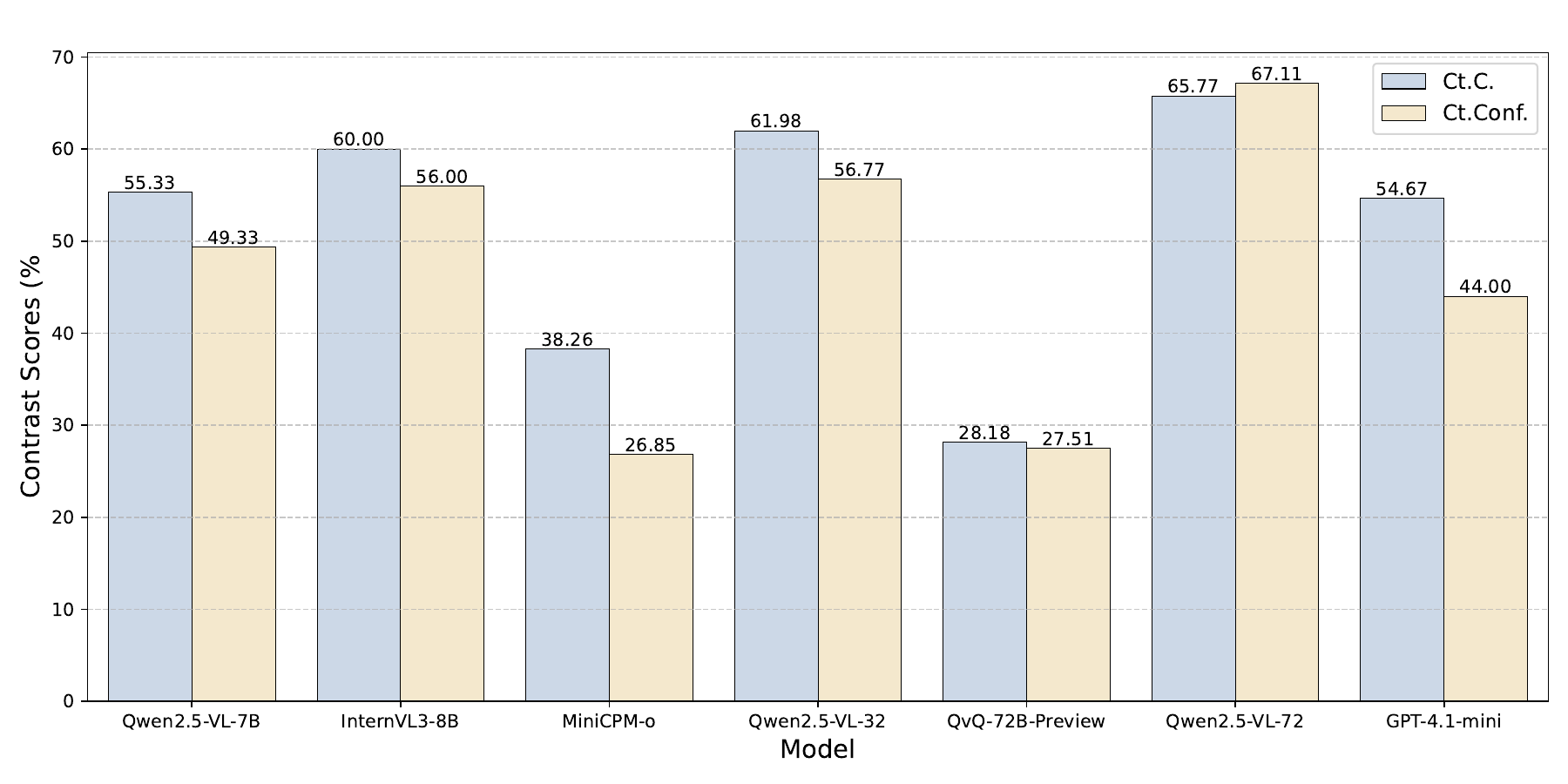}
 \caption{History}
 \label{fig：5}
 \end{subfigure}
 \hfill
 \begin{subfigure}[b]{0.32\textwidth}
 \centering
\includegraphics[width=\linewidth]{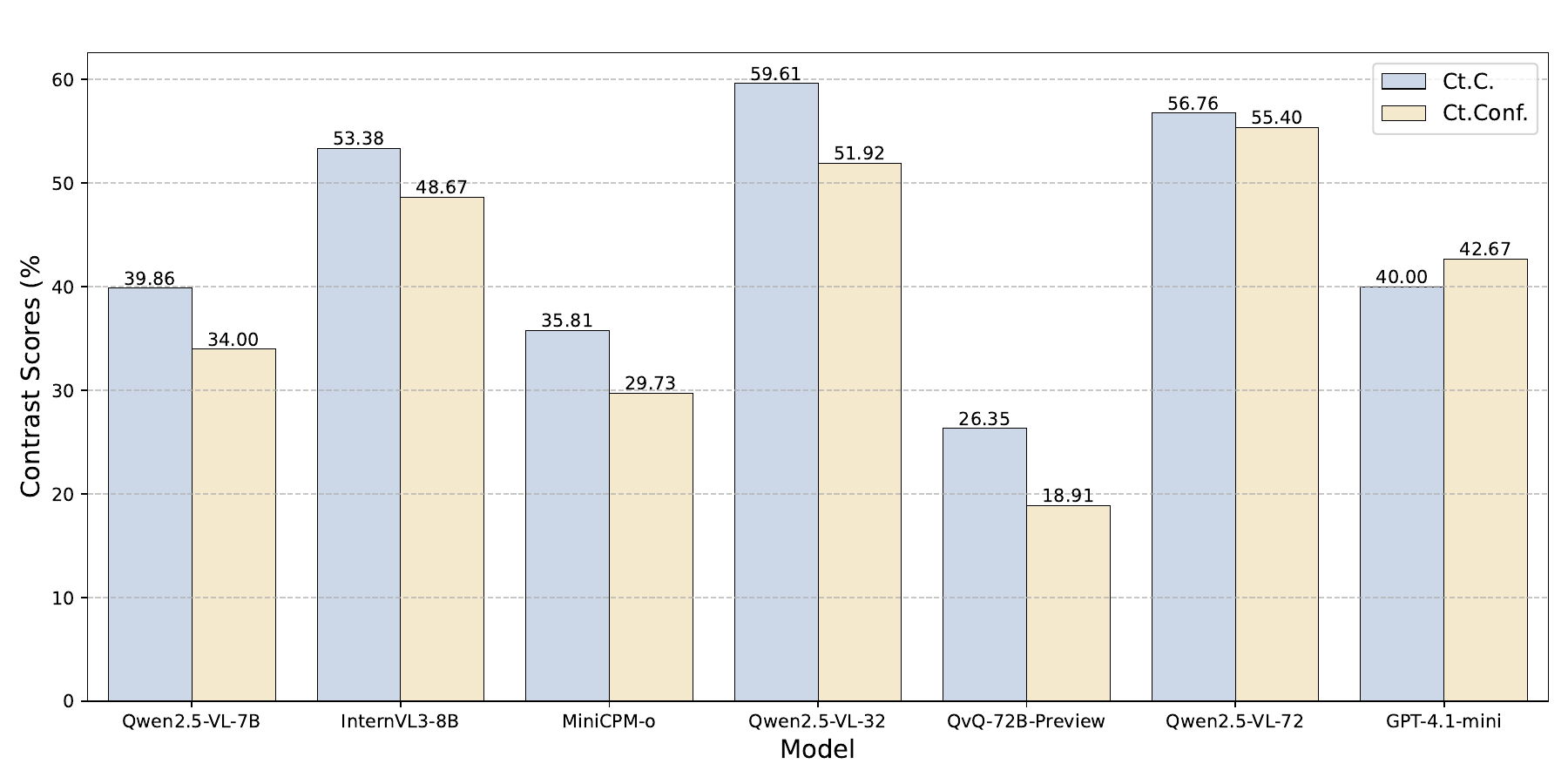}
 \caption{Social sciences}
 \label{fig：6}
 \end{subfigure}

 \caption{Contrast Scores (\%) with confounding option of MLLMs on HSSBench.}
 \label{fig：six_figs-table5}
\end{figure}

Table~\ref{apptab:question_contrast2} and Figure~\ref{fig：six_figs-table5} provides a detailed presentation of the experimental results after adding a confounding option. We sampled 900 data points. "Conf." indicates that, in addition to the given options, an extra option—"None of the above answers is correct"—was added. The model's output performance was then compared under these conditions.

\subsection{Visual Information Extraction Experiment Details}\label{sec.appendix.visual_extraction}

\begin{figure}[htbp]
 \centering
 \begin{subfigure}[b]{0.32\textwidth}
 \centering
 \includegraphics[width=\linewidth]{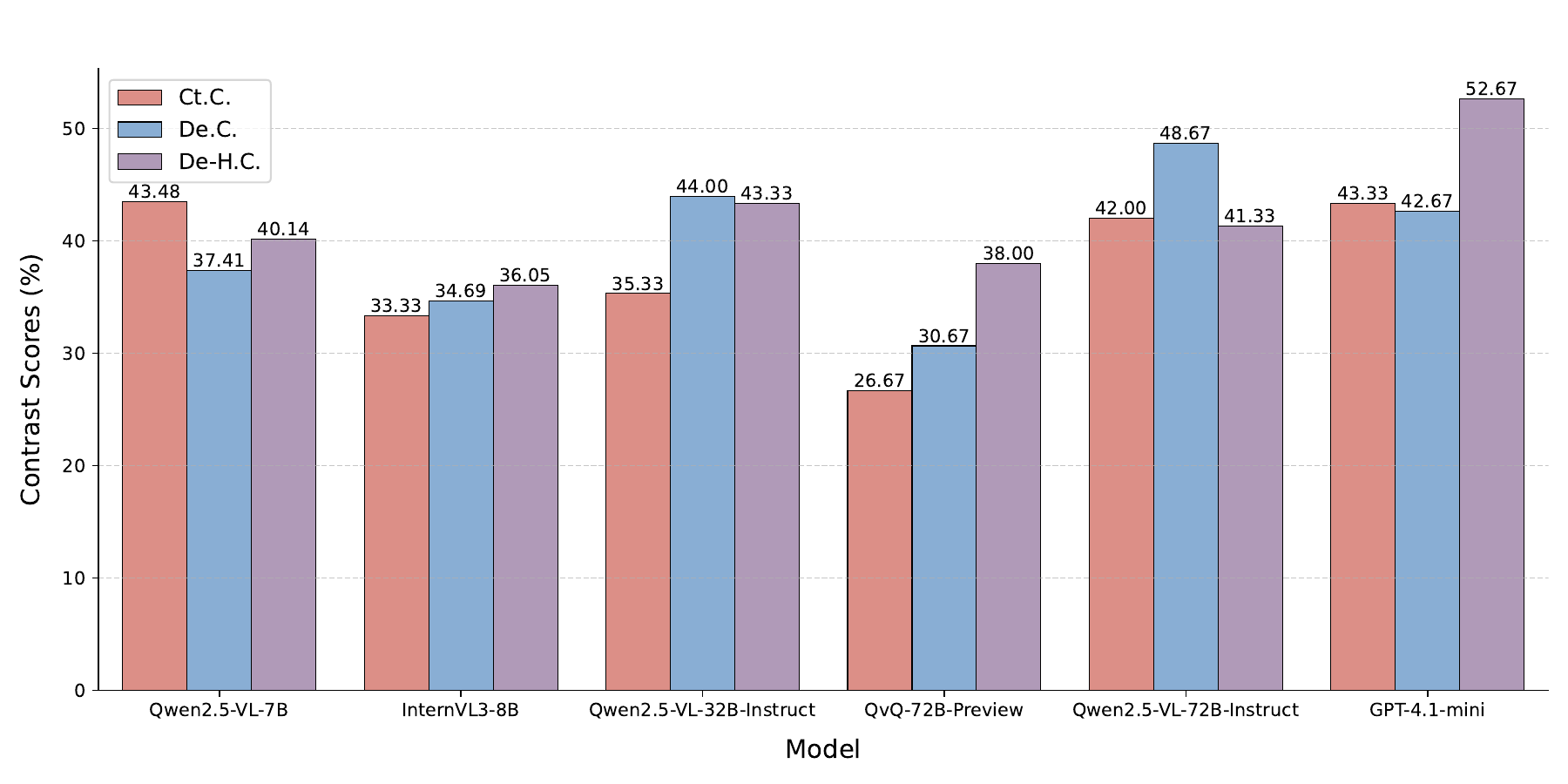}
 \caption{Art}
 \label{fig：1}
 \end{subfigure}
 \hfill
 \begin{subfigure}[b]{0.32\textwidth}
 \centering
 \includegraphics[width=\linewidth]{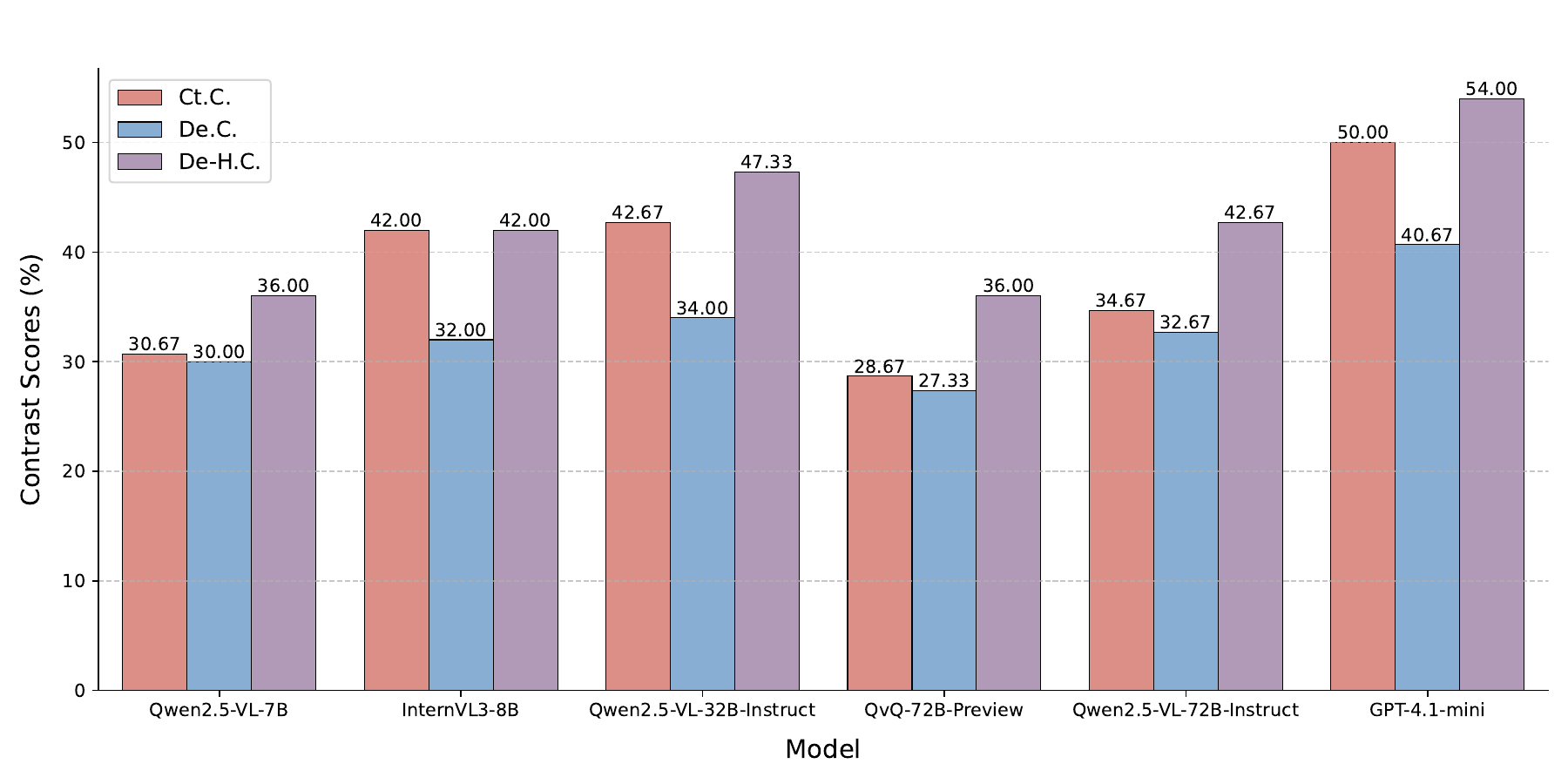}
 \caption{Culture}
 \label{fig：2}
 \end{subfigure}
 \hfill
 \begin{subfigure}[b]{0.32\textwidth}
 \centering
 \includegraphics[width=\linewidth]{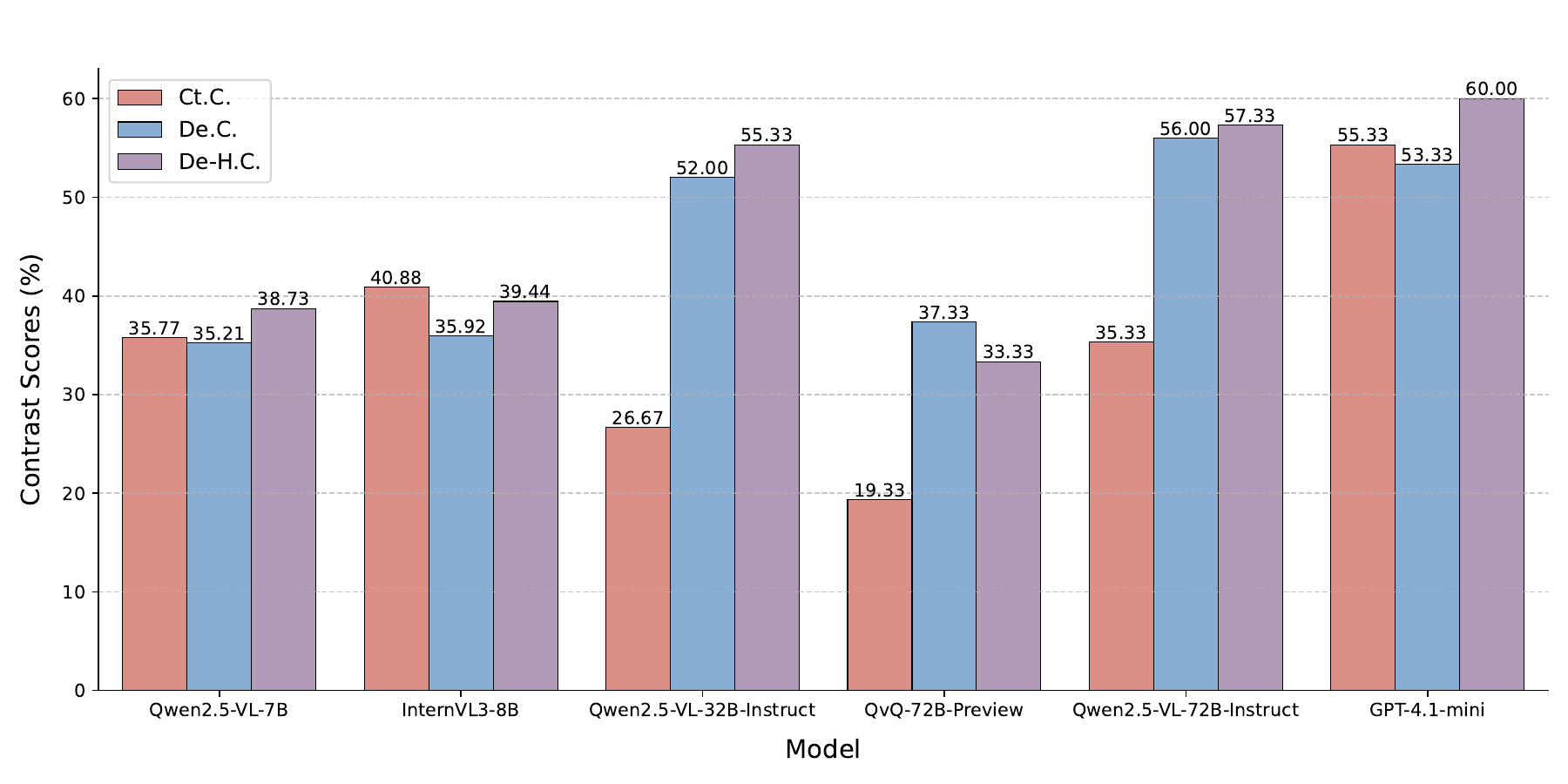}
 \caption{Economy}
 \label{fig：3}
 \end{subfigure}

 \vspace{1em} 

 \begin{subfigure}[b]{0.32\textwidth}
 \centering
\includegraphics[width=\linewidth]{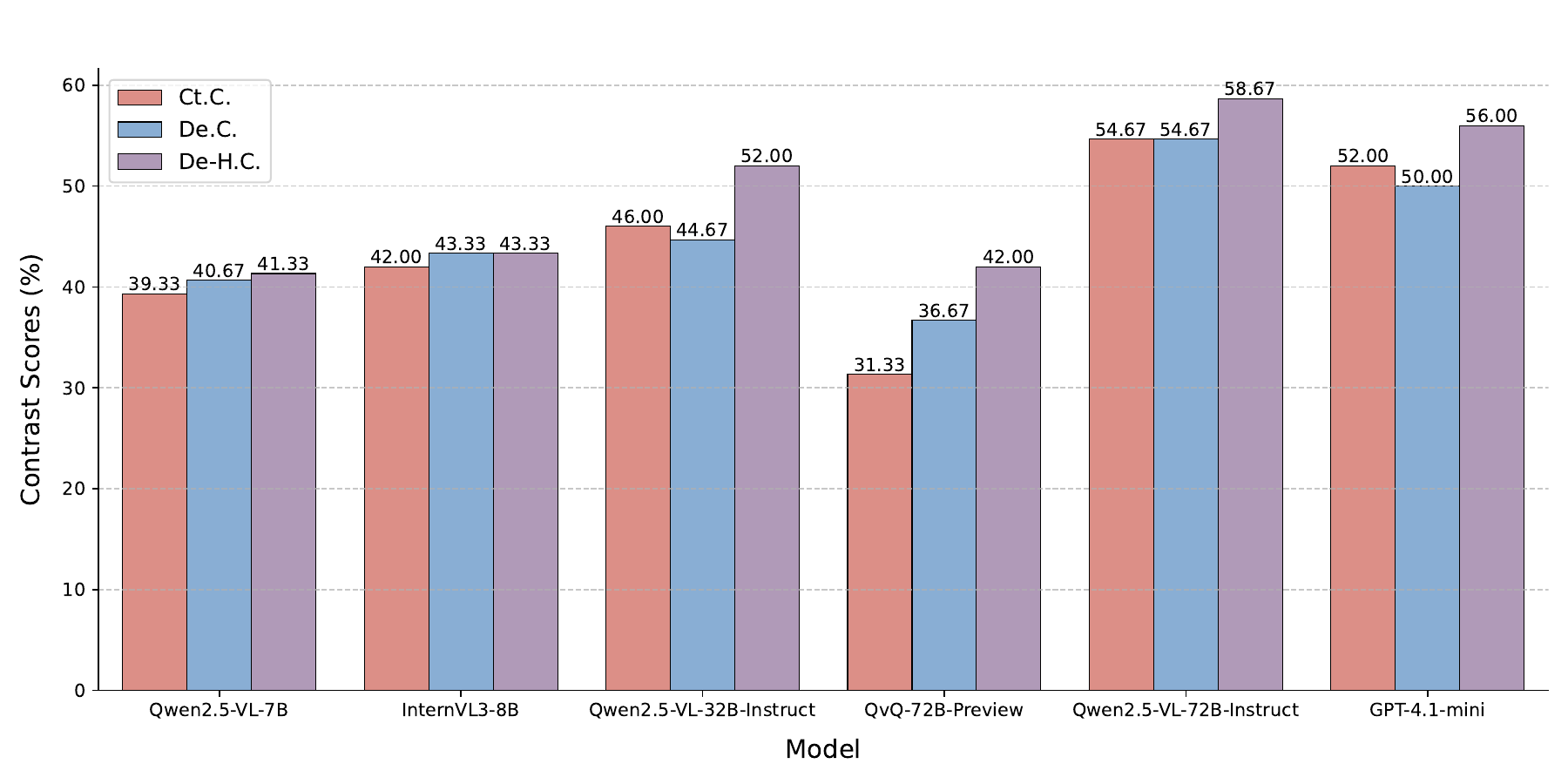}
 \caption{Geography}
 \label{fig：4}
 \end{subfigure}
 \hfill
 \begin{subfigure}[b]{0.32\textwidth}
 \centering
 \includegraphics[width=\linewidth]{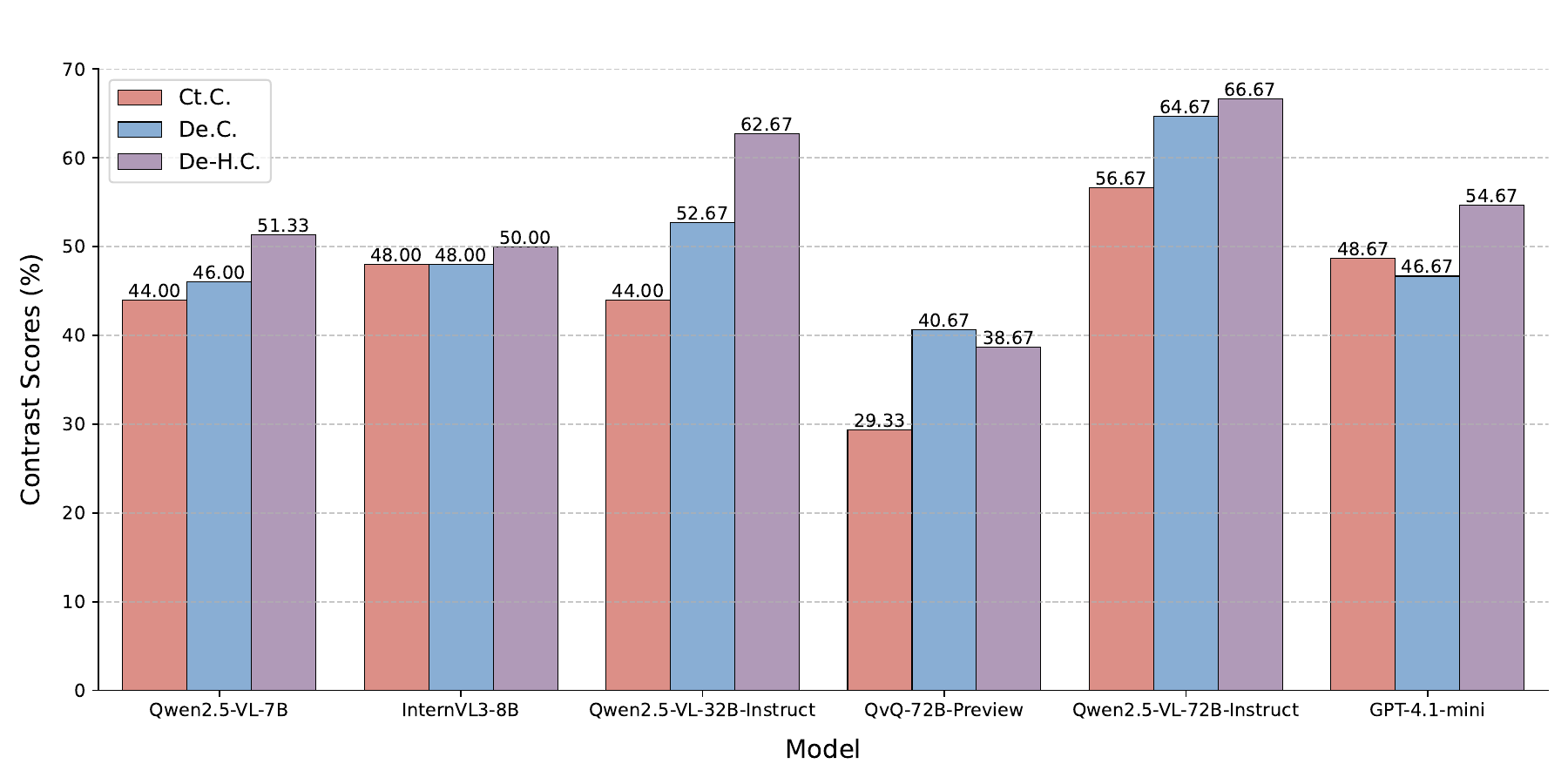}
 \caption{History}
 \label{fig：5}
 \end{subfigure}
 \hfill
 \begin{subfigure}[b]{0.32\textwidth}
 \centering
\includegraphics[width=\linewidth]{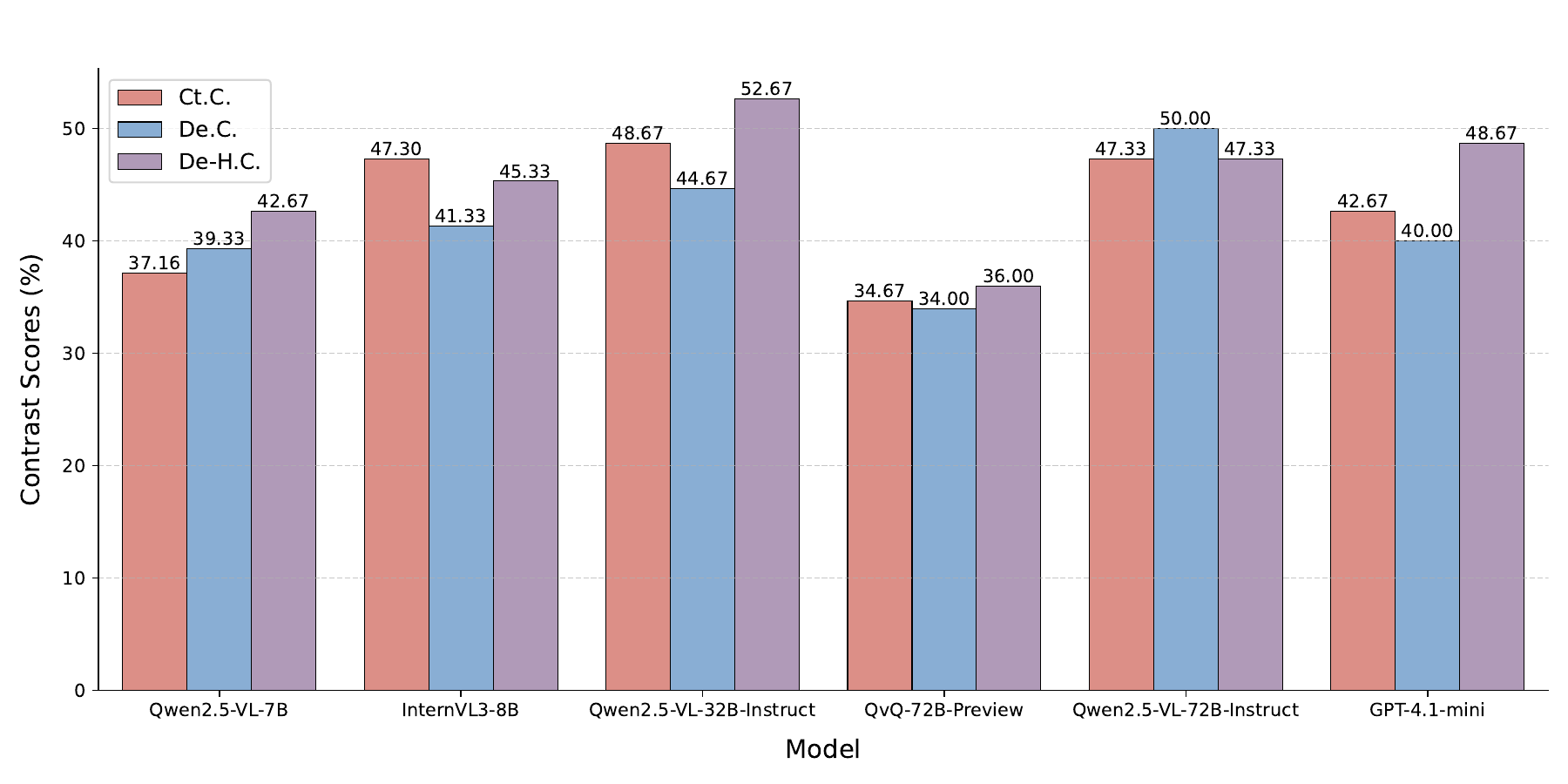}
 \caption{Social sciences}
 \label{fig：6}
 \end{subfigure}

 \caption{Contrast Scores (\%) about Visual Information Extraction of MLLMs on HSSBench.}
 \label{fig：six_figs-table6}
\end{figure}

Table~\ref{apptab:question_contrast} and Figure~\ref{fig：six_figs-table6} presents the extraction of image information into text (where "De." indicates a detailed description of the image generated directly by GPT-4.1, and "De-H." indicates a detailed explanation provided by a domain expert for each image).
The table below shows the detailed results of comparative experiments in which only the extracted image information, rather than the images themselves, was provided to the model.

\begin{table}[!ht]
    \centering
    \renewcommand{\arraystretch}{1.2}
\resizebox{\textwidth}{!}{
    \begin{tabular}{c|ccc|ccc|ccc|ccc|ccc|ccc}
    \toprule
    \multirow{2}{*}{Model} 
    & \multicolumn{3}{c|}{Geography} 
    & \multicolumn{3}{c|}{Art} 
    & \multicolumn{3}{c|}{Culture} 
    & \multicolumn{3}{c|}{Social Sciences} 
    & \multicolumn{3}{c|}{History} 
    & \multicolumn{3}{c}{Economy}  \\
    \cmidrule(lr){2-4} \cmidrule(lr){5-7} \cmidrule(lr){8-10} \cmidrule(lr){11-13} \cmidrule(lr){14-16} \cmidrule(lr){17-19} 
    & Ct.C. & De.C. & De-H.C. & Ct.C. & De.C. & De-H.C. & Ct.C. & De.C. & De-H.C. &Ct.C. & De.C. & De-H.C.& Ct.C. & De.C. & De-H.C.&Ct.C. & De.C. & De-H.C.\\
    \midrule
    Qwen2.5-VL-7B & 39.33 & 40.67 & 41.33 & 43.48 & 37.41 & 40.14 & 30.67 & 30.00 & 36.00 & 37.16 & 39.33 & 42.67 & 44.00 & 46.00 & 51.33 & 35.77 & 35.21 & 38.73 \\
    InternVL3-8B & 42.00 & 43.33 & 43.33 & 33.33 & 34.69 & 36.05 & 42.00 & 32.00 & 42.00 & 47.30 & 41.33 & 45.33 & 48.00 & 48.00 & 50.00 & 40.88 & 35.92 & 39.44 \\
    Qwen2.5-VL-32B-Instruct & 46.00 & 44.67 & 52.00 & 35.33 & 44.00 & 43.33 & 42.67 & 34.00 & 47.33 & 48.67 & 44.67 & 52.67 & 44.00 & 52.67 & 62.67 & 26.67 & 52.00 & 55.33 \\
    QvQ-72B-Preview & 31.33 & 36.67 & 42.00 & 26.67 & 30.67 & 38.00 & 28.67 & 27.33 & 36.00 & 34.67 & 34.00 & 36.00 & 29.33 & 40.67 & 38.67 & 19.33 & 37.33 & 33.33 \\
    Qwen2.5-VL-72B-Instruct & 54.67 & 54.67 & 58.67 & 42.00 & 48.67 & 41.33 & 34.67 & 32.67 & 42.67 & 47.33 & 50.00 & 47.33 & 56.67 & 64.67 & 66.67 & 35.33 & 56.00 & 57.33 \\
    GPT-4.1-mini &52.00 & 50.00 & 56.00 &  43.33& 42.67 & 52.67 & 50.00 &40.67   & 54.00 &42.67  &  40.00 &  48.67 &48.67 &46.67  &  54.67& 55.33  & 53.33 & 60.00 \\
    \bottomrule
    \end{tabular}
}
    \caption{Contrast Scores (\%) about Visual Information Extraction of MLLMs on HSSBench.}
    \label{apptab:question_contrast}
\end{table}

\subsection{Comparison with HSS-Related Benchmarks}\label{sec.appendix.related.Benchmark}

We acknowledge that some existing benchmarks, such as MME, include HSS-related test data, particularly within the Art domain. To better understand the relationship between these datasets and our proposed HSSBench, we conducted a detailed comparative analysis focusing on the overlapping Art category. The evaluation was performed under identical prompt settings, including both Direct and Chain-of-Thought (CoT) prompting. The combined results are presented in Table~\ref{tab:hssbench_comparison}.

\begin{table}[h]
  \centering  
  \begin{tabular}{lcccccc}
    \toprule
    \multirow{2}{*}{Model} & \multicolumn{4}{c}{Benchmark (\% Accuracy)} & \multicolumn{2}{c}{Prompting} \\
    \cmidrule(r){2-5} \cmidrule(l){6-7}
                           & CMMMU & MME & MMMU & HSSBench & Direct & CoT \\
    \midrule
    Qwen2.5-VL-3B-Instruct & 47.73 & 77.32 & 57.76 & 29.01 & 47.73 & 48.86 \\
    Qwen2.5-VL-7B-Instruct & 43.18 & 71.21 & 61.21 & 37.88 & 43.18 & 46.59 \\
    InternVL3-8B           & 65.91 & 85.28 & 68.10 & 42.14 & 65.91 & 55.68 \\
    llava-onevision-qwen2-7b & 53.41 & 77.21 & 49.14 & 36.20 & 53.41 & 52.27 \\
    \bottomrule
  \end{tabular}
  \caption{Performance comparison on the Art category across HSS-related benchmarks under Direct and CoT prompting.}
  \label{tab:hssbench_comparison}
\end{table}

As shown, the relative performance trends across these benchmarks are broadly consistent, indicating that the challenges within the Art domain are similar across datasets. Notably, HSSBench presents a more challenging evaluation, reflected by generally lower accuracy scores, which underscores its value in pushing forward research on Humanities and Social Sciences tasks.

\subsection{Comparison with STEM Benchmarks}\label{sec.appendix.STEM.Benchmark}

While our primary focus is on Humanities and Social Sciences (HSS), we also recognize the importance of situating HSSBench within the broader landscape of STEM benchmarks. To this end, we provide a comparative overview of model performance on several representative STEM benchmarks alongside HSSBench under the Chain-of-Thought (CoT) prompting setting. The results are summarized in Table~\ref{tab:stem_hss_comparison}.

\begin{table}[h]
  \centering
  \setlength{\tabcolsep}{2pt}
  \small
  
  \begin{tabular}{lccccccccc}
       \toprule
       Model & \makecell{MMLU \\ Pro} & \makecell{GPQA \\ Diamond} & \makecell{SWE-bench \\ Verified} & \makecell{MATH-\\500} & \makecell{AIME \\2024} & \makecell{LiveCode\\Bench} & \makecell{OpenCompass- \\ Reasoning} & \textbf{HSSBench} \\
       \midrule
    Qwen2.5-VL-72B      & 71.2     & —            & —                  & —        & —         & —             & 50.2                  & 51.87             \\
    GPT-4.1 mini        & —        & 65.00        & 23.60              & —        & 49.60     & —             & 46.0                  & 48.03             \\
    GPT-4o              & 79.80    & 66.90        & —                  & —        & —         & 35.80         & 54.8                  & 46.88             \\
    GPT-4.1             & 81.80    & 66.30        & 54.6                & 92.80    & 48.10     & 40.50         & 54.0                  & 42.66             \\
    InternVL3-8B        & —        & —            & —                  & —        & —         & —             & 41.4                  & 41.42             \\
    GPT-4.1 nano        & —        & 50.30        & —                  & —        & 29.40     & —             & 34.2                  & 35.83             \\
    Janus-Pro-7B        & —        & —            & —                  & —        & —         & —             & 19.1                  & 31.66             \\
    \bottomrule
  \end{tabular}
  \caption{Model performance comparison on STEM benchmarks and HSSBench (CoT prompting).}
  \label{tab:stem_hss_comparison}
  \setlength{\tabcolsep}{6pt} 
  \renewcommand{\arraystretch}{1} 
\end{table}

\noindent\textbf{Notes:}  
All STEM benchmark and OpenCompass results are sourced from official publications or model release notes. A dash (—) indicates that the model did not publicly report results for that benchmark. Minor discrepancies may exist due to testing variability but do not affect the overall trend.

This comparison reveals that while models generally achieve higher accuracy on STEM benchmarks (e.g., MMLU Pro, MATH-500), their performance on HSSBench is comparatively lower. This gap highlights the unique challenges posed by HSS tasks and the necessity of dedicated benchmarks like HSSBench to drive progress in this domain.

\subsection{Retrieval-Augmented Generation Evaluation}\label{sec.appendix.detail.RAG}
We conducted additional experiments integrating Retrieval-Augmented Generation (RAG) with several smaller MLLMs. The retrieval database was constructed from a general knowledge corpus comprising Wikipedia and publicly available documents related to HSS. Model performance was evaluated under two prompting strategies: direct prompting and CoT prompting.

\begin{table}[htbp]
\centering
\setlength{\tabcolsep}{5pt}
\small 
\caption{Performance of models under direct prompting.}
\label{tab:direct_prompting}
\begin{tabular}{l|*{5}{c}c|c}
\toprule
Model & Geography & Art & Culture & \makecell{Social\\Science} & History & Economy & Overall \\
\midrule
\multicolumn{8}{c}{\textit{Without RAG}} \\
\midrule
Qwen2.5-VL-7B-Instruct & 60.09 & 28.05 & 27.19 & 53.50 & 54.51 & 40.09 & 43.89 \\
Qwen2.5-VL-3B-Instruct & 55.40 & 33.48 & 44.24 & 54.50 & 50.82 & 35.59 & 45.56 \\
InternVL3-8B & 54.93 & 37.56 & 43.78 & 62.00 & 51.64 & 44.14 & 48.82 \\
llava-onevision-qwen2-7b & 46.01 & 30.77 & 34.10 & 46.50 & 38.93 & 29.28 & 37.43 \\
\midrule
\multicolumn{8}{c}{\textit{With RAG}} \\
\midrule
Qwen2.5-VL-7B-Instruct & 47.42 & 29.86 & 29.95 & 51.00 & 45.08 & 38.74 & 40.24 \\
Qwen2.5-VL-3B-Instruct & 49.77 & 33.03 & 35.48 & 48.50 & 43.44 & 33.78 & 40.55 \\
InternVL3-8B & 53.52 & 35.75 & 35.94 & 59.50 & 49.59 & 37.39 & 45.10 \\
llava-onevision-qwen2-7b & 43.19 & 32.13 & 26.73 & 41.00 & 32.38 & 21.17 & 32.57 \\
\bottomrule
\end{tabular}
\end{table}

\vspace{1em}

\begin{table}[htbp]
\centering
\setlength{\tabcolsep}{5pt}
\small 
\caption{Performance of models under COT prompting.}
\label{tab:COT_prompting}
\begin{tabular}{l|*{5}{c}c|c}
\toprule
Model & Geography & Art & Culture & \makecell{Social\\Science} & History & Economy & Overall \\
\midrule
\multicolumn{8}{c}{\textit{Without RAG}} \\
\midrule
Qwen2.5-VL-7B-Instruct & 55.40 & 25.79 & 27.65 & 51.00 & 47.54 & 39.64 & 41.08 \\
Qwen2.5-VL-3B-Instruct & 51.17 & 30.32 & 35.48 & 49.00 & 46.31 & 32.43 & 40.70 \\
InternVL3-8B & 53.52 & 34.39 & 40.55 & 57.00 & 51.64 & 38.74 & 45.86 \\
llava-onevision-qwen2-7b & 43.66 & 31.22 & 28.57 & 43.00 & 33.61 & 29.73 & 34.78 \\
\midrule
\multicolumn{8}{c}{\textit{With RAG}} \\
\midrule
Qwen2.5-VL-7B-Instruct & 47.89 & 27.15 & 27.19 & 46.50 & 42.62 & 36.49 & 37.89 \\
Qwen2.5-VL-3B-Instruct & 41.78 & 31.22 & 29.95 & 44.00 & 40.57 & 36.04 & 37.21 \\
InternVL3-8B & 57.75 & 30.77 & 34.10 & 55.50 & 50.41 & 39.19 & 44.50 \\
llava-onevision-qwen2-7b & 42.25 & 29.86 & 28.11 & 42.50 & 31.56 & 23.42 & 32.73 \\
\bottomrule
\end{tabular}
\end{table}

The results indicate that, although RAG occasionally yields modest improvements in specific domains or models, it does not consistently outperform direct prompting without retrieval augmentation. This suggests that augmenting MLLMs with a general retrieval corpus and straightforward prompting strategies may be insufficient to fully exploit the complex and nuanced knowledge required for HSS tasks. 

These findings highlight the challenges inherent in applying RAG to Humanities and Social Sciences benchmarks such as HSSBench. We hypothesize that more specialized, domain-specific retrieval corpora, combined with advanced retrieval and integration techniques, are necessary to unlock the full potential of RAG in this context.

\subsection{Prompt for Model Inference}\label{sec.appendix.detail.prompt}

Table~\ref{tab:prompt_model_inference} details the configurations of the four prompts employed in our experiments, specifying the presence or absence of a CoT prompt and indicating whether the questions were open-ended or multiple-choice.
\begin{table*}[ht]
\centering
\small
\begin{tabularx}{\linewidth}{m{1cm}<{\centering}m{\dimexpr\textwidth-1cm-4\tabcolsep}}
\toprule
\multicolumn{2}{c}{\textbf{Prompt for Model Inference}} \\\midrule
\makecell{w/ MC \\ w/ CoT} &
Question: [question]

Options: [options]

Think step by step to determine the correct answer. 

End your response with [[X]] where X is your final answer (A, B, C, D or E).
 \\\midrule
\makecell{w/ MC \\ w/o CoT} &
Question: [question]

Options: [options]

Give the correct answer directly. 

End your response with [[X]] where X is your final answer (A, B, C, D or E).
 \\\midrule
\makecell{w/o MC \\ w/ CoT} &
Question: [question]

Think step by step to determine the correct answer.

End your response with [[X]] where X is your final answer.
 \\\midrule
\makecell{w/o MC \\ w/o CoT} &
Question: [question]

Give the correct answer directly. 

End your response with [[X]] where X is your final answer. \\\bottomrule
\end{tabularx}
\caption{Prompt for model inference.}
\label{tab:prompt_model_inference}
\end{table*}

\subsection{Evaluation for model's output}\label{sec.appendix.detail.validation}
Table~\ref{tab:prompt_model_evaluation} presents the evaluation prompts employed to assess the accuracy of the model's responses.
\begin{table*}[ht]
\centering
\small
\begin{tabularx}{\linewidth}{m{1cm}<{\centering}m{\dimexpr\textwidth-1cm-4\tabcolsep}}
\toprule
\multicolumn{2}{c}{\textbf{Prompt for Model Evaluation}} \\\midrule
\makecell{w/ MC} &
You are an evaluation assistant. 

Please determine whether the answers output by the model below are correct.

Question: [question]

Options: [options]

Correct answer: [correct answer]

Model output content: [model output]

Please extract its final answer from the model output and determine whether it is consistent with the content of the correct answer. 

If the answer is correct, reply with "1". Otherwise, reply with "0".

Only reply with numbers and no superfluous content.
 \\\midrule
\makecell{w/o MC} &
You are an evaluation assistant. 

Please determine whether the answers output by the model below are correct.

Question: [question]

Correct answer: [correct answer]

Model output content: [model output]

Please extract its final answer from the model output and determine whether it is consistent with the content of the correct answer. 
Attention! If the meanings are roughly consistent, it can be regarded as correct.
It is not necessary to be strictly and completely consistent.
One needs to carefully consider whether the model's response is reasonable and correct.

If the answer is correct, reply with "1". Otherwise, reply with "0".

Only reply with numbers and no superfluous content.
 \\\bottomrule
\end{tabularx}
\caption{Prompt for model evaluation.}
\label{tab:prompt_model_evaluation}
\end{table*}

In our evaluation process, we employed both expert sampling and automated assessment using GPT models. To verify the consistency of the evaluation results produced by the models, we randomly sampled 2,000 data points from the overall experiment, including 1,000 multiple-choice questions and 1,000 open-ended questions. These samples were independently evaluated by three human experts, as well as by Qwen2.5-7B and GPT-4.1.

For both multiple-choice and open-ended questions, the overall agreement among the two models and the three experts, measured by Fleiss' Kappa, was 96.82\% and 94.79\%, respectively. The agreement between the two models alone, measured by Cohen's Kappa, was 97.32\% for multiple-choice questions and 94.79\% for open-ended questions.

These results demonstrate that the consistency between the three experts and the two models remains at a reliable level. This indicates that either evaluation method yields results that are largely consistent with those reported in the paper, thereby ensuring the reproducibility of our findings.

\section{More Cases}\label{sec.appendix.case}
\clearpage
\newpage
\begin{figure}
    \centering
    \includegraphics[width=1\linewidth]{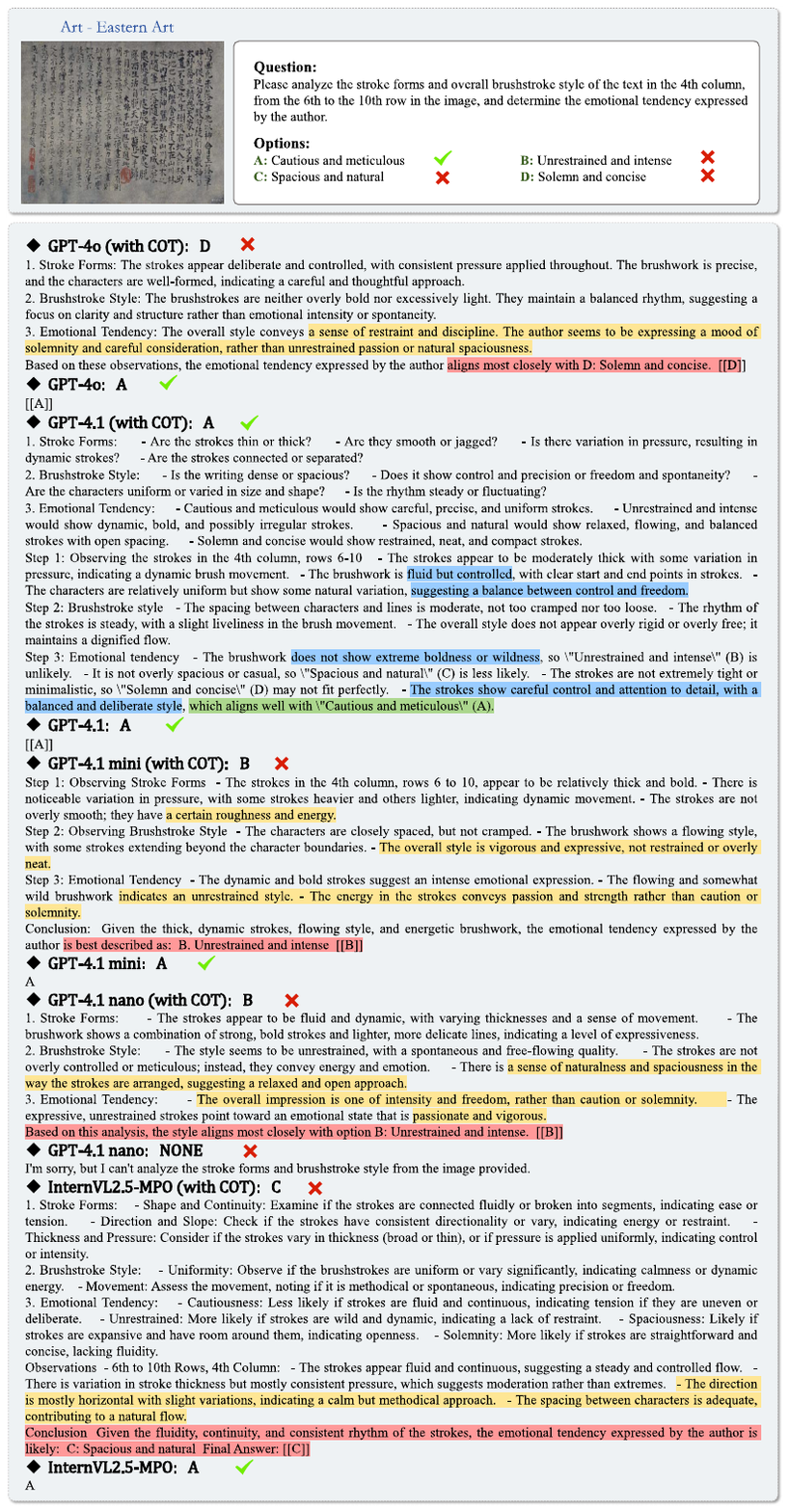}
    \label{fig:enter-label}
\end{figure}

\begin{figure}
    \centering
    \includegraphics[width=1\linewidth]{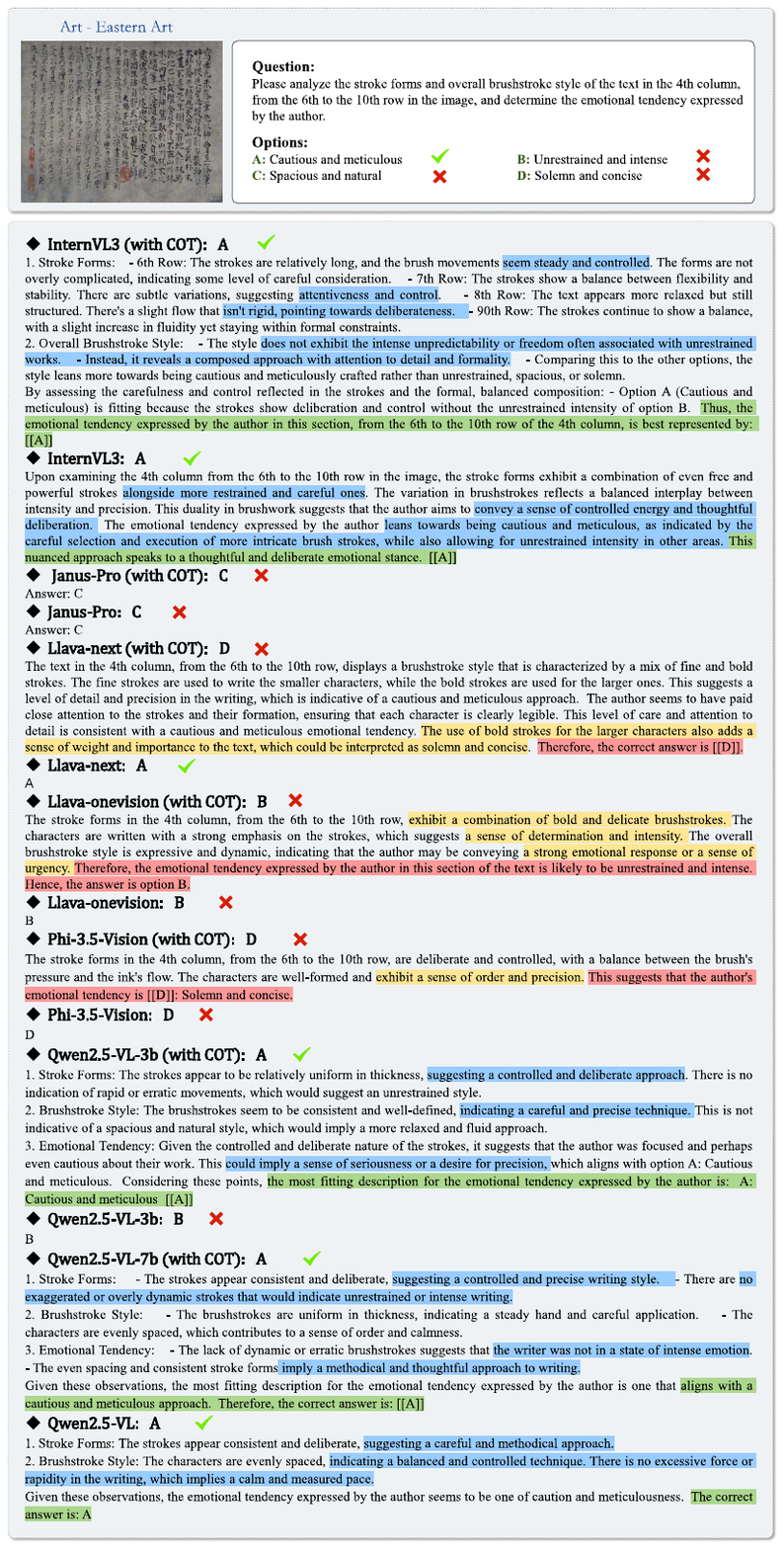}
\end{figure}

\begin{figure}
    \centering
    \includegraphics[width=1\linewidth]{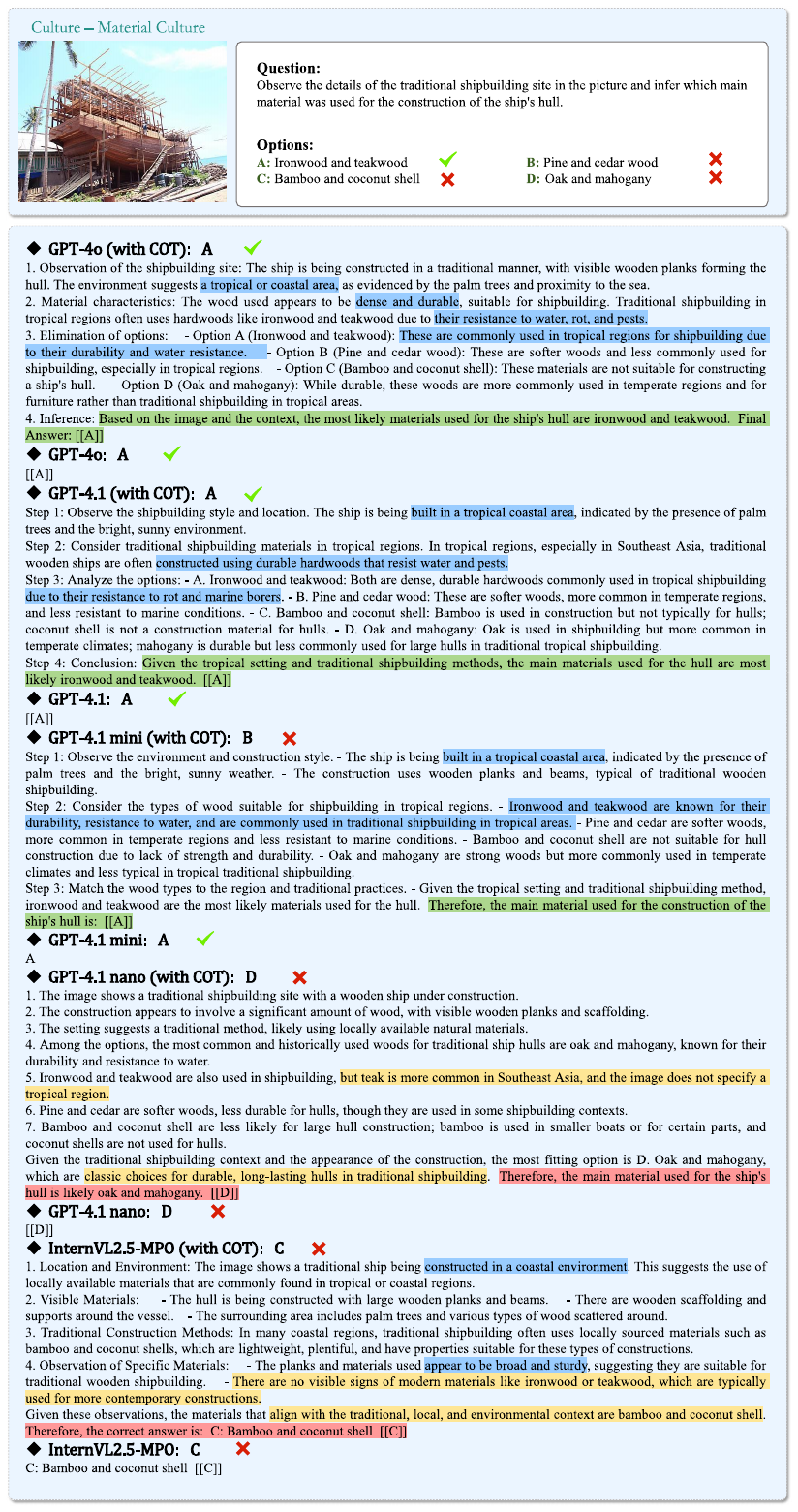}

\end{figure}
\begin{figure}
    \centering
    \includegraphics[width=1\linewidth]{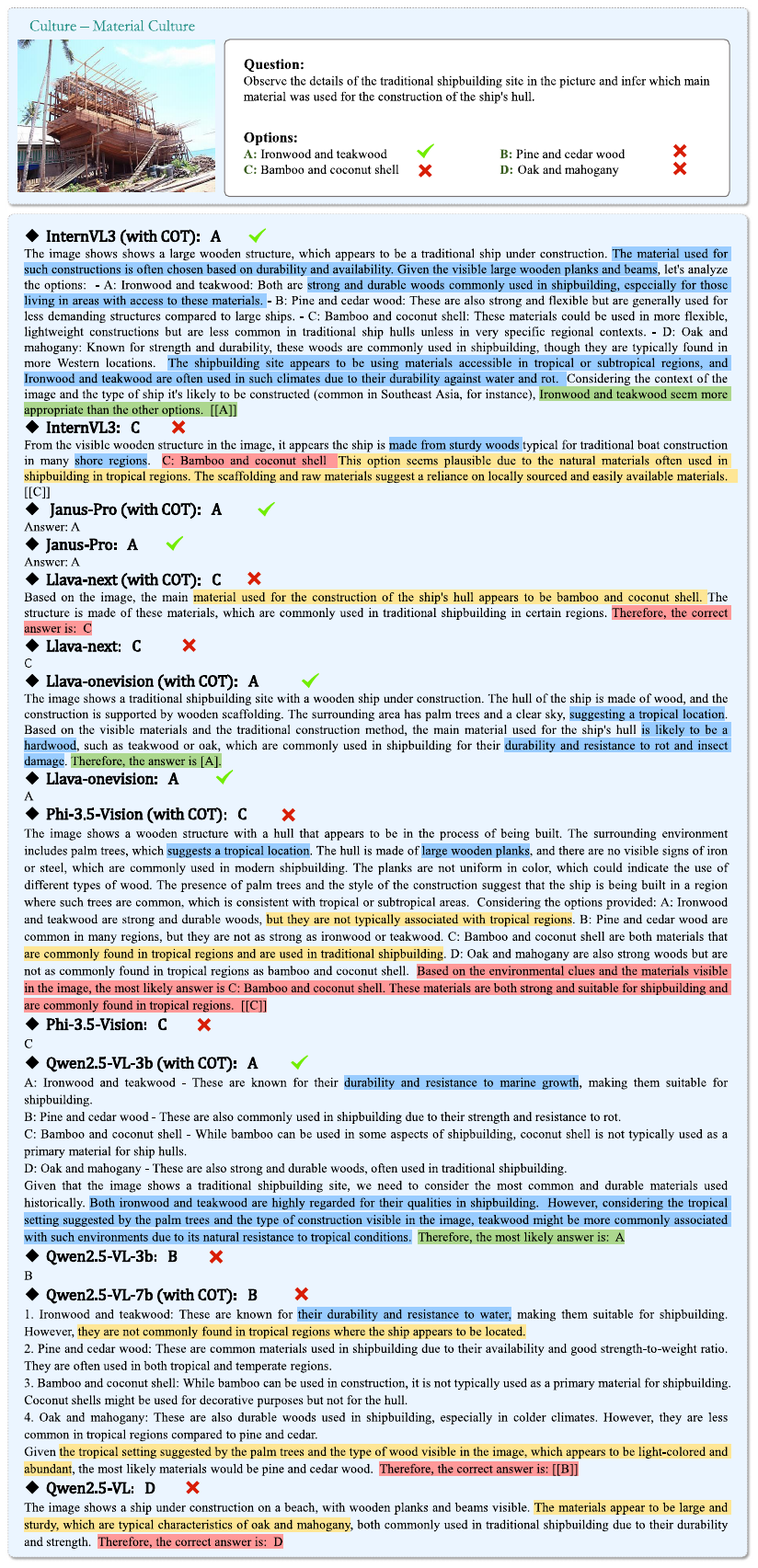}
\end{figure}

\begin{figure}
    \centering
    \includegraphics[width=1\linewidth]{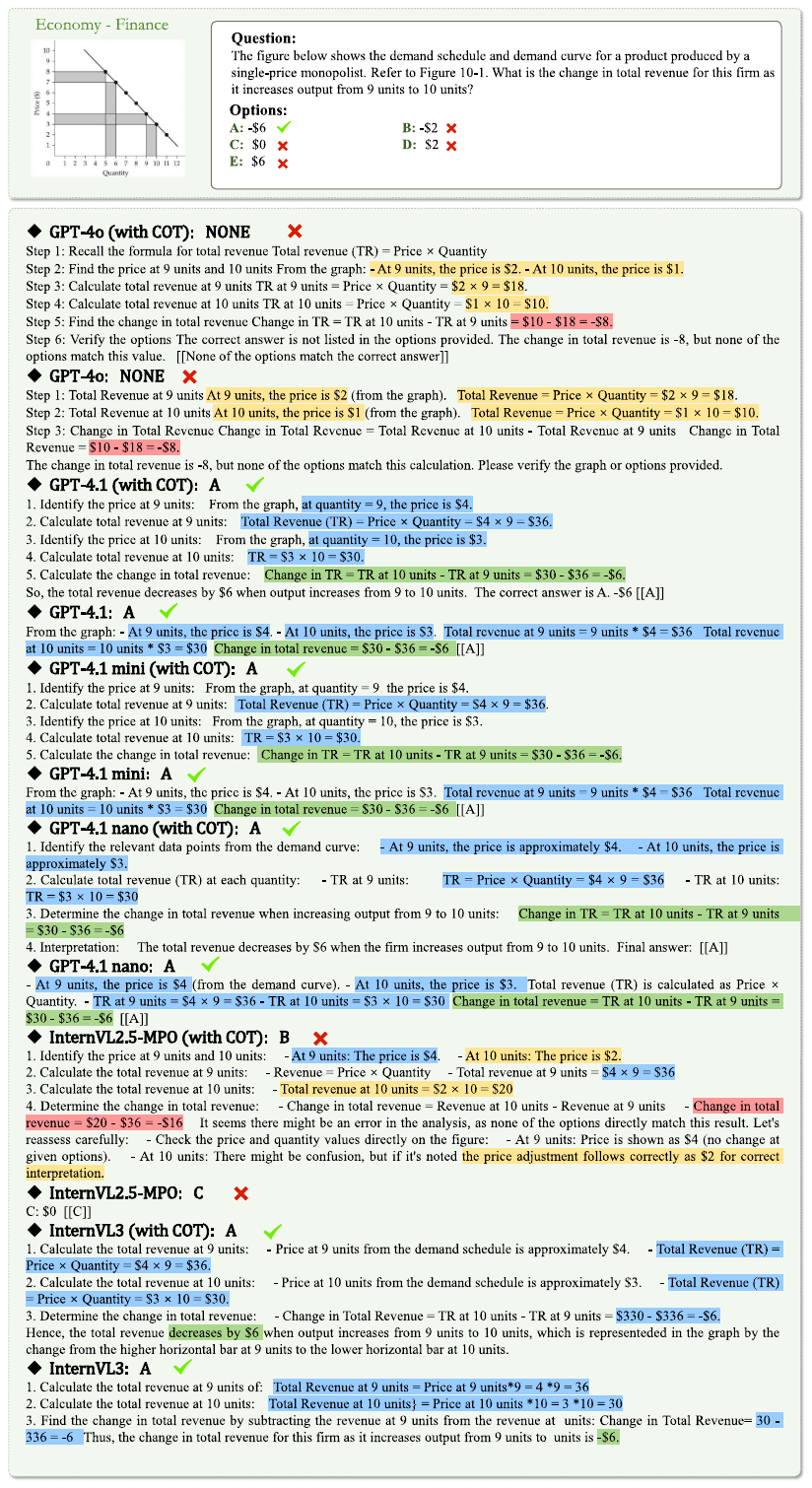}
\end{figure}
\begin{figure}
    \centering
    \includegraphics[width=1\linewidth]{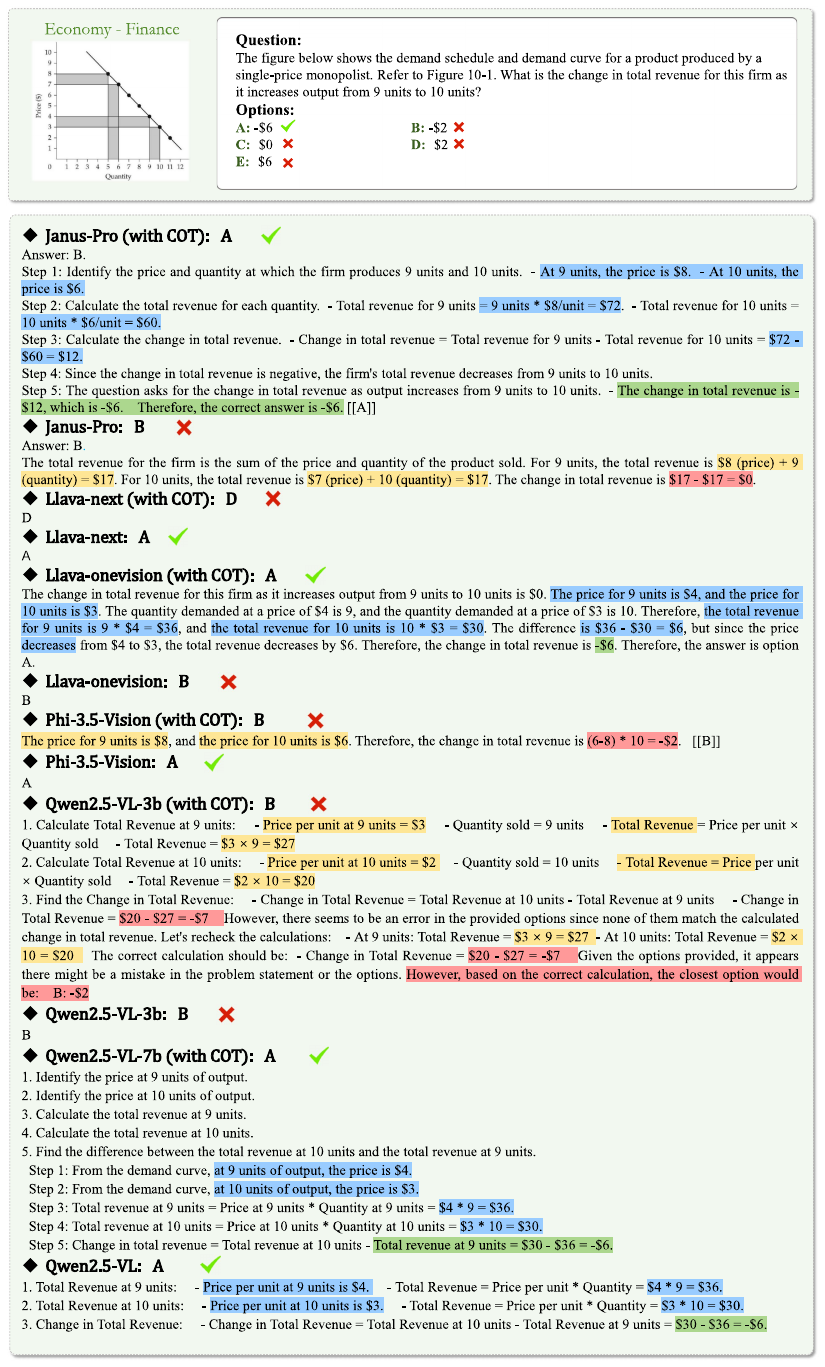}
\end{figure}

\begin{figure}
    \centering
    \includegraphics[width=1\linewidth]{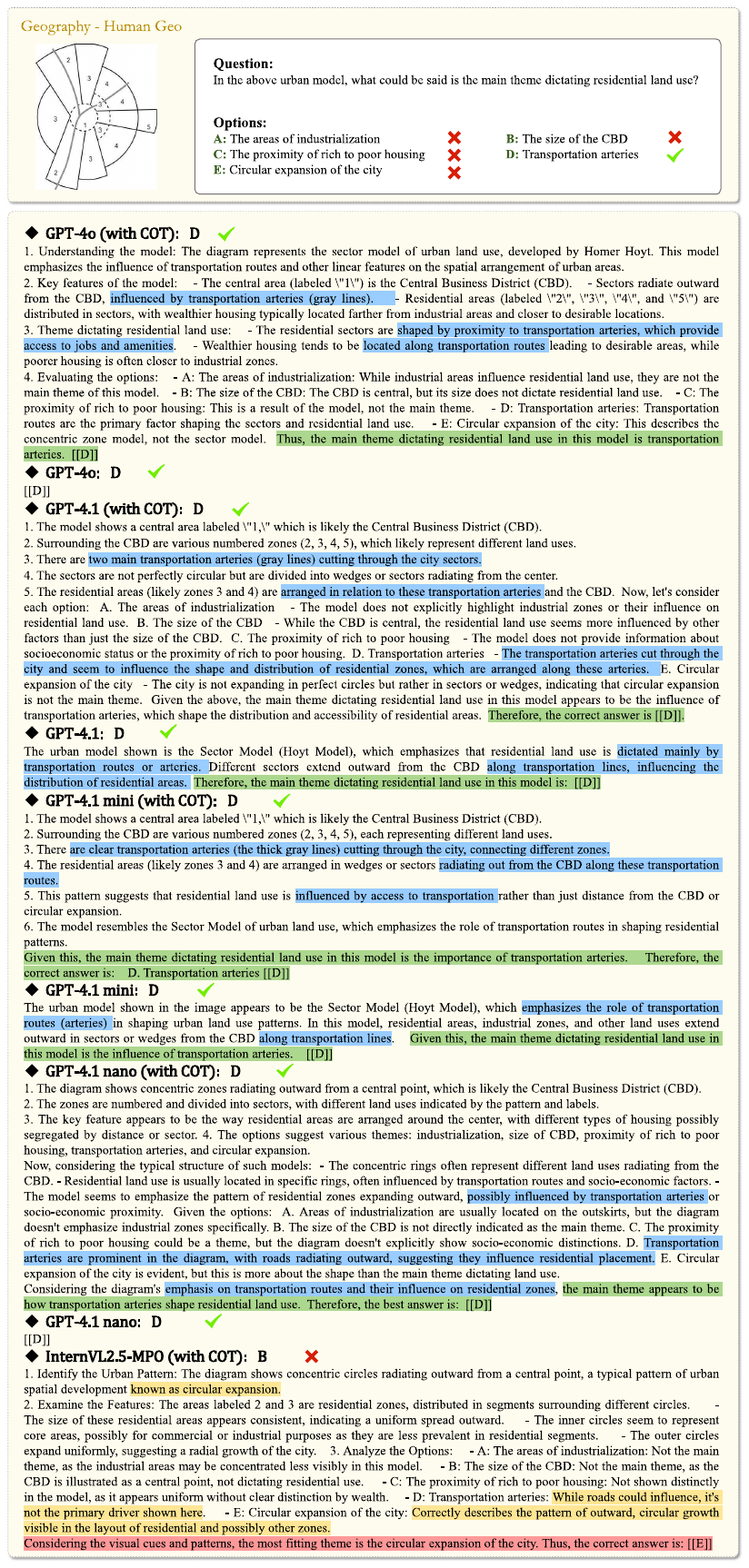}
\end{figure}
\begin{figure}
    \centering
    \includegraphics[width=1\linewidth]{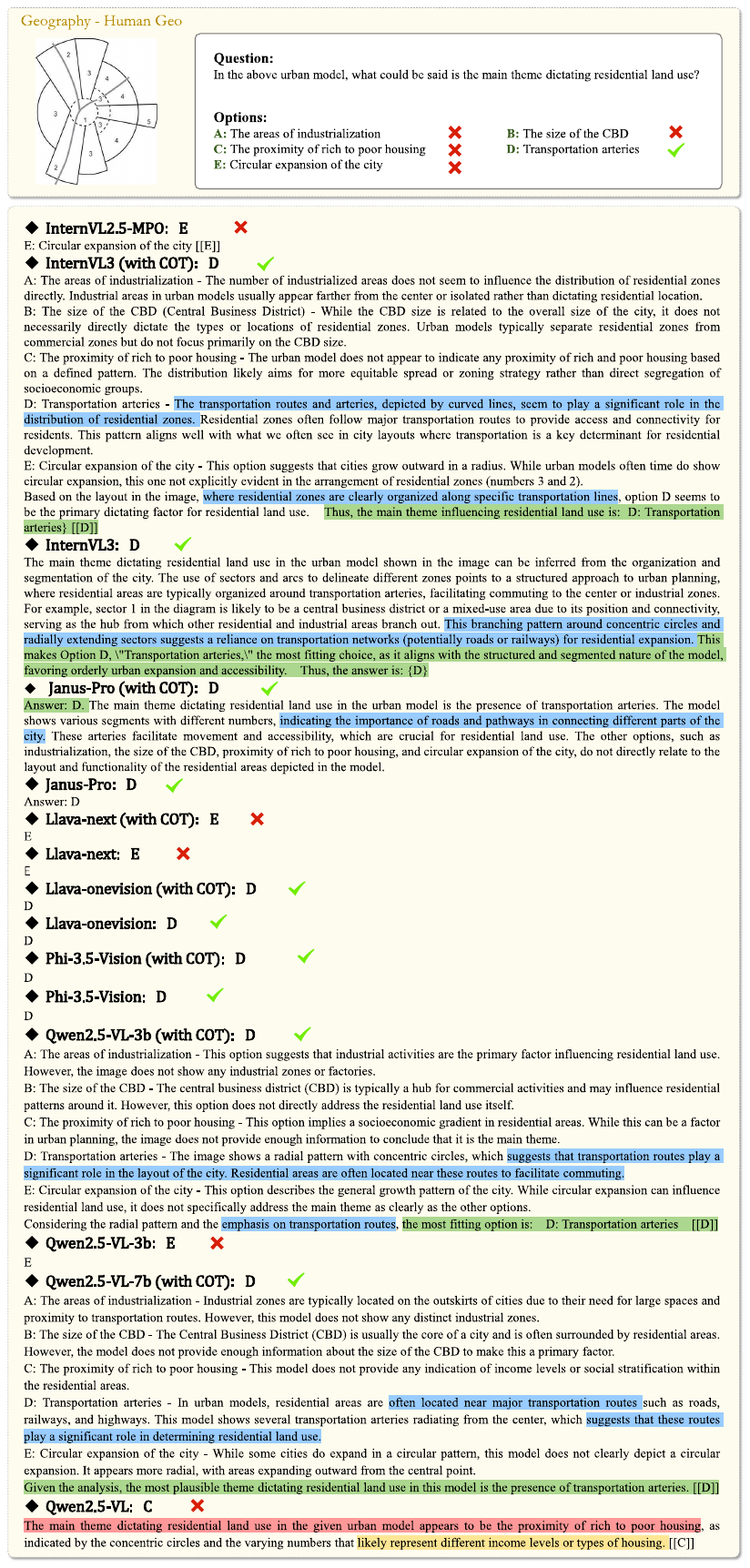}
\end{figure}

\begin{figure}
    \centering
    \includegraphics[width=1\linewidth]{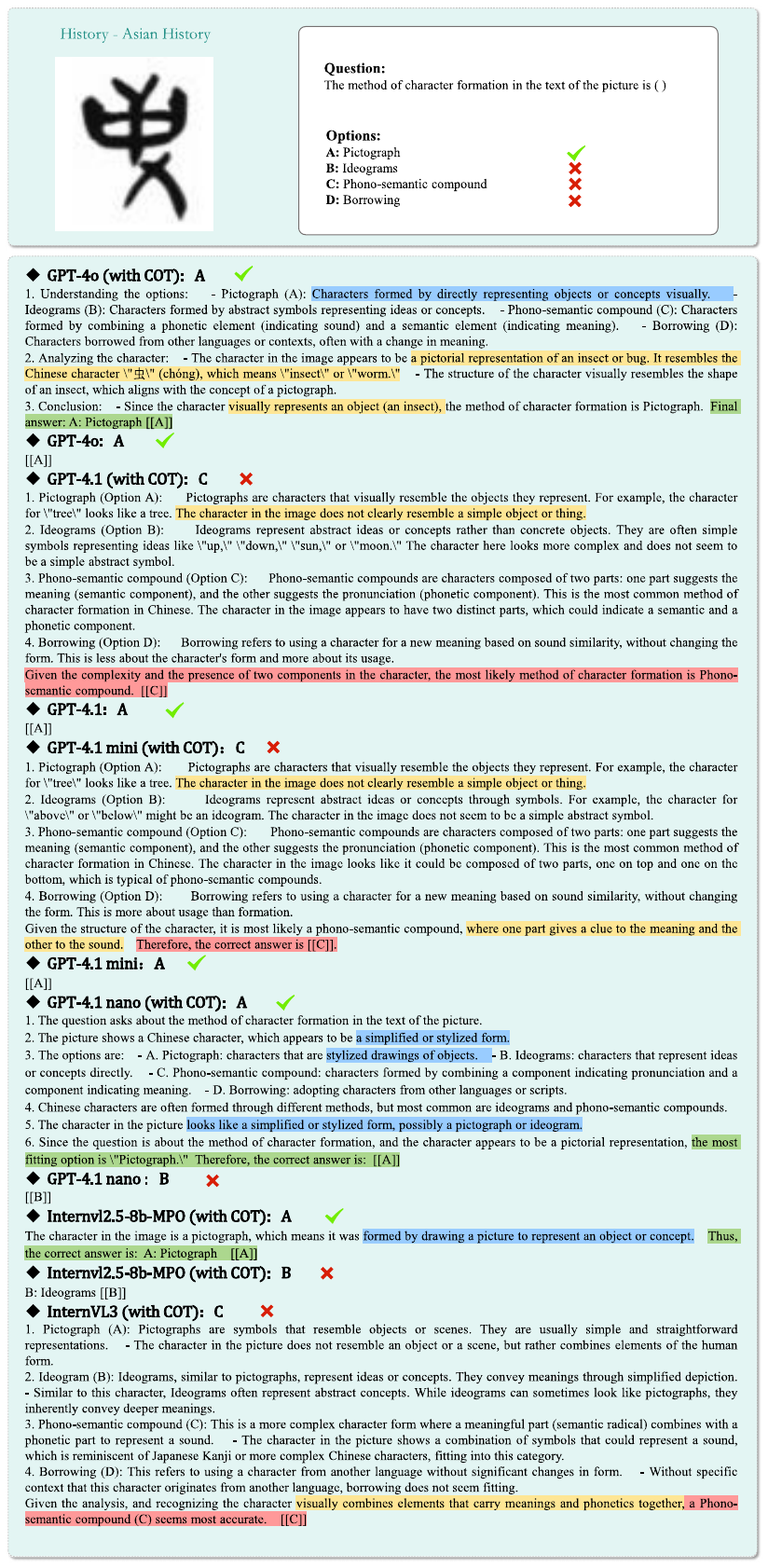}
\end{figure}
\begin{figure}
    \centering
    \includegraphics[width=1\linewidth]{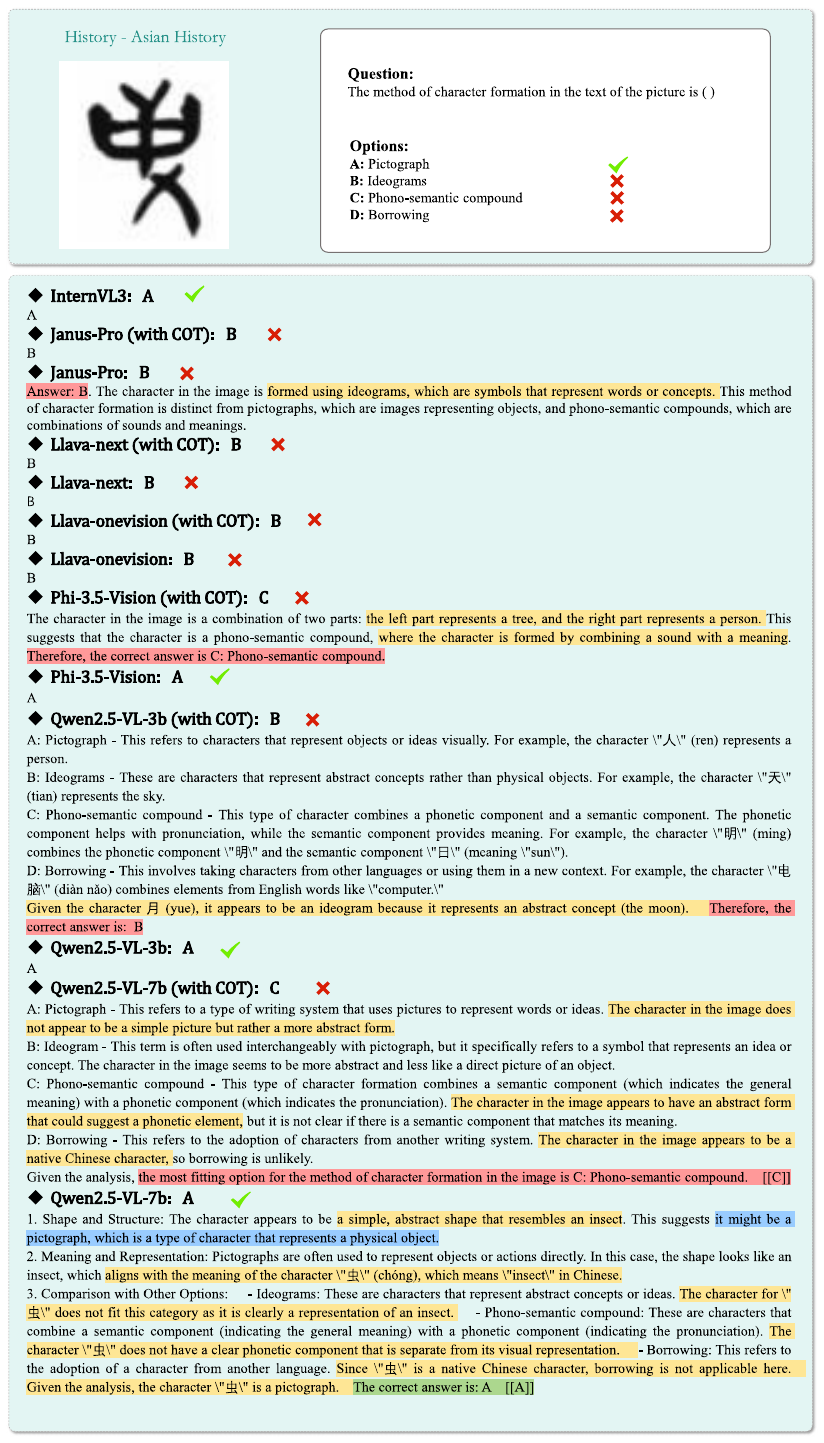}
\end{figure}

\begin{figure}
    \centering
    \includegraphics[width=1\linewidth]{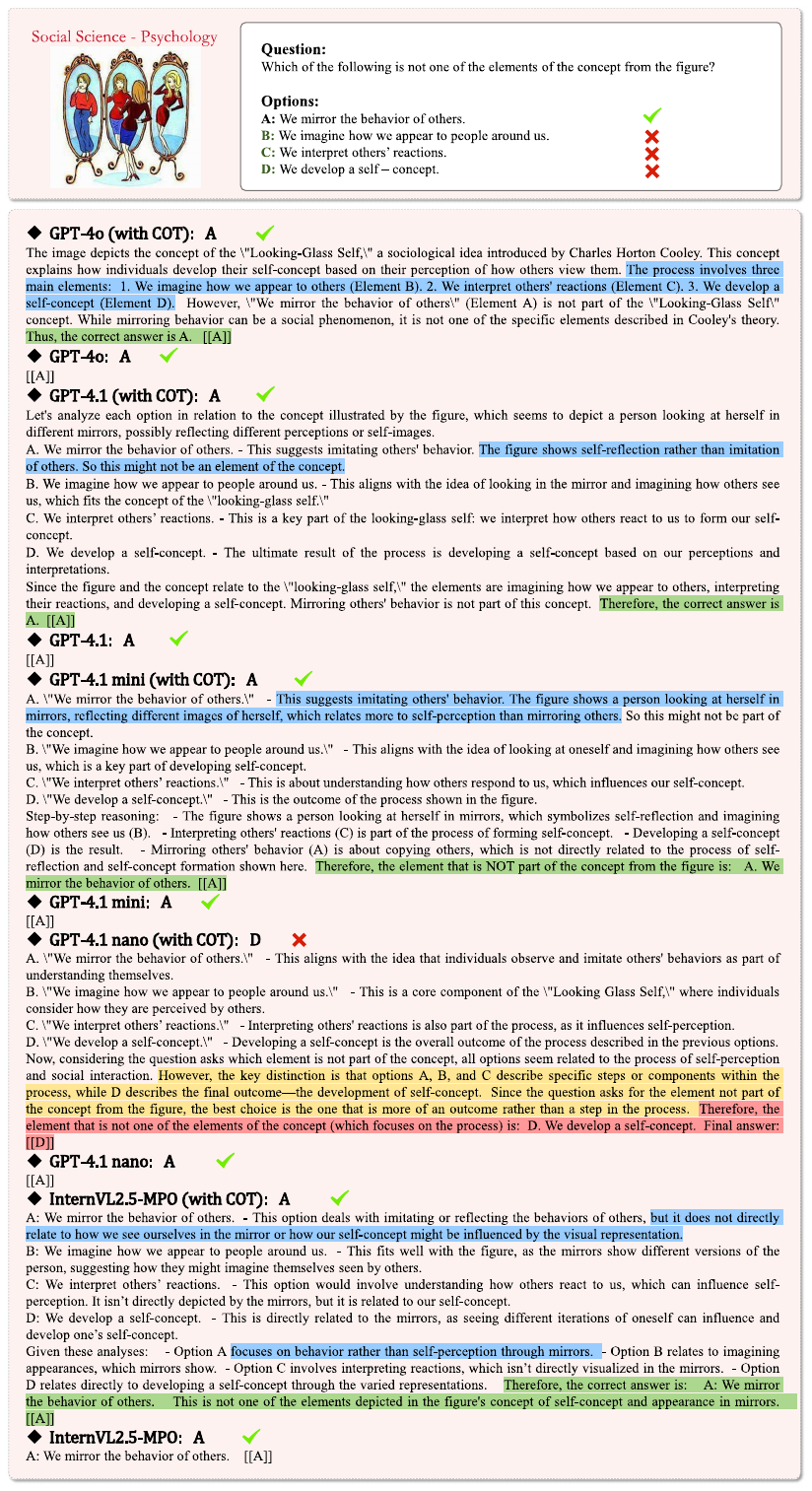}
\end{figure}

\begin{figure}
    \centering
    \includegraphics[width=1\linewidth]{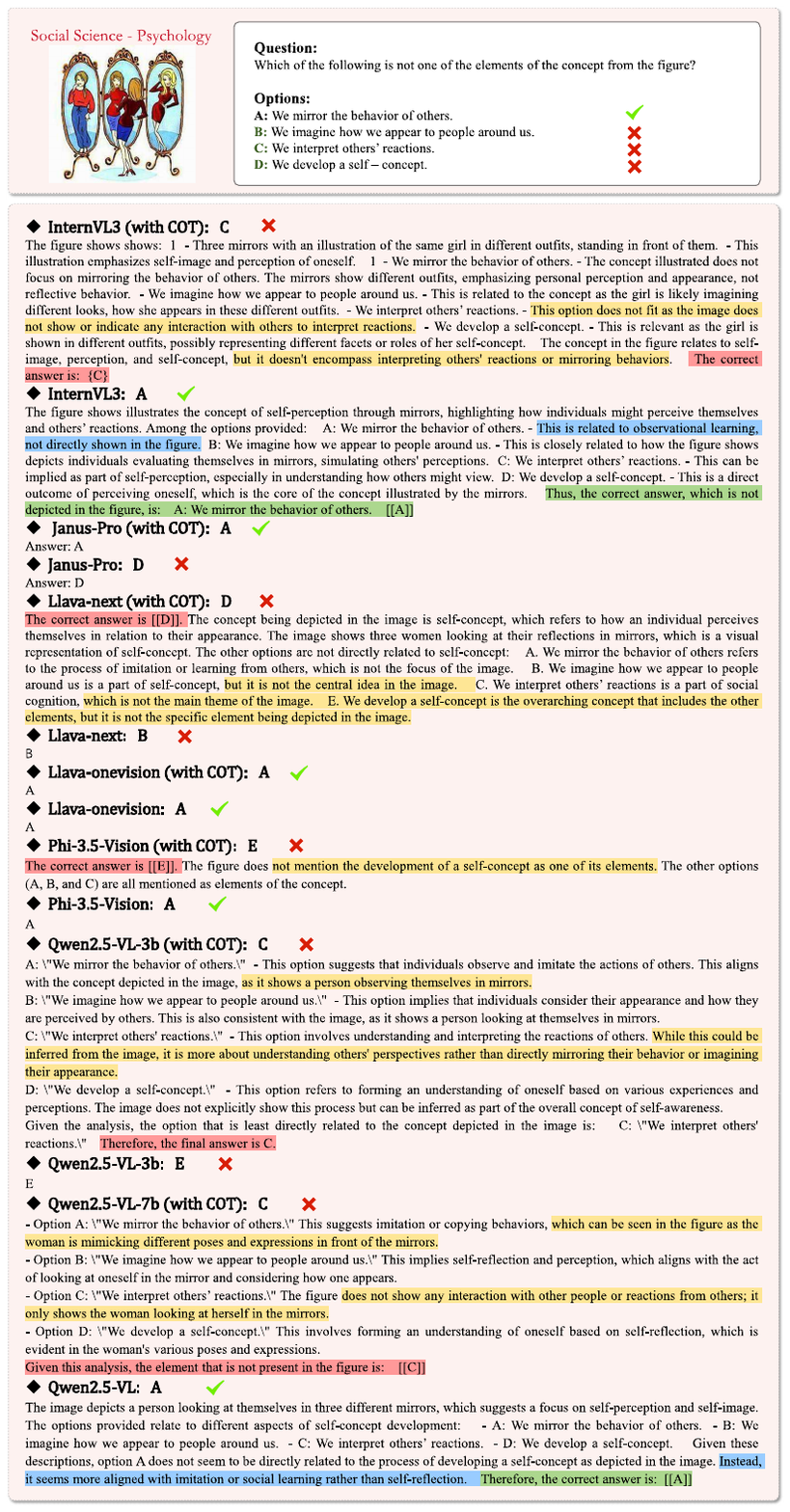}
\end{figure}

\begin{figure}
    \centering
    \includegraphics[width=1\linewidth]{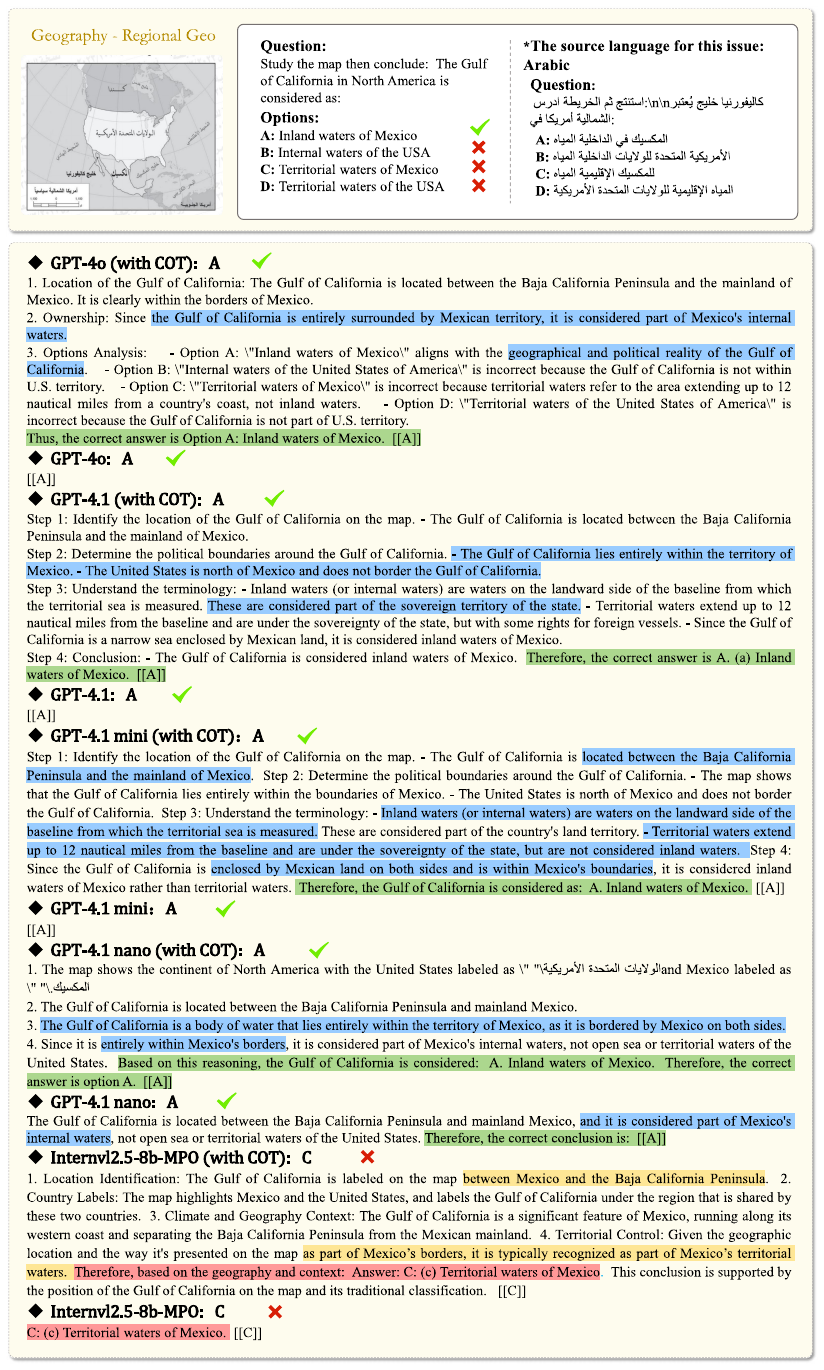}
\end{figure}

\begin{figure}
    \centering
    \includegraphics[width=1\linewidth]{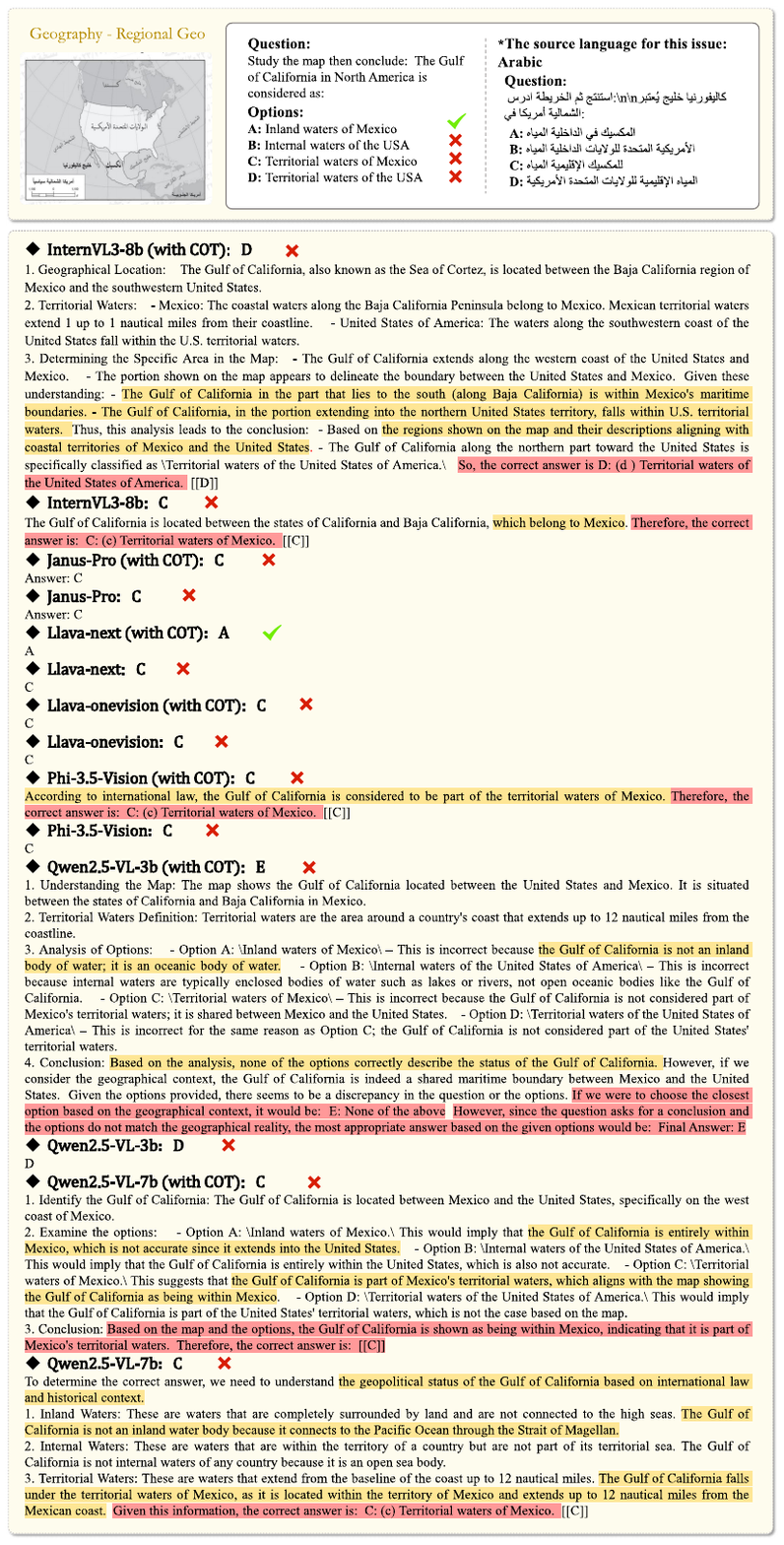}
\end{figure}

\begin{figure}
    \centering
    \includegraphics[width=1\linewidth]{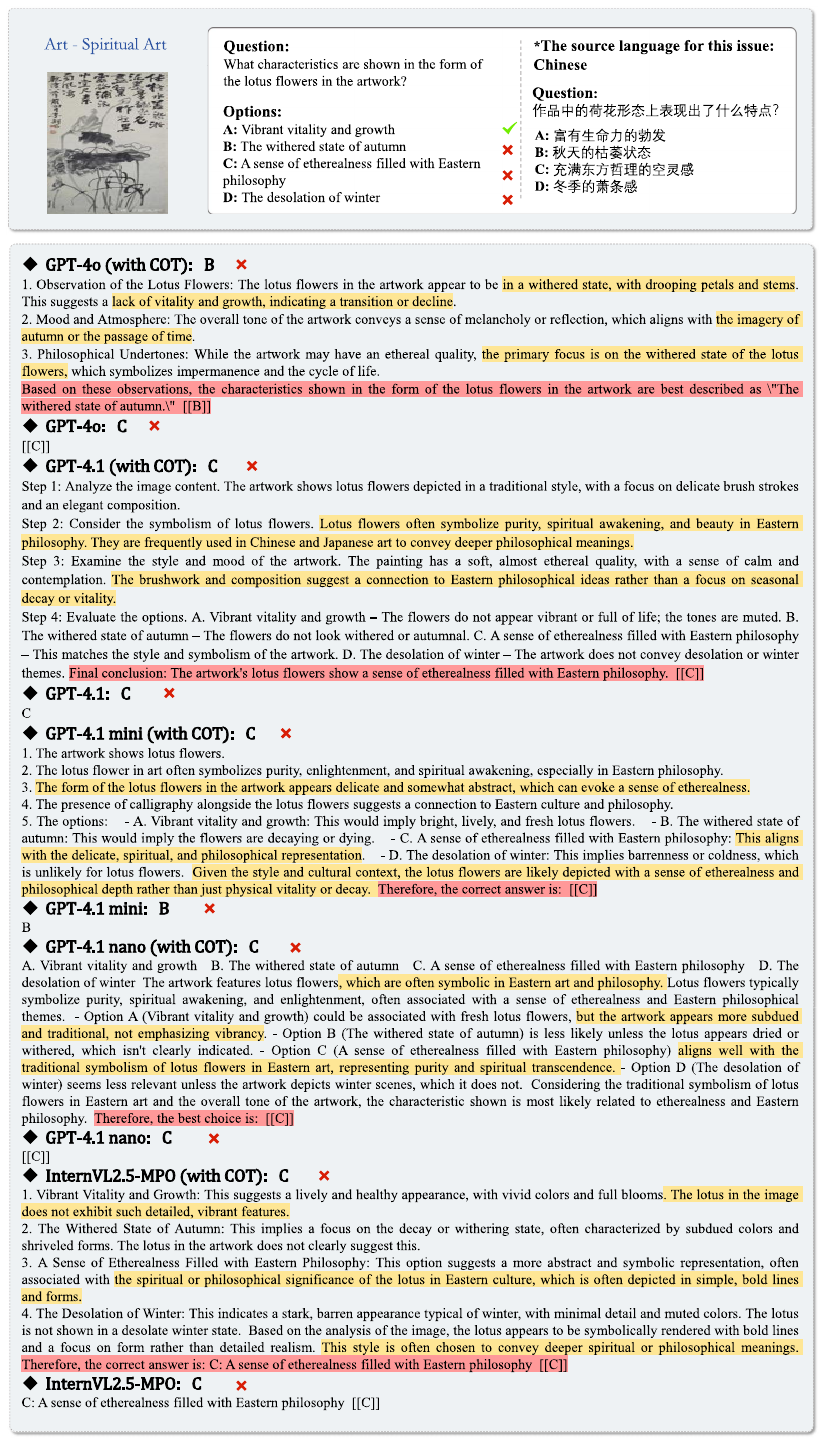}
\end{figure}

\begin{figure}
    \centering
    \includegraphics[width=1\linewidth]{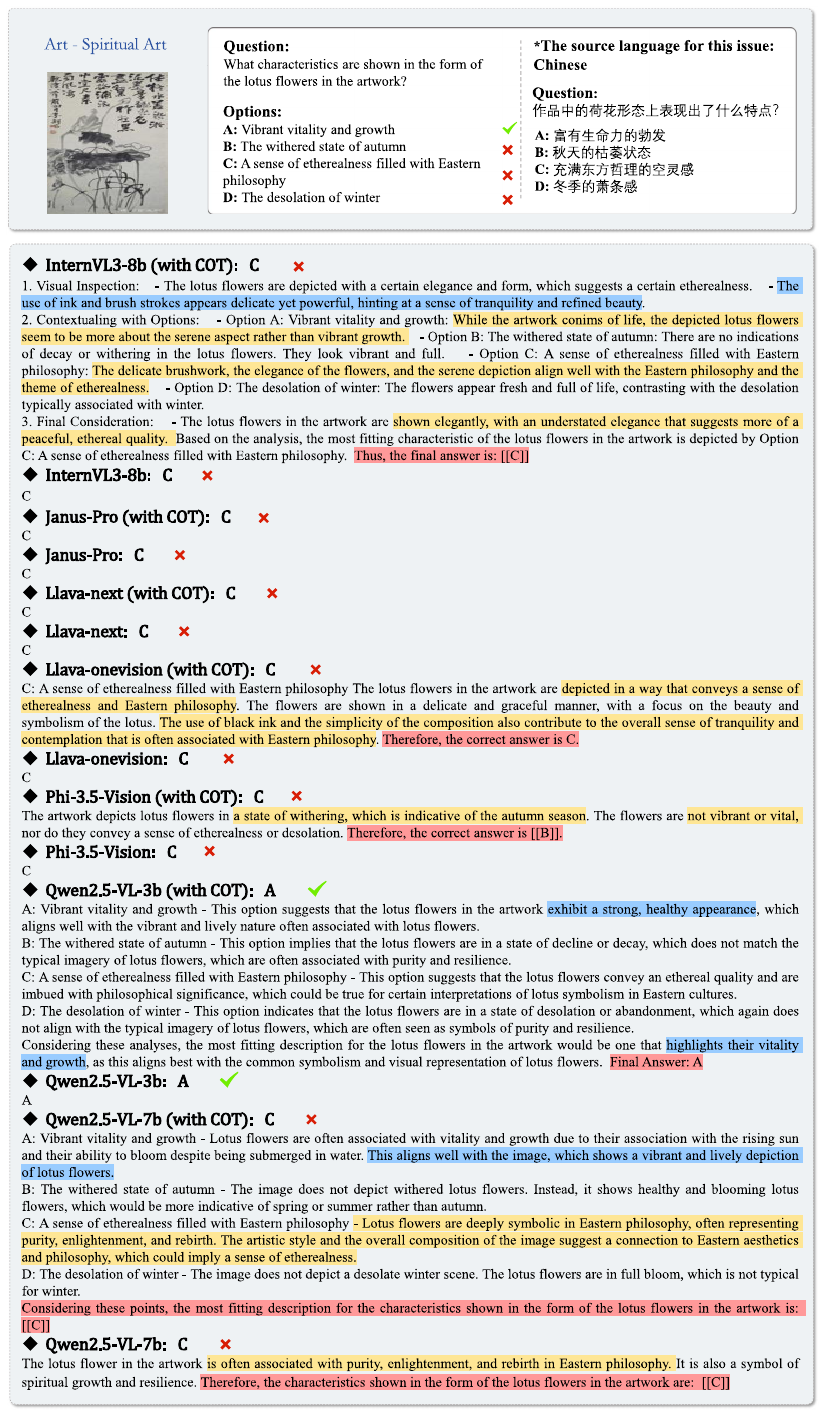}
\end{figure}

\begin{figure}
    \centering
    \includegraphics[width=1\linewidth]{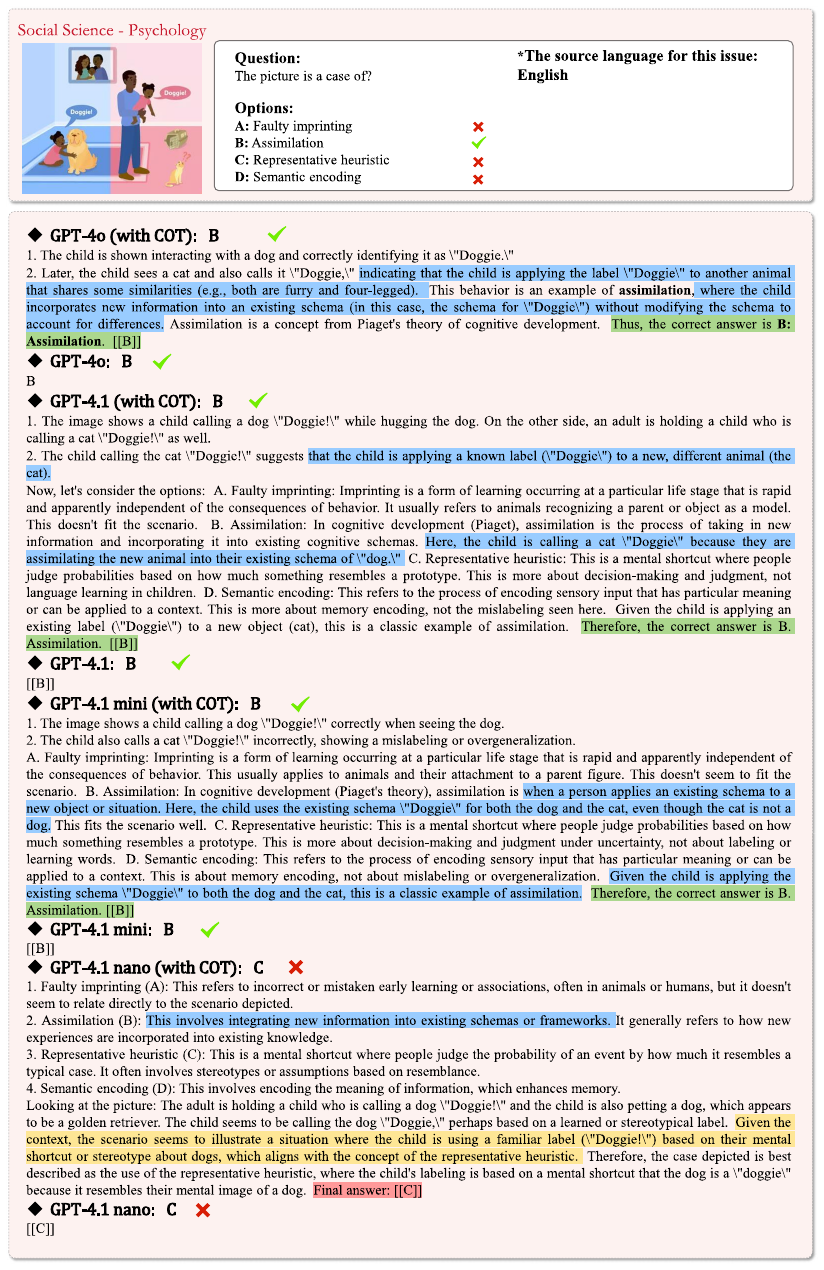}
\end{figure}

\begin{figure}
    \centering
    \includegraphics[width=1\linewidth]{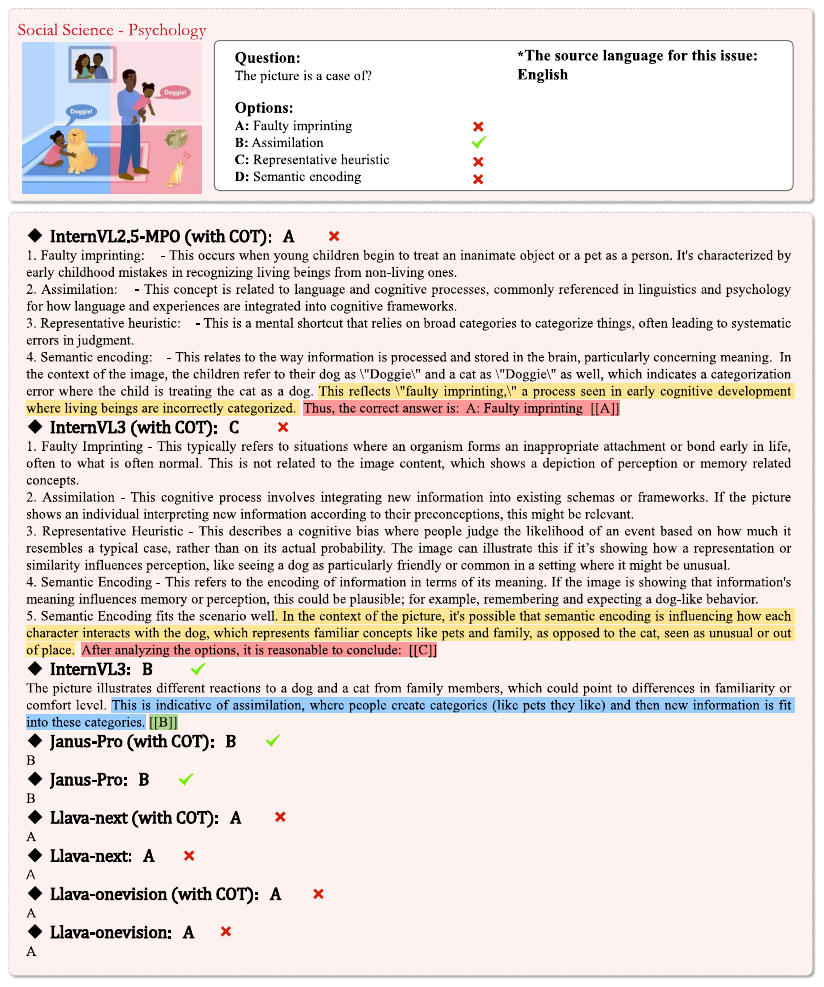}
\end{figure}
\begin{figure}
    \centering
    \includegraphics[width=1\linewidth]{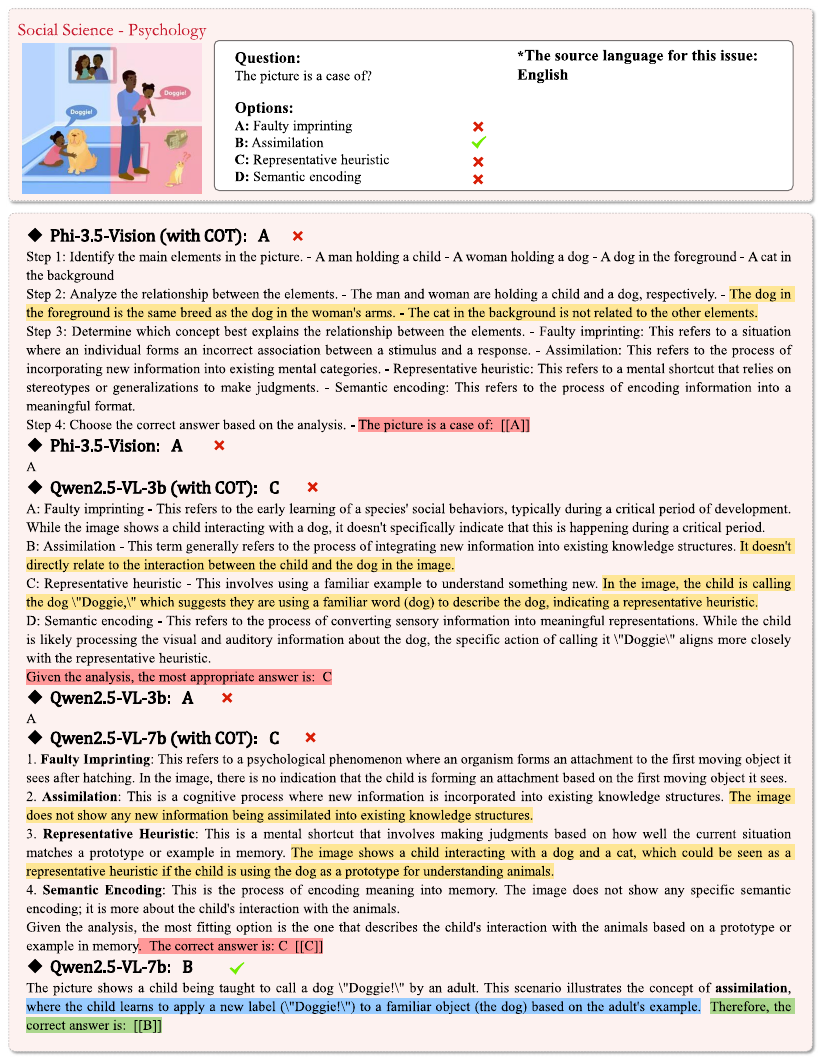}
\end{figure}

\begin{figure}
    \centering
    \includegraphics[width=1\linewidth]{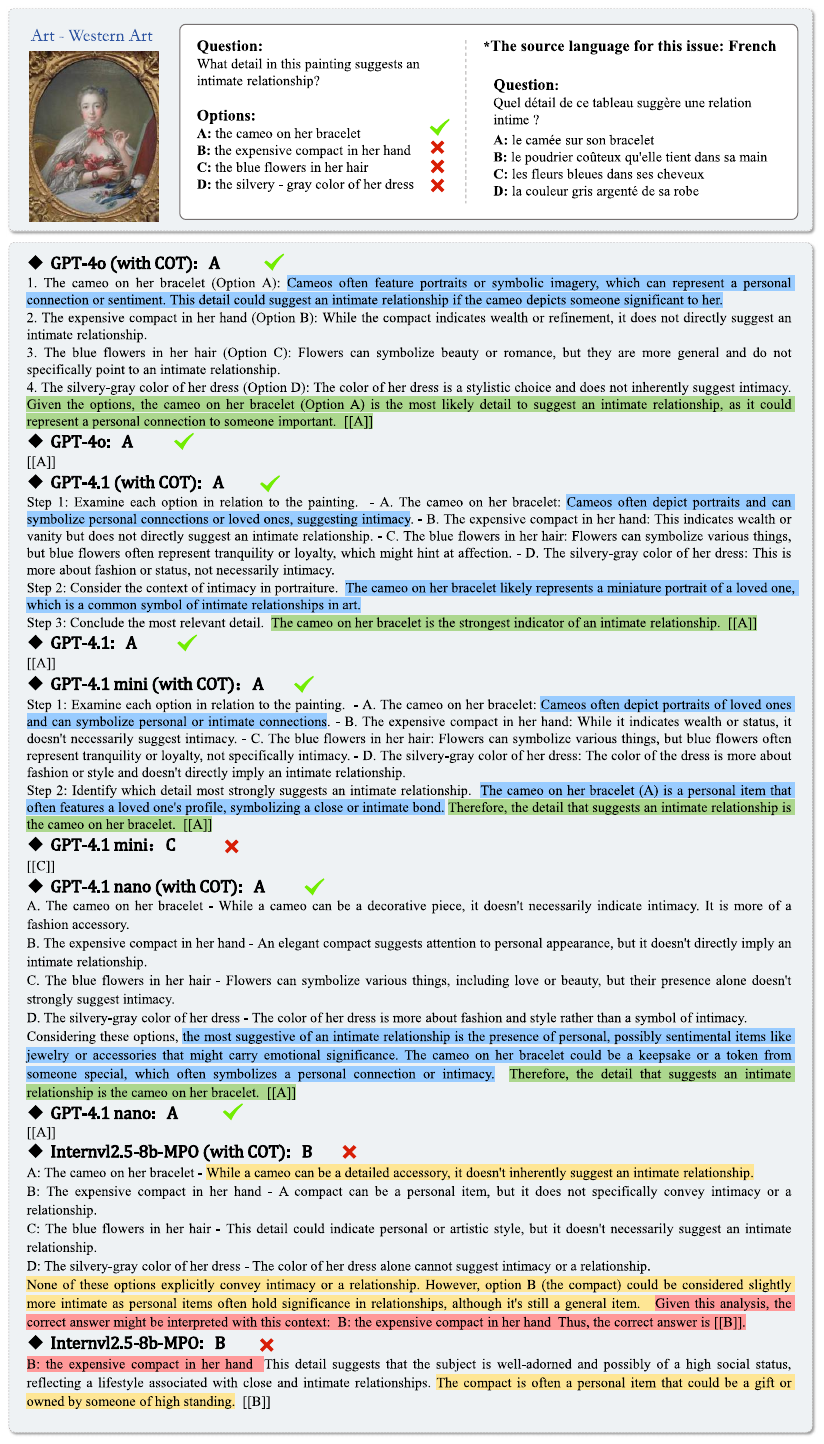}
\end{figure}

\begin{figure}
    \centering
    \includegraphics[width=1\linewidth]{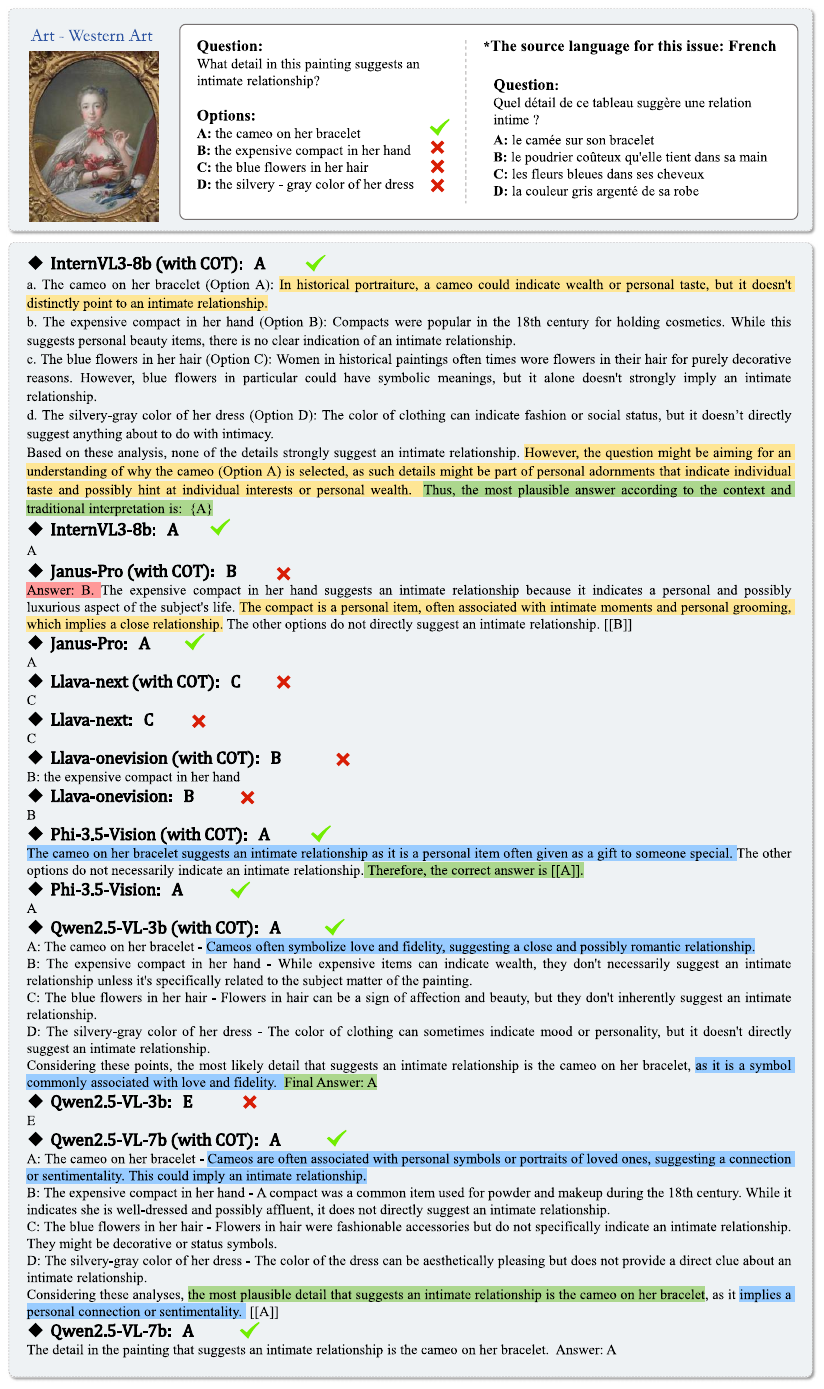}
\end{figure}

\begin{figure}
    \centering
    \includegraphics[width=1\linewidth]{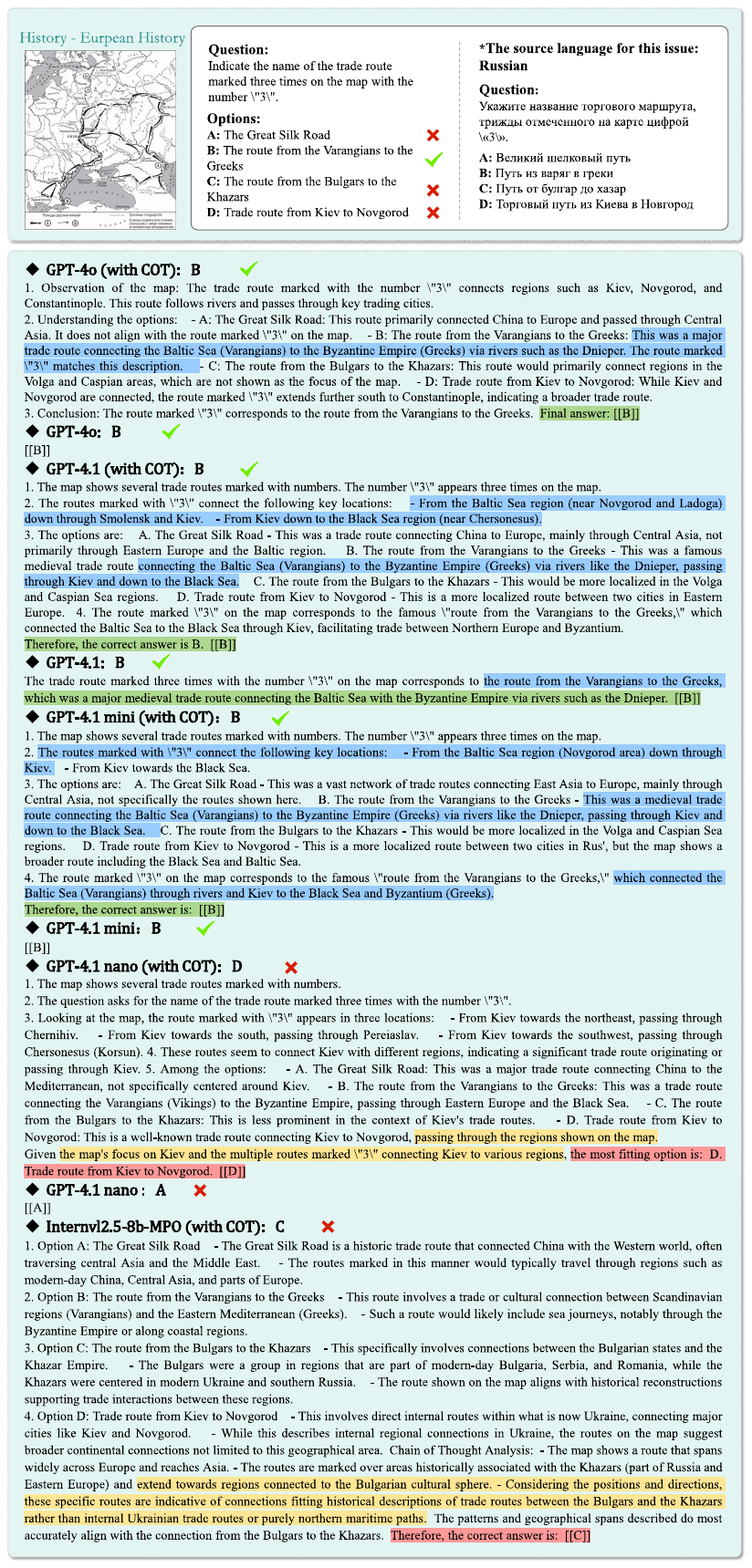}
\end{figure}

\begin{figure}
    \centering
    \includegraphics[width=1\linewidth]{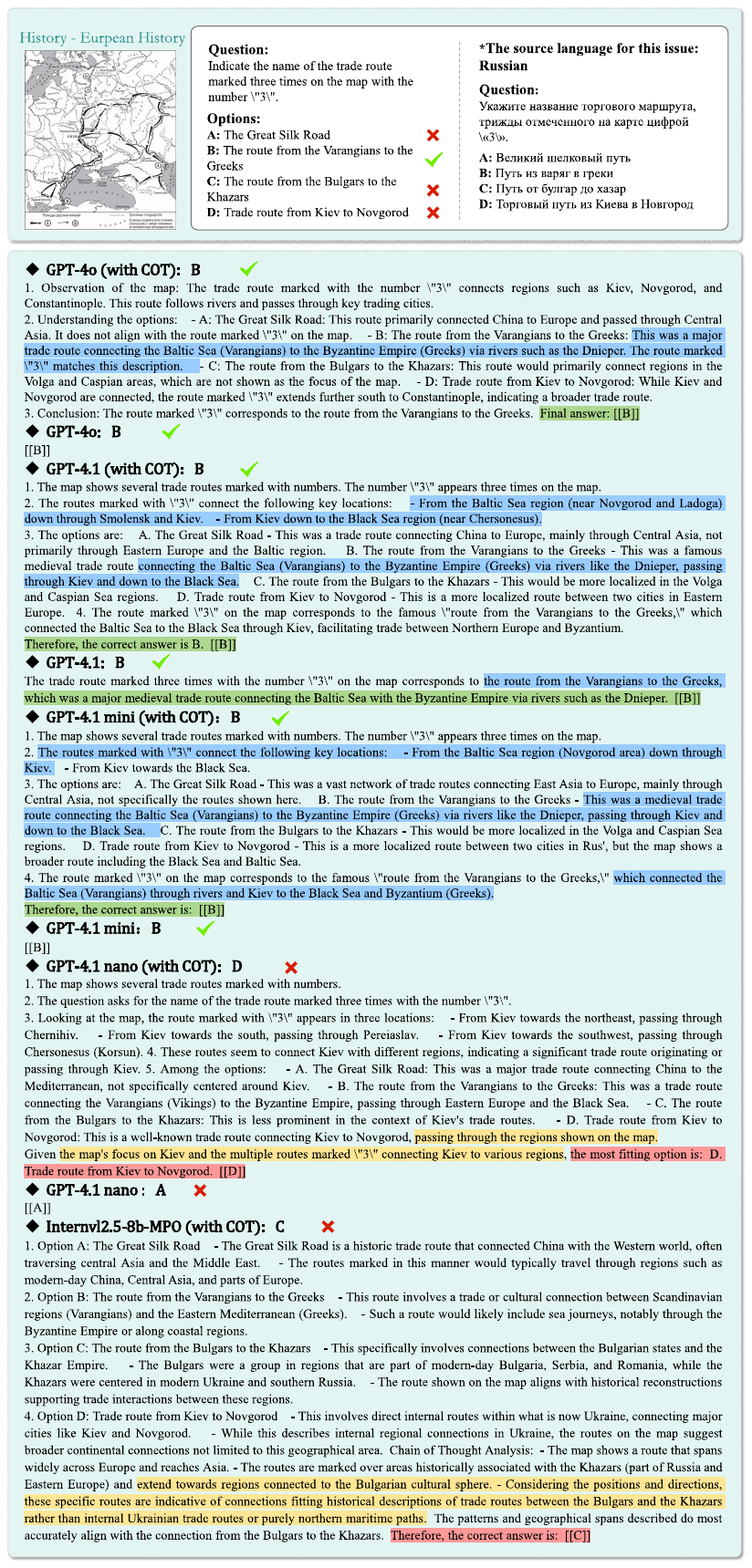}
\end{figure}
\begin{figure}
    \centering
    \includegraphics[width=1\linewidth]{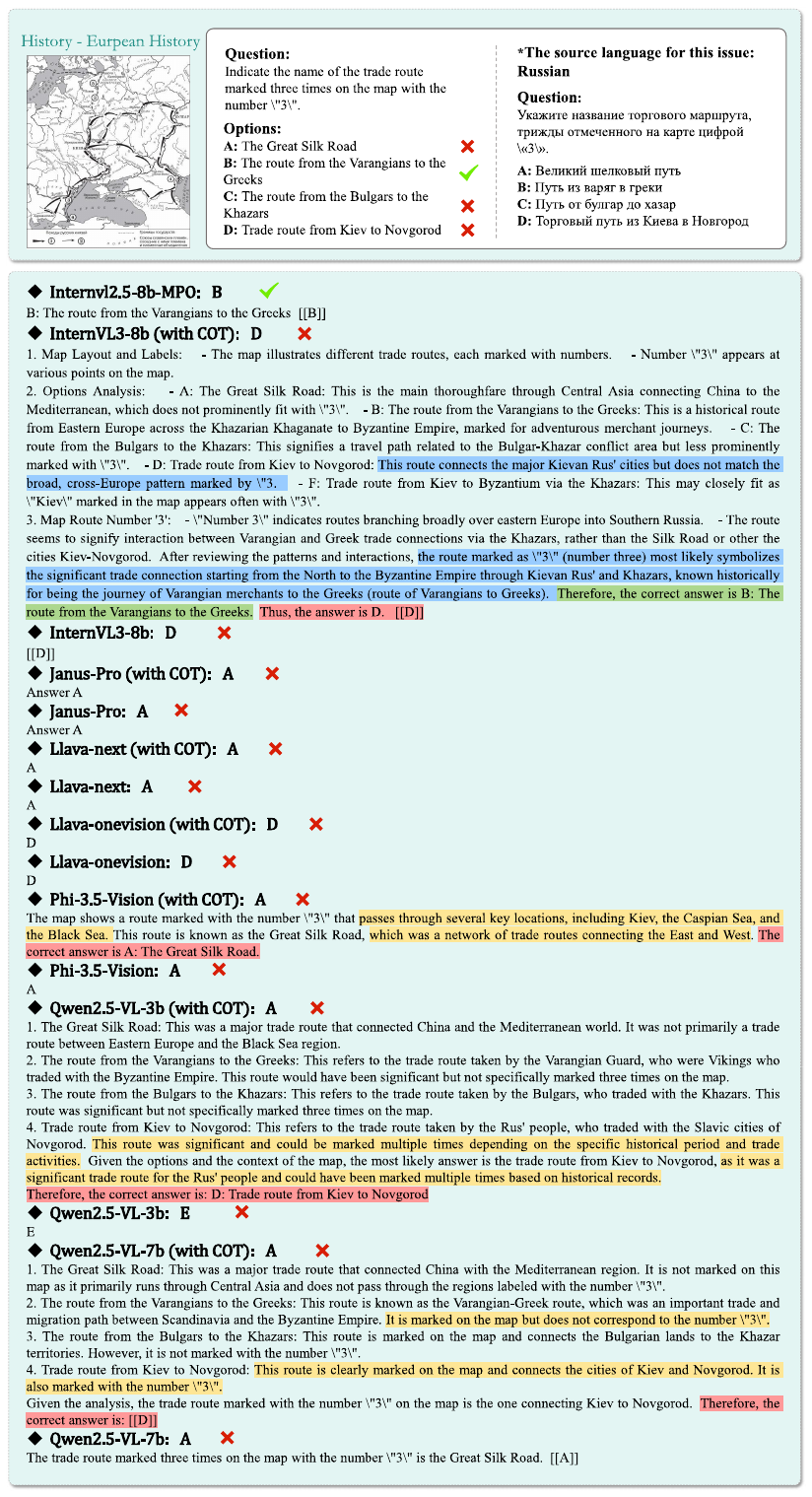}
\end{figure}

\begin{figure}
    \centering
    \includegraphics[width=1\linewidth]{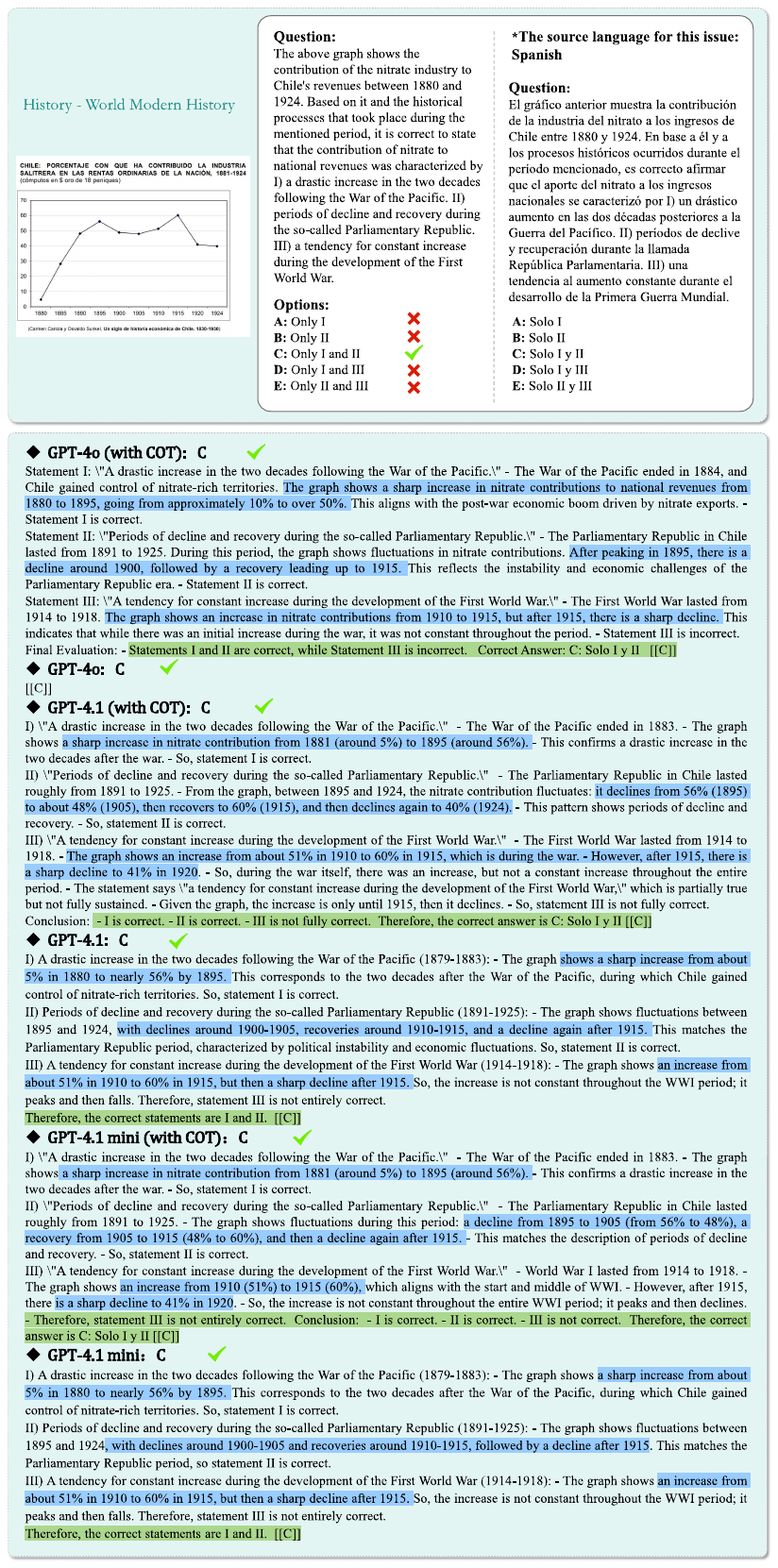}
\end{figure}

\begin{figure}
    \centering
    \includegraphics[width=1\linewidth]{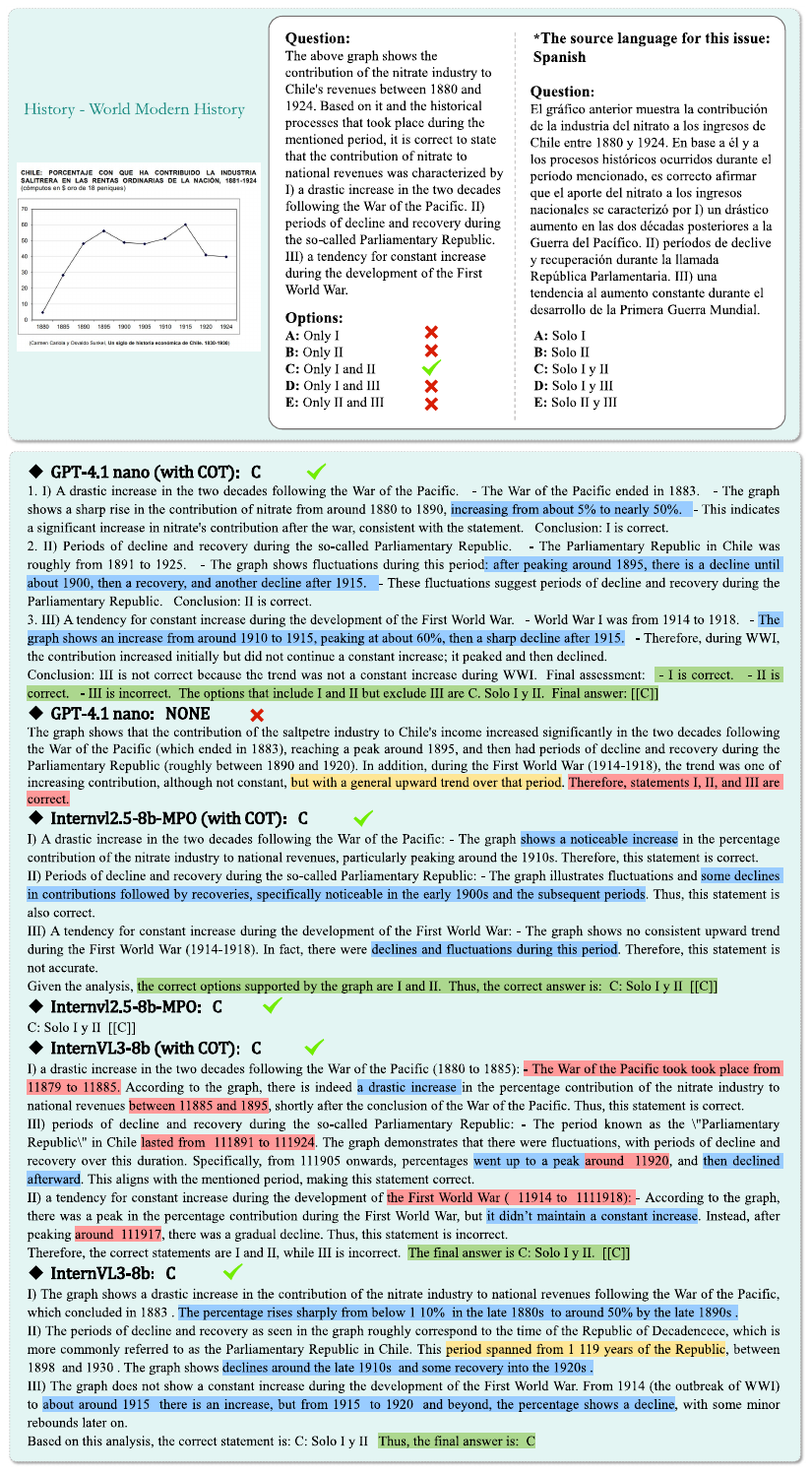}
\end{figure}
\begin{figure}
    \centering
    \includegraphics[width=1\linewidth]{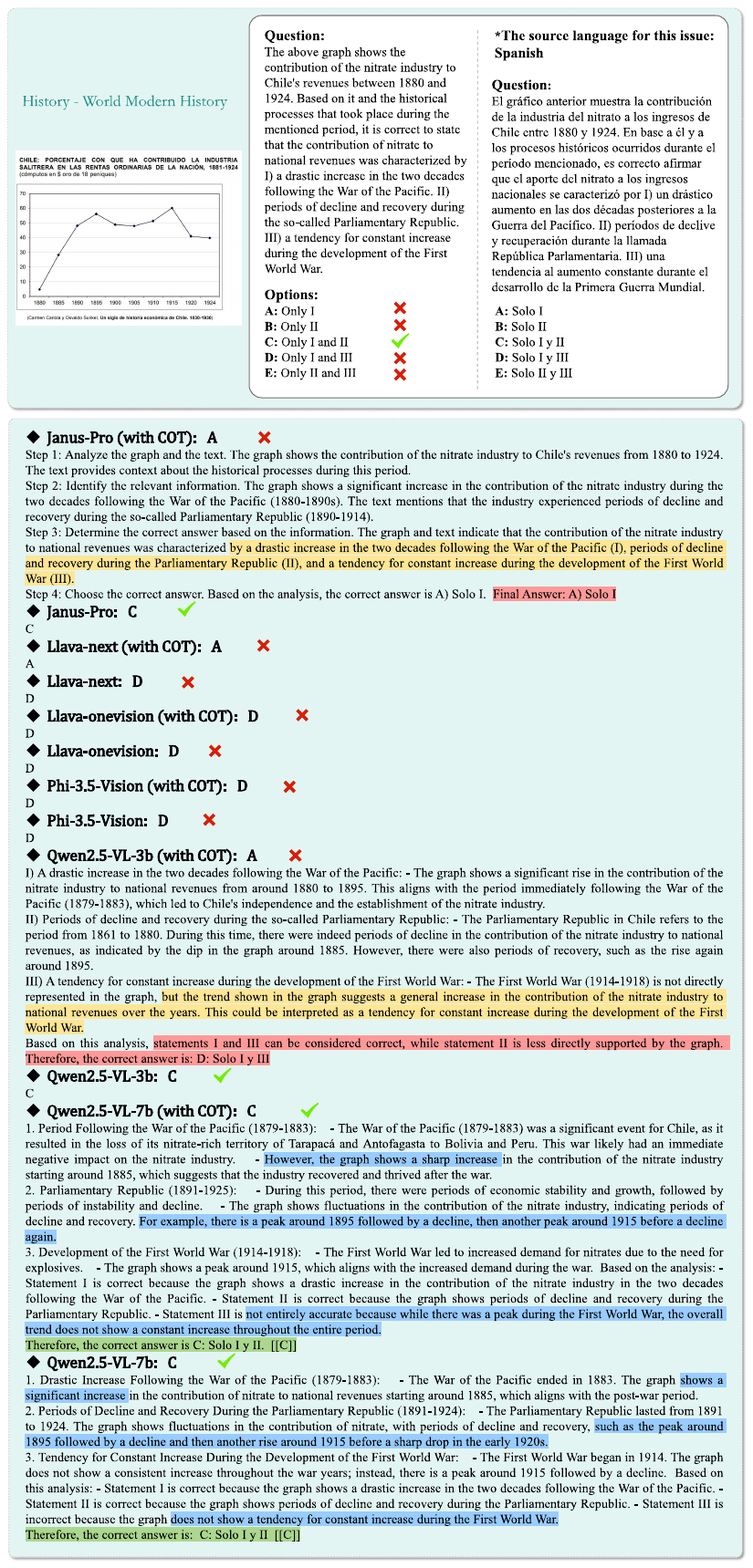}
\end{figure}

\end{document}